\documentclass{article}

\usepackage{arxiv}
\usepackage{comment}

\usepackage[utf8]{inputenc} 
\usepackage[T1]{fontenc}    
\usepackage{hyperref}       
\usepackage{url}            
\usepackage{longtable}
\usepackage{booktabs}       
\usepackage{multirow}
\usepackage{amsfonts}       
\usepackage{nicefrac}       
\usepackage{microtype}      
\usepackage{lipsum}
\usepackage{graphicx}
\usepackage{subcaption}
\usepackage{float}

\usepackage{xcolor}

\usepackage{authblk}

\newcommand{\head}[1]{\multicolumn{1}{c}{#1}}
\usepackage{caption} 
\usepackage{array}
\usepackage{etoolbox,siunitx}
\usepackage[normalem]{ulem}
\robustify\bfseries
\robustify\itshape
\robustify\uline
\robustify\slshape
\robustify\ttfamily
\def\Uline#1{#1\llap{\uline{\phantom{#1}}}}

\hypersetup{
   colorlinks=true
}

\title{Deep Learning in Video Multi-Object Tracking: A Survey}

\begin{document}

\author[1,2]{Gioele Ciaparrone}
\author[2]{Francisco Luque Sánchez}
\author[2]{Siham Tabik}
\author[3]{Luigi Troiano}
\author[1]{Roberto Tagliaferri}
\author[2]{Francisco Herrera}
\affil[1]{Department of Management and Innovation Systems, University of Salerno, 84084 Fisciano (SA), Italy}
\affil[2]{Andalusian Research Institute in Data Science and Computational Intelligence, University of Granada, 18071 Granada, Spain}
\affil[3]{Department of Engineering, University of Sannio, 82100 Benevento, Italy}
\affil[ ]{\texttt{\{\href{mailto:gciaparrone@unisa.it}{gciaparrone},\href{mailto:robtag@unisa.it}{robtag}\}@unisa.it, \{\href{mailto:fluque@decsai.ugr.es}{fluque},\href{mailto:herrera@decsai.ugr.es}{herrera}\}@decsai.ugr.es, \href{mailto:siham@ugr.es}{siham@ugr.es}, \href{mailto:troiano@unisannio.it}{troiano@unisannio.it}}}

\maketitle

\begin{abstract}
The problem of Multiple Object Tracking (MOT) consists in following the trajectory of different objects in a sequence, usually a video. In recent years, with the rise of Deep Learning, the algorithms that provide a solution to this problem have benefited from the representational power of deep models. This paper provides a comprehensive survey on works that employ Deep Learning models to solve the task of MOT on single-camera videos. Four main steps in MOT algorithms are identified, and an in-depth review of how Deep Learning was employed in each one of these stages is presented. A complete experimental comparison of the presented works on the three MOTChallenge datasets is also provided, identifying a number of similarities among the top-performing methods and presenting some possible future research directions.
\end{abstract}


\newpage
\section{Introduction}
\begin{figure}[ht]
    \centering
    \includegraphics[width=.55\textwidth]{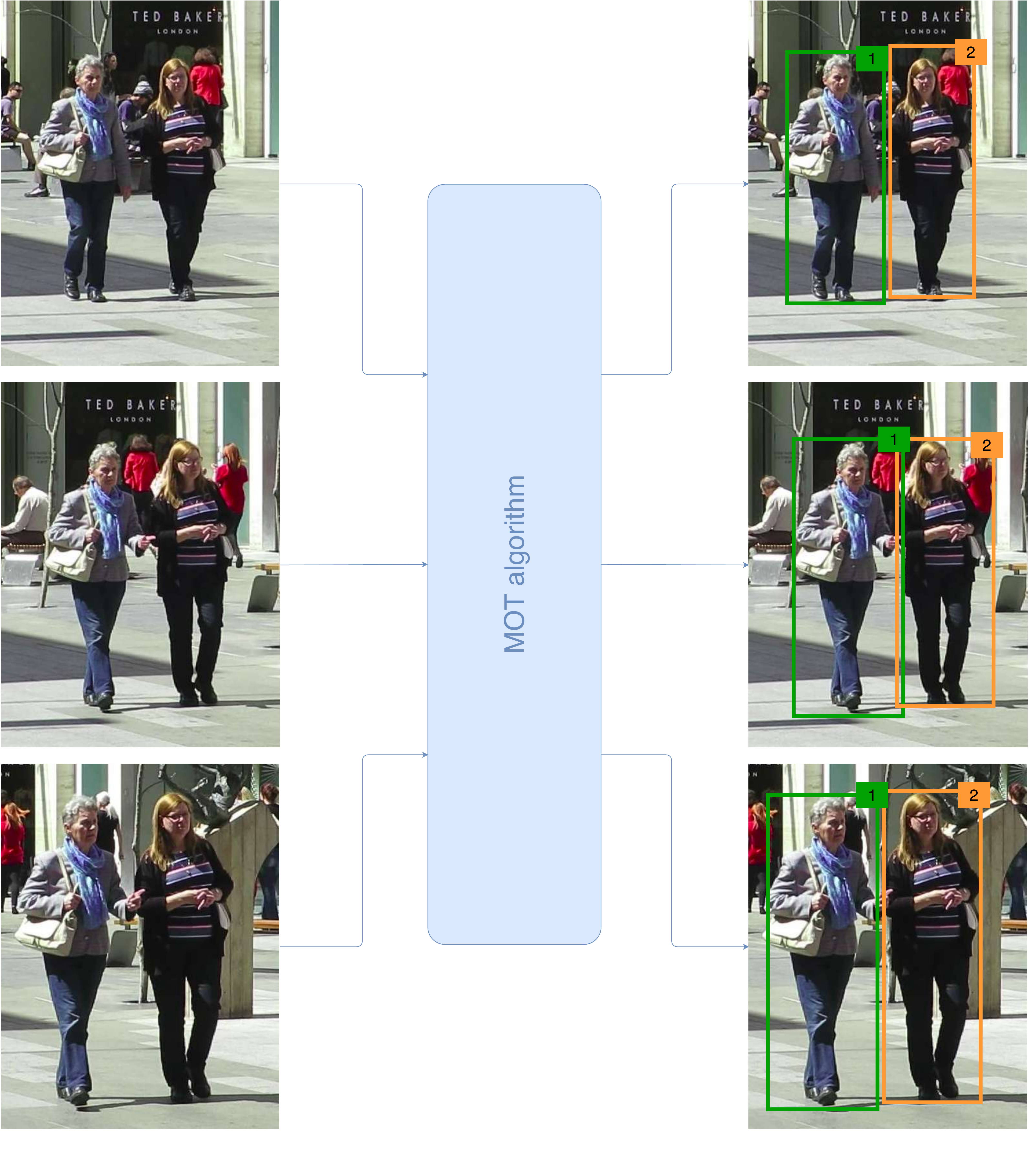}
    \caption{An illustration of the output of a MOT algorithm. Each output bounding box has a number that identifies a specific person in the video.}
    \label{fig:mot_algorithm_output}
\end{figure}

Multiple Object Tracking (MOT), also called Multi-Target Tracking (MTT), is a computer vision task that aims to analyze videos in order to identify and track objects belonging to one or more categories, such as pedestrians, cars, animals and inanimate objects, without any prior knowledge about the appearance and number of targets. Differently from object detection algorithms, whose output is a collection of rectangular bounding boxes identified by their coordinates, height and width, MOT algorithms also associate a target ID to each box (known as a \textit{detection}), in order to distinguish among intra-class objects. An example of the output of a MOT algorithm is illustrated in figure \ref{fig:mot_algorithm_output}. The MOT task plays an important role in computer vision: from video surveillance to autonomous cars, from action recognition to crowd behaviour analysis, many of these problems would benefit from a high-quality tracking algorithm.

While in Single Object Tracking (SOT) the appearance of the target is known a priori, in MOT a detection step is necessary to identify the targets, that can leave or enter the scene. The main difficulty in tracking multiple targets simultaneously stems from the various occlusions and interactions between objects, that can sometimes also have similar appearance. Thus, simply applying SOT models directly to solve MOT leads to poor results, often incurring in target drift and numerous ID switch errors, as such models usually struggle in distinguishing between similar looking intra-class objects. A series of algorithms specifically tuned to multi-target tracking have then been developed in recent years to address these issues, together with a number of benchmark datasets and competitions to ease the comparisons between the different methods.

Recently, more and more of such algorithms have started exploiting the representational power of deep learning (DL). The strength of Deep Neural Networks (DNN) resides in their ability to learn rich representations and to extract complex and abstract features from their input. Convolutional neural networks (CNN) currently constitute the state-of-the-art in spatial pattern extraction, and are employed in tasks such as image classification \cite{simonyan2014very, szegedy2015going,  he2016deep} or object detection \cite{ren2015faster, liu2016ssd, redmon2017yolo9000}, while recurrent neural networks (RNN) like the Long Short-Term Memory (LSTM) are used to process sequential data, like audio signals, temporal series and text \cite{sak2014long, sundermeyer2012lstm, fan2014tts, marchi2014multi}. Since DL methods have been able to reach top performance in many of those tasks, we are now progressively seeing them used in most of the top performing MOT algorithms, aiding to solve some of the subtasks in which the problem is divided.

This work presents a survey of algorithms that make use of the capabilities of deep learning models to perform Multiple Object Tracking, focusing on the different approaches used for the various components of a MOT algorithm and putting them in the context of each of the proposed methods. While the MOT task can be applied to both 2D and 3D data, and to both single-camera and multi-camera scenarios, in this survey we focus on 2D data extracted from videos recorded by a single camera.

Some reviews and surveys have been published on the subject of MOT. Their main contributions and limitations are the following:

\begin{itemize}
    \item Luo et al. \cite{luo2014multiple} presented the first comprehensive review to focus specifically on MOT, in particular on pedestrian tracking. They provided a unified formulation of the MOT problem and described the main techniques used in the key steps of a MOT system. They presented deep learning as one of the future research directions, since at the time it had only been employed by very few algorithms.

    \item Camplani et al. \cite{camplani2016multiple} presented a survey on Multiple Pedestrian Tracking, but they focused on RGB-D data, while our focus is on 2D RGB images, without additional inputs. Moreover, their review does not cover deep learning based algorithms.

    \item Emami et al. \cite{emami2018machine} proposed a formulation of single and multi-sensor tracking tasks as a Multidimensional Assignment Problem (MDAP). They also presented a few approaches that employed deep learning in tracking problems, but it wasn't the focus of their paper and they didn't provide any experimental comparison among such methods. 

    \item Leal-Taixé et al. \cite{leal2017tracking} presented an analysis of the results obtained by algorithms on the MOT15 \cite{leal2015motchallenge} and MOT16 \cite{milan2016mot16} datasets, providing a summary of the trending lines of research and statistics about the results. They found that after 2015, methods have been shifting from trying to find better optimization algorithms for the association problem to focusing on improving the affinity models, and they predict that many more approaches would tackle this issue by using deep learning. However, this work also did not focus on deep learning, and it does not cover more recent MOT algorithms, published in the last years.
\end{itemize}

In this paper, based on the discussed limitations, our aim is to provide a survey with the following main contributions:

\begin{itemize}
    \item We provide the first comprehensive survey on the use of Deep Learning in Multiple Object Tracking, focusing on 2D data extracted from single-camera videos, including recent works that have not been covered by past surveys and reviews. The use of DL in MOT is in fact recent, and many approaches have been published in the last three years.
    \item We identify four common steps in MOT algorithms and describe the different DL models and approaches employed in each of those steps, including the algorithmic context in which they are used. The techniques utilized by each analyzed work are also summarized in a table, together with links to the available source code, to serve as a quick reference for future research.
    \item We collect experimental results on the most commonly used MOT datasets to perform a numerical comparison among them, also identifying the main trends in the best performing algorithms.
    \item As final point, we discuss the possible future directions of research.
\end{itemize}

The survey is further organized in this manner. We first describe the general structure of MOT algorithms and the most commonly used metrics and datasets in section \ref{sec:steps}. Section \ref{sec:main} explores the various DL-based models and algorithms in each of the four identified steps of a MOT algorithm. Section \ref{sec:comparisons} presents a numerical comparison among the presented algorithms and identifies common trends and patterns in current approaches, as well as some limitations and possible future research directions. Finally, section \ref{sec:conclusion} summarizes the findings of the previous sections and presents some final remarks.

\section{MOT: algorithms, metrics and datasets}\label{sec:steps}

In this section, a general description about the problem of MOT is provided. The main characteristics and common steps of MOT algorithms are identified and described in section \ref{sec:motsteps}. The metrics that are usually employed to evaluate the performance of the models are discussed in section \ref{sec:metrics}, while the most important benchmark datasets are presented in section \ref{sec:datasets}.

\subsection{Introduction to MOT algorithms} \label{sec:motsteps}
The standard approach employed in MOT algorithms is \textit{tracking-by-detection}: a set of detections (i.e. bounding boxes identifying the targets in the image) are extracted from the video frames and are used to guide the tracking process, usually by associating them together in order to assign the same ID to bounding boxes that contain the same target. For this reason, many MOT algorithms formulate the task as an assignment problem. Modern detection frameworks \cite{ren2015faster, he2017mask, dai2016r, liu2016ssd, redmon2017yolo9000} ensure a good detection quality, and the majority of MOT methods (with some exceptions, as we will see) have been focusing on improving the association; indeed, many MOT datasets provide a standard set of detections that can be used by the algorithms (that can thus skip the detection stage) in order to exclusively compare their performances on the quality of the association algorithm, since the detector performance can heavily affect the tracking results.

MOT algorithms can also be divided into batch and online methods. Batch tracking algorithms are allowed to use future information (i.e. from future frames) when trying to determine the object identities in a certain frame. They often exploit global information and thus result in better tracking quality. Online tracking algorithms, on the contrary, can only use present and past information to make predictions about the current frame. This is a requirement in some scenarios, like autonomous driving and robot navigation. Compared to batch methods, online methods tend to perform worse, since they cannot fix past errors using future information. It is important to note that while a real-time algorithm is required to run in an online fashion, not every online method necessarily runs in real-time; quite often, in fact, with very few exceptions, online algorithms are still too slow to be employed in a real-time environment, especially when exploiting deep learning algorithms, that are often computationally intensive.

\begin{figure}[htb]
    \centering
    \includegraphics[width=.7\textwidth]{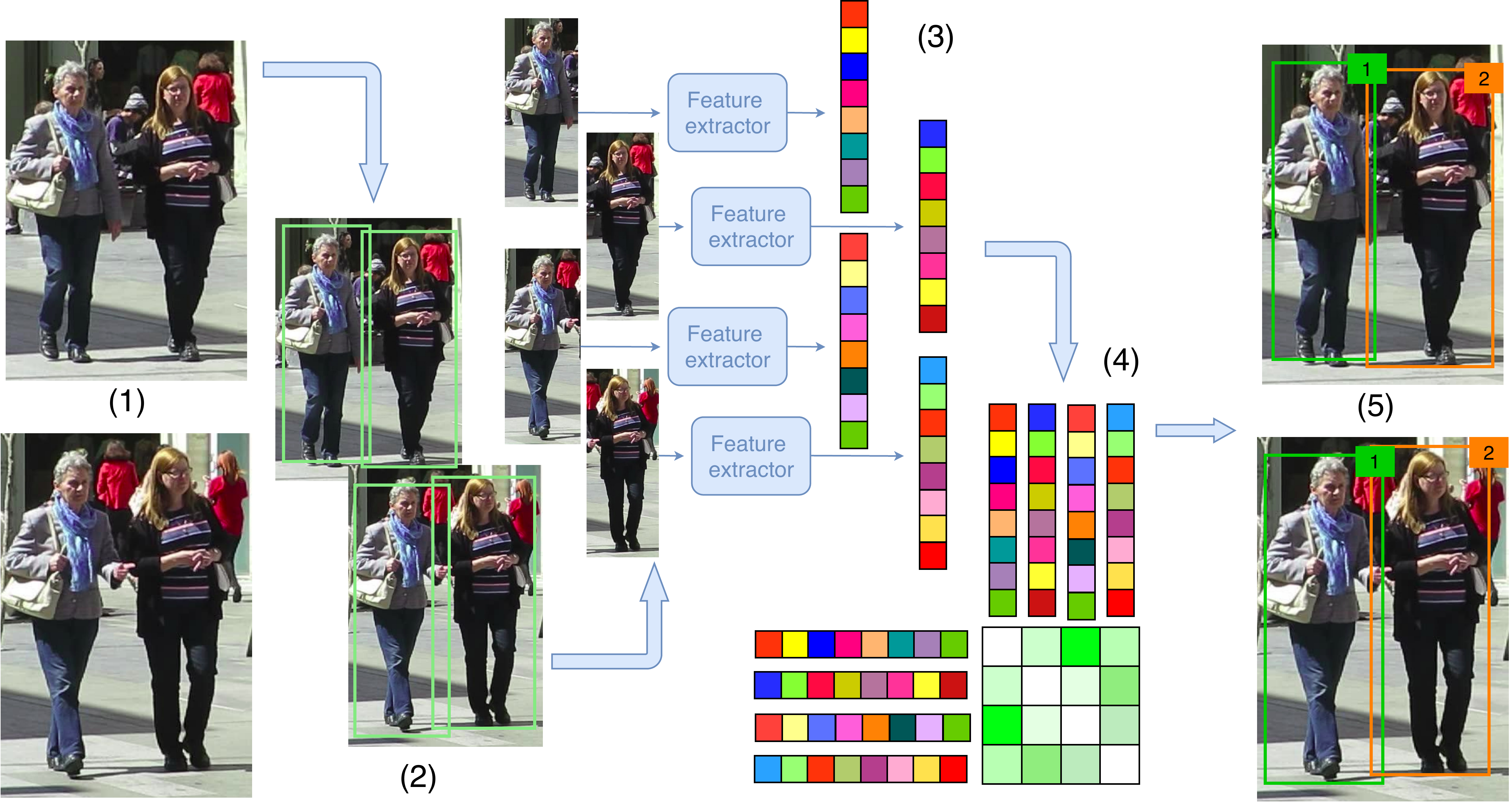}
    \caption{Usual workflow of a MOT algorithm: given the raw frames of a video (1), an object detector is run to obtain the bounding boxes of the objects (2). Then, for every detected object, different features are computed, usually visual and motion ones (3). After that, an affinity computation step calculates the probability of two objects belonging to the same target (4), and finally an association step assigns a numerical ID to each object (5).}
    \label{fig:four_steps}
\end{figure}

Despite the huge variety of approaches presented in the literature, the vast majority of MOT algorithms share part or all of the following steps (summarized in figure \ref{fig:four_steps}):
\begin{itemize}
    \item Detection stage: an object detection algorithm analyzes each input frame to identify objects belonging to the target class(es) using bounding boxes, also known as `detections' in the context of MOT;
    \item Feature extraction/motion prediction stage: one or more feature extraction algorithms analyze the detections and/or the tracklets to extract appearance, motion and/or interaction features. Optionally, a motion predictor predicts the next position of each tracked target;
    \item Affinity stage: features and motion predictions are used to compute a similarity/distance score between pairs of detections and/or tracklets;
    \item Association stage: the similarity/distance measures are used to associate detections and tracklets belonging to the same target by assigning the same ID to detections that identify the same target.
\end{itemize}

While these stages can be performed sequentially in the order presented here (often once per frame for online methods and once for the whole video for batch methods), there are many algorithms that merge some of these steps together, or intertwine them, or even perform them multiple times using different techniques (e.g. in algorithms that work in two phases). Moreover, some methods do not directly associate detections together, but use them to refine trajectory predictions and to manage initialization and termination of new tracks; nonetheless, many of the presented steps can often still be identified even in such cases, as we will see.

\subsection{Metrics}\label{sec:metrics}
In order to provide a common experimental setup where algorithms can be fairly tested and compared, a group of metrics have been \textit{de facto} established as standard, and they are used in almost every work. The most relevant ones are metrics defined by Wu and Nevatia \cite{wu2006tracking}, the so-called CLEAR MOT metrics \cite{bernardin2008evaluating}, and recently the ID metrics \cite{ristani2016performance}. These sets of metrics aim to reflect the overall performance of the tested models, and point out the possible drawbacks of each one. Therefore, those metrics are defined as follows:

\subsubsection*{Classical metrics}

These metrics, defined by Wu and Nevatia \cite{wu2006tracking}, highlight the different types of errors a MOT algorithm can make. In order to show those problems, the following values are computed:

\begin{itemize}
    \item \textit{Mostly Tracked} (MT) trajectories: number of ground-truth trajectories that are correctly tracked in at least 80\% of the frames.
    \item \textit{Fragments}: trajectory hypotheses which cover at most 80\% of a ground truth trajectory. Observe that a true trajectory can be covered by more than one fragment.
    \item \textit{Mostly Lost} (ML) trajectories: number of ground-truth trajectories that are correctly tracked in less than 20\% of the frames.
    \item \textit{False trajectories}: predicted trajectories which do not correspond to a real object (i.e. to a ground truth trajectory).
    \item \textit{ID switches}: number of times when the object is correctly tracked, but the associated ID for the object is mistakenly changed.
\end{itemize}

\subsubsection*{CLEAR MOT metrics}
The CLEAR MOT metrics were developed for the \textit{Classification of Events, Activities and Relationships} (CLEAR) workshops held in 2006 \cite{stiefelhagen2007multimodal} and 2007 \cite{stiefelhagen2008multimodal}. The workshops were jointly organized by the the European CHIL project, the U.S. VACE project, and the National Institute of Standards and Technology (NIST). Those metrics are MOTA (Multiple Object Tracking Accuracy) and MOTP (Multiple Object Tracking Precision). They serve as a summary of other simpler metrics which compose them. We will explain the simpler metrics at first and build the complex ones over them. A detailed description on how to match the real objects (ground truth) with the tracker hypothesis can be found in \cite{bernardin2008evaluating}, as it is not trivial how to consider when a hypothesis is related to an object, and it depends on the precise tracking task to be evaluated. In our case, as we are focusing on 2D tracking with single camera, the most used metric to decide whether an object and a prediction are related or not is Intersection over Union (IoU) of bounding boxes, as it was the measure established in the presentation paper of MOT15 dataset \cite{leal2015motchallenge}. Specifically, the mapping between ground truth and hypotheses is established as follows: if  the ground truth object $o_i$ and the hypothesis $h_j$ are matched in frame $t-1$, and in frame $t$ the $IoU(o_i, h_j) \geq 0.5$, then $o_i$ and $h_j$ are matched in that frame, even if there exists another hypothesis $h_k$ such that $IoU(o_i, h_j) < IoU(o_i, h_k)$, considering the continuity constraint. After the matching from previous frames has been performed, the remaining objects are tried to be matched with the remaining hypotheses, still using a 0.5 IoU threshold. The ground truth bounding boxes that cannot be associated with a hypothesis are counted as \textit{false negatives} (FN), and the hypotheses that cannot be associated with a real bounding box are marked as \textit{false positives} (FP). Also, every time a ground truth object tracking is interrupted and later resumed is counted as a \textit{fragmentation}, while every time a tracked ground truth object ID is incorrectly changed during the tracking duration is counted as an \textit{ID switch}. Then, the simple metrics computed are the following:

\begin{itemize}
    \item FP: the number of false positives in the whole video;
    \item FN: the number of false negatives in the whole video;
    \item Fragm: the total number of fragmentations;
    \item IDSW: the total number of ID switches.
\end{itemize}

The MOTA score is then defined as follows:

\[
\mathit{MOTA} = 1 - \frac{(\mathit{FN} + \mathit{FP} + \mathit{IDSW})}{\mathit{GT}} \quad \in (-\infty , 1 ]
\]

where $GT$ is the number of ground truth boxes. It is important to note that the score can be negative, as the algorithm can commit a number of errors greater than the number of ground truth boxes. Usually, instead of reporting MOTA, it is common to report the percentage MOTA, which is just the previous expression expressed as a percentage. On the other hand, MOTP is computed as: 
\[
\mathit{MOTP} = \frac{\sum_{t,i} d_{t,i}}{\sum_t c_t}
\]

where $c_t$ denotes the number of matches in frame $t$, and $d_{t,i}$ is the bounding box overlap between the hypothesis $i$ with its assigned ground truth object. It is important to note that this metric takes few information about tracking into account, and rather focuses on the quality of the detections.

\subsubsection*{ID scores}
The main problem of MOTA score is that it takes into account the number of times a tracker makes an incorrect decision, such as an ID switch, but in some scenarios (e.g. airport security) one could be more interested in rewarding a tracker that can follow an object for the longest time possible, in order to not lose its position. Because of that, in \cite{ristani2016performance} a couple of alternative new metrics are defined, that are supposed to complement the information given by the CLEAR MOT metrics. Instead of matching ground truth and detections frame by frame, the mapping is performed globally, and the trajectory hypothesis assigned to a given ground truth trajectory is the one that maximizes the number of frames correctly classified for the ground truth. In order to solve that problem, a bipartite graph is constructed, and the minimum cost solution for that problem is taken as the problem solution. For the bipartite graph, the sets of vertices are defined as follows: the first set of vertices, $V_T$, has a so-called regular node for each true trajectory, and a false positive node for each computed trajectory. The second set, $V_C$, has a regular node for each computed trajectory and a false negative for each true one. The costs of the edges are set in order to count the number of false negative and false positive frames in case that edge were chosen (more information can be found in \cite{ristani2016performance}). After the association is performed, there are four different possible pairs, attending to the nature of the involved nodes. If a regular node from $V_T$ is matched with a regular node of $V_C$ (i.e. a true trajectory is matched with a computed trajectory), a \textit{true positive ID} is counted. Every false positive from $V_T$ matched with a regular node from $V_C$ counts as a \textit{false positive ID}. Every regular node from $V_T$ matched with a false negative from $V_C$ counts as a \textit{false negative ID}, and finally, every false positive matched with a false negative counts as a \textit{true negative ID}. Afterwards, three scores are calculated. IDTP is the sum of the weights of the edges selected as \textit{true positive ID} matches (it can be seen as the percentage of detections correctly assigned in the whole video). IDFN is the sum of weights from the selected \textit{false negative ID} edges, and IDFP is the sum of weights from the selected \textit{false positive ID} edges. With these three basic measures, another three measures are computed:

\begin{itemize}
    \item Identification precision: $\mathit{IDP} = \frac{\mathit{IDTP}}{\mathit{IDTP} + \mathit{IDFP}}$
    \item Identification recall: $\mathit{IDR} = \frac{\mathit{IDTP}}{\mathit{IDTP} + \mathit{IDFN}}$
    \item Identification F1: $\mathit{IDF1} = \frac{2}{\frac{1}{\mathit{IDP}} + \frac{1}{\mathit{IDR}}} = \frac{2\mathit{IDTP}}{2\mathit{IDTP} + \mathit{IDFP} + \mathit{IDFN}}$
\end{itemize}

Usually, the reported metrics in almost every piece of work are the CLEAR MOT metrics, mostly tracked trajectories (MT), mostly lost trajectories (ML) and IDF1, since this metrics are the ones shown in MOTChallenge leaderboards (see section \ref{sec:datasets} for details). Additionally, the number of frames per second (FPS) the tracker can process is often reported, and is also included in the leaderboards. However, we find this metric difficult to compare among different algorithms, since some of the methods include the detection phase while others skip that computation. Also, the dependency on the hardware employed is relevant in terms of speed.

\subsection{Benchmark datasets}\label{sec:datasets}
In the past few years, a number of datasets for MOT have been published. In this section we are going to describe the most important ones, starting from a general description of the MOTChallenge benchmark, then focusing on its datasets, and finally describing KITTI and other less commonly used MOT datasets.

{\bf MOTChallenge.} MOTChallenge\footnote{\url{https://motchallenge.net/}} is the most commonly used benchmark for multiple object tracking. It provides, among others, some of largest datasets for pedestrian tracking that are currently publicly available. For each dataset, the ground truth for the training split, and detections for both training and test splits are provided. The reason why MOTChallenge datasets frequently provide detections (often referred to as \textit{public detections}, as opposed to the \textit{private detections}, that are obtained by the algorithm authors by using a detector of their own) is that the detection quality has a big impact on the final performance of the tracker, but the detection part of the algorithms is often independent from the tracking part and usually uses already existing models; providing public detections that every model can use makes the comparison of the tracking algorithms easier, since the detection quality is factored out from the performance computation and trackers start on a common ground. The evaluation of an algorithm on the test dataset is done by submitting the results to a test server. The MOTChallenge website contains a leaderboard for each of the datasets, showing in separate pages models using the publicly provided detections and the ones using private detections. Online methods are also marked as so. MOTA is the primary evaluation score for the MOTChallenge, but many other metrics are shown, including all the ones presented in section \ref{sec:metrics}. As we will see, since the vast majority of MOT algorithms that use deep learning focus on pedestrians, the MOTChallenge datasets are the most widely used, as they are the most comprehensive ones currently available, providing more data to train deep models.

{\bf MOT15.} The first MOTChallenge dataset is 2D MOT 2015\footnote{Dataset: \url{https://motchallenge.net/data/2D_MOT_2015/}, leaderboard: \url{https://motchallenge.net/results/2D_MOT_2015/}.} \cite{leal2015motchallenge} (often just called MOT15). It contains a series of 22 videos (11 for training and 11 for testing), collected from older datasets, with a variety of characteristics (fixed and moving cameras, different environments and lighting conditions, and so on) so that the models would need to generalize better in order to obtain good results on it. In total, it contains 11283 frames at various resolutions, with 1221 different identities and 101345 boxes. The provided detections were obtained using the ACF detector \cite{dollar2014fast}.

{\bf MOT16/17.} A new version of the dataset was presented in 2016, called MOT16\footnote{Dataset: \url{https://motchallenge.net/data/MOT16/}, leaderboard: \url{https://motchallenge.net/results/MOT16/}.} \cite{milan2016mot16}. This time, the ground truth was made from scratch, so that it was consistent throughout the dataset. The videos are also more challenging, since they have a higher pedestrian density. A total of 14 videos are included in the set (7 for training and 7 for testing), with public detections obtained using the Deformable Part-based Model (DPM) v5 \cite{felzenszwalb2009object, girshick2012DPMv5}, that they found to obtain better performance in detecting pedestrians on the dataset when compared to other models. This time the dataset includes 11235 frames with 1342 identities and 292733 boxes in total. The MOT17 dataset\footnote{Dataset: \url{https://motchallenge.net/data/MOT17/}, leaderboard: \url{https://motchallenge.net/results/MOT17/}.} includes the same videos as MOT16, but with more accurate ground truth and with three sets of detections for each video: one from Faster R-CNN \cite{ren2015faster}, one from DPM and one from the Scale-Dependent Pooling detector (SDP) \cite{yang2016exploit}. The trackers would then have to prove to be versatile and robust enough to get a good performance using different detection qualities.

{\bf MOT19.} Very recently, a new version of the dataset for the CVPR 2019 Tracking Challenge\footnote{\url{https://motchallenge.net/workshops/bmtt2019/tracking.html}} has been released, containing 8 videos (4 for training, 4 for testing) with extremely high pedestrian density, reaching up to 245 pedestrians per frame on average in the most crowded video. The dataset contains 13410 frames with 6869 tracks and a total of 2259143 boxes, much more than the previous datasets. While submissions for this dataset have only been allowed for a limited amount of time, this data will be the basis for the release of MOT19 in late 2019 \cite{dendorfer2019cvpr19}.

{\bf KITTI.} While the MOTChallenge datasets focus on pedestrian tracking, the KITTI tracking benchmark\footnote{\url{http://www.cvlibs.net/datasets/kitti/eval_tracking.php}} \cite{geiger2012we, geiger2013vision} allows for tracking of both people and vehicles. The dataset was collected by driving a car around a city and it was released in 2012. It consists of 21 training videos and 29 test ones, with a total of about 19000 frames (32 minutes). It includes detections obtained using the DPM\footnote{The website says the detections were obtained using a model based on a latent SVM, or L-SVM. That model is now known as Deformable Parts Model (DPM).} and RegionLets\footnote{\url{http://www.xiaoyumu.com/project/detection}} \cite{wang2013regionlets} detectors, as well as stereo and laser information; however, as explained, in this survey we are only going to focus on models using 2D images. The CLEAR MOT metrics, MT, ML, ID switches and fragmentations are used to evaluate the methods. It is possible to submit results only for pedestrians or only for cars, and two different leaderboards are maintained for the two classes.

{\bf Other datasets.} Besides the previously described datasets, there is a number of older, and now less frequently used, ones. Among those we can find the UA-DETRAC tracking benchmark\footnote{\url{https://detrac-db.rit.albany.edu/Tracking}} \cite{wen2015ua}, that focuses on vehicles tracked from traffic cameras, and the TUD\footnote{\url{https://www.d2.mpi-inf.mpg.de/node/428}} \cite{andriluka2010monocular} and PETS2009\footnote{\url{http://www.cvg.reading.ac.uk/PETS2009/a.html}} \cite{ferryman2009pets2009} datasets, that both focus on pedestrians. Many of their videos are now part of the MOTChallenge datasets.

\section{Deep learning in MOT} \label{sec:main}
As this survey focuses on the use of deep learning in the MOT task, we organize this section into five subsections. Each of the first four subsections provides a review on how deep learning is exploited in each one of the four MOT stages defined previously\footnote{Note that the classification of the models should not be considered as a strict categorization, since it's not rare that one of them has been used for multiple purposes and drawing a line is sometimes difficult. For example, some deep learning models, Siamese networks in particular, are often trained to output an affinity score, but at inference time they are only used to extract `association features', and a simple hardcoded distance measure is then used instead to compute the affinities. In those cases, we decided to consider the network as performing feature extraction, since the similarity measure is not directly learned. However, those models could have also been considered to use deep learning for affinity computation.}. Subsection \ref{sec:association}, besides presenting the use of deep learning in the association process, will also include its use in the overall track management process (e.g. initialization/termination of tracks), since it is strictly linked to the association step. Subsection \ref{sec:others} will finally describe uses of deep learning in MOT that do not fit into the four-step scheme.

We have included a summary table in \ref{app:summary} that shows the main techniques used in each of the four steps in each paper presented in this survey. The mode of operation (batch vs. online) is indicated and a link to the source code or to other provided material is also included (when available).

\subsection{DL in detection step}
\label{sec:detections}

While many works have used as input to their algorithms dataset-provided detections generated by various detectors (for example Aggregated Channel Features \cite{dollar2014fast} for MOT15 \cite{leal2015motchallenge} or Deformable Parts Model \cite{felzenszwalb2009object} for MOT16 \cite{milan2016mot16}), there have also been algorithms that integrated a custom detection step, that often contributed to improve the overall tracking performance by enhancing the detection quality.

As we will see, most of the algorithms that employed custom detections made use of Faster R-CNN and its variants (section \ref{sec:det-faster}) or SSD (section \ref{sec:det-ssd}), but approaches that used different models also exist (section \ref{sec:det-otherdet}). Despite the vast majority of algorithms utilized deep learning models to extract rectangular bounding boxes, a few works made a different use of deep networks in the detection step: these works are the focus of section \ref{sec:det-otheruses}.

\begin{figure}[htb]
    \centering
    \includegraphics[width=.7\textwidth]{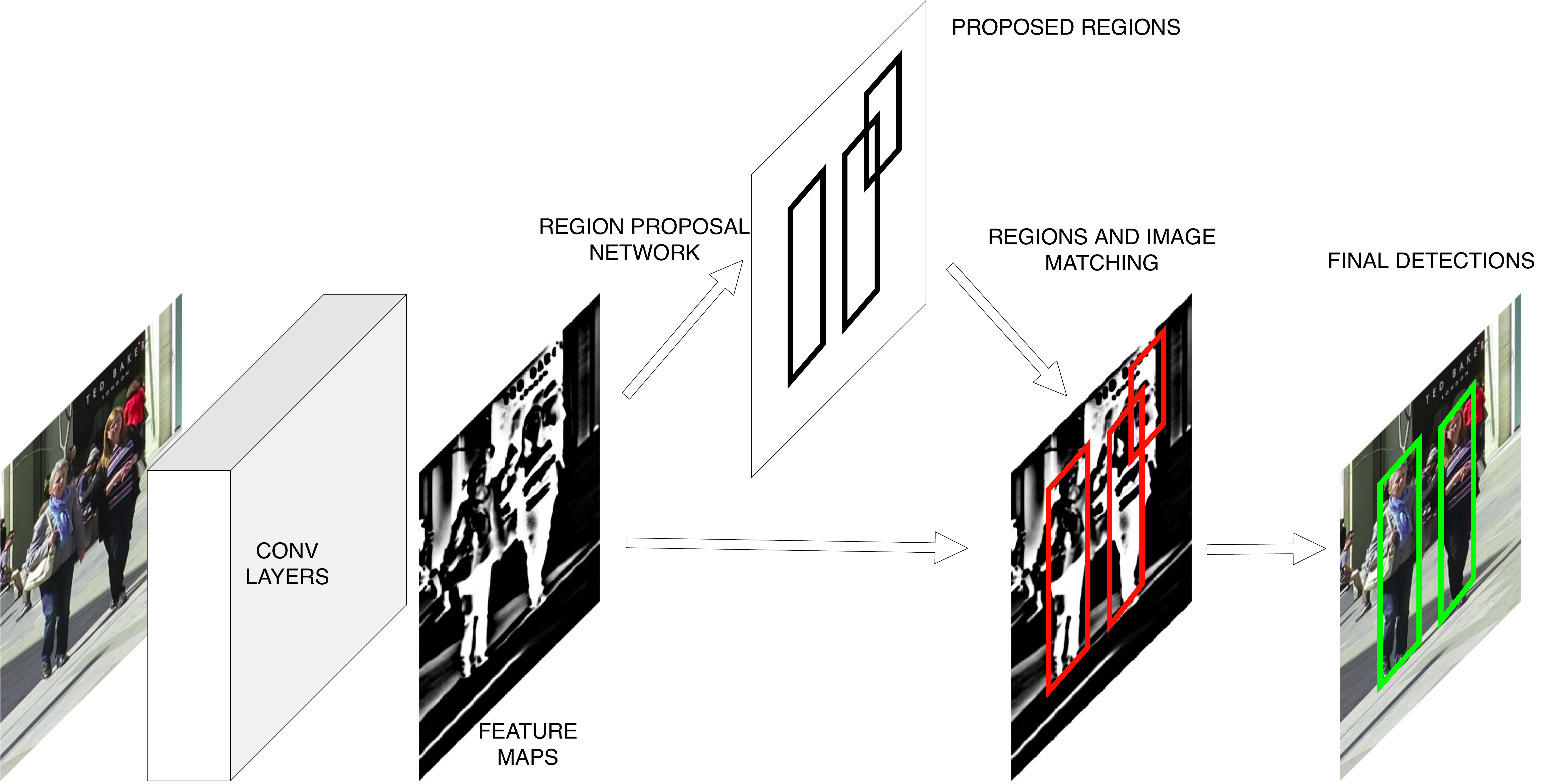}
    \caption{Example of a deep learning based detector (Faster R-CNN architecture \cite{ren2015faster})}
    \label{fig:detector}
\end{figure}

\subsubsection{Faster R-CNN}\label{sec:det-faster}

The Simple Online and Realtime Tracking (SORT) algorithm \cite{bewley2016simple} has been one of the first MOT pipelines to leverage convolutional neural networks for the detection of pedestrians. Bewley et al. showed that replacing detections obtained using Aggregated Channel Features (ACF) \cite{dollar2014fast} with detections computed by Faster R-CNN \cite{ren2015faster} (illustrated in figure \ref{fig:detector}) could improve the MOTA score by 18.9\% (absolute change) on the MOT15 dataset \cite{leal2015motchallenge}. They used a relatively simple approach that consisted in predicting object motion using the Kalman filter \cite{kalman1960new} and then associating the detections together with the help of the Hungarian algorithm \cite{kuhn1955hungarian}, using intersection-over-union (IoU) distances to compute the cost matrix. At the time of publishing, SORT was ranked as the best-performing open source algorithm on the MOT15 dataset.

Yu et al. reached the same conclusions in \cite{yu2016poi} using a modified Faster R-CNN, that included skip-pooling \cite{bell2016inside} and multi-region features \cite{gidaris2015object} and that was fine-tuned on multiple pedestrian detection datasets. With this architecture they were able to improve the performance of the algorithm they proposed (see section \ref{sec:featext-cnn}) by more than 30\% (absolute change, measured in MOTA), reaching state-of-the-art performance on the MOT16 dataset \cite{milan2016mot16}. They also showed that having higher-quality detections reduces the need of complex tracking algorithms while still obtaining similar results: this is because the MOTA score is heavily influenced by the amount of false positives and false negatives, and using accurate detections is an effective way of reducing both. The detections computed by \cite{yu2016poi} on the MOT16 dataset have also been made available to the public\footnote{\url{https://drive.google.com/file/d/0B5ACiy41McAHMjczS2p0dFg3emM/view}} and many MOT algorithms have since exploited them \cite{wojke2017simple, mahmoudi2019cnnmtt, wan2018online, ujiie2018interpolation, he2017sot, li2017multi, li2018did, jorquera2019probability, zhong2019decision, lu2019multi, tang2017multiple}.

In the following years, other works have taken advantage of the detection accuracy of Faster R-CNN, that has since been applied as part of MOT algorithms to detect athletes \cite{ran2019robust}, cells \cite{hu2018automatic} and pigs \cite{zhang2019automatic}. Moreover, an adaptation of Faster R-CNN that adds a segmentation branch, Mask R-CNN \cite{he2017mask}, has been used for example by Zhou et al. \cite{zhou2018online} both to detect and to track pedestrians,

\subsubsection{SSD}\label{sec:det-ssd}
The SSD \cite{liu2016ssd} detector is another commonly used network in the detection step. In particular, Zhang et al. \cite{zhang2019automatic} compared it with Faster R-CNN and R-FCN \cite{dai2016r} in their pig tracking pipeline, showing that it worked better on their dataset. They employed a Discriminative Correlation Filters (DCF) based online tracking method \cite{danelljan2017eco} with the use of HOG \cite{dalal2005histograms} and Colour Names \cite{van2009learning} features to predict the position of so-called \textit{tag-boxes}, small regions around the center of each animal. The Hungarian algorithm was used for the association between tracked tag-boxes and detections, and in the case of tracking failure the output of the DCF tracker was used to refine the bounding boxes. Lu et al. \cite{lu2017online} also used SSD, but in this case to detect a variety of object classes to track (people, animals, cars, etc., see section \ref{sec:featext-complex}).

Some works have tried to refine the detections obtained with SSD by taking into account the information obtained in other steps of the tracking algorithm. Kieritz et al. \cite{kieritz2018joint}, in their joint detection and tracking framework, used the affinity scores computed between tracks and detections to replace the standard Non-Maximum Suppression (NMS) step included in the SSD network with a version that refines detection confidence scores based on their correspondence to tracked targets.

Zhao et al. \cite{zhao2018multi} instead employed the SSD detector to search for pedestrians and vehicles in a scene, but they used a CNN-based Correlation Filter (CCF) to allow SSD to generate more accurate bounding boxes. The CCF exploited PCA-compressed \cite{pearson1901liii} CNN features to predict the position of a target in the subsequent frame; the predicted position was then used to crop a ROI (Region Of Interest) around it, that was given as input to SSD. In that way, the network was able to compute small detections using deeper layers, that extract more valuable semantic information and that are thus known to produce more accurate bounding boxes and less false negatives. The algorithm then combined these detections with the ones obtained on the full image with a NMS step and then association between tracks and detections was performed using the Hungarian algorithm, with a cost matrix that took into account geometric (IoU) and appearance (Average Peak-to-Correlation Energy - APCE \cite{wang2017large}) cues. APCE was also used for an object re-identification (ReID) step, to recover from occlusions. The authors showed that training a detector with multi-scale augmentation could lead to much better performance in tracking and the algorithm reached accuracy comparable to state-of-the-art online algorithms on KITTI and MOT15.

\subsubsection{Other detectors}\label{sec:det-otherdet}
Among the other CNN models used as detectors in MOT, we can mention the YOLO series of detectors \cite{redmon2016you, redmon2017yolo9000, redmon2018yolov3}; in particular, YOLOv2 has been used by Kim et al. \cite{kim2018online} also to detect pedestrians. Sharma et al. \cite{sharma2018beyond} used instead a Recurrent Rolling Convolution (RRC) CNN \cite{ren2017accurate} and a SubCNN \cite{xiang2017subcategory} to detect vehicles in videos recorded on a moving camera in the context of autonomous driving (see section \ref{sec:featext-complex}). Pernici et al. \cite{pernici2018memory} used the Tiny CNN detector \cite{hu2017finding} in their face tracking algorithm, obtaining a better performance when compared to the Deformable Parts Model detector (DPM) \cite{felzenszwalb2009object}, that does not use deep learning techniques.

\subsubsection{Other uses of CNNs in the detection step}\label{sec:det-otheruses}
Sometimes CNNs have been employed in the MOT detection step for uses other than directly computing object bounding boxes.

For example, CNNs have been exploited to reduce false positives in \cite{min2018new}, where vehicle detections were obtained with a modified version of the ViBe algorithm \cite{barnich2011vibe} that performed background subtraction on the input. These detections were first given as input to a SVM \cite{cortes1995support} and, in case the SVM was not confident enough to either discard or confirm them, a Faster-CNN based network \cite{yu2017model} would then be used to decide whether to keep or discard each of them. In this way, only a few objects would have to be analyzed by the CNN, making the detection step faster.

Bullinger et al. explored a different approach in \cite{bullinger2017instance}, where instead of computing classical bounding boxes in the detection step, a Multi-task Network Cascade \cite{dai2016instance} was instead employed to obtain instance-aware semantic segmentation maps. The authors argue that since the 2D shape of instances, differently from rectangular bounding boxes, do not contain background structures or parts of other objects, optical flow based tracking algorithms would perform better, especially when the target position in the image is also subject to camera motion in addition to the object's own motion. After obtaining the segmentation maps for the various instances present in the current frame, an optical flow method (\cite{farneback2003two, revaud2016deepmatching, hu2016efficient}) was applied to predict the position and shape of each instance in the next frame. An affinity matrix between predicted and detected instances was then computed and given as input to the Hungarian algorithm for association. While the method obtained slightly lower MOTA score on the whole MOT15 dataset when compared to SORT, the authors showed that it performed better on videos with moving camera.

\subsection{DL in feature extraction and motion prediction}
\label{sec:featext}

The feature extraction phase is the preferred one for the employment of deep learning models, due to their strong representational power that makes them good at extracting meaningful high-level features. The most typical approach in this area is the use of CNNs to extract visual features, as it is commented in section \ref{sec:featext-cnn}. Instead of using classical CNN models, another recurrent idea consists in training them as Siamese CNNs, using contrastive loss functions, in order to find the set of features that best distinguish between subjects. Those approaches are explained in section \ref{sec:featext-siamese}. Furthermore, some authors explored the capabilities of CNNs to predict object motion inside correlation filter based algorithms: these are commented in section \ref{sec:featext-cnnmotion}. Finally, other types of deep learning models have been employed, usually including them in more complex systems, combining deep features with classical ones. They are explained in sections \ref{sec:featext-complex} (specifically for visual features) and \ref{sec:featext-other} (for approaches that don't fit in the other categories).

\subsubsection{Autoencoders: first usage of DL in a MOT pipeline}
To the best of our knowledge, the first approach using deep learning in MOT was presented by Wang et al. \cite{wang2014learning} in 2014. They proposed a network of autoencoders stacked in two layers that were used to refine visual features extracted from natural scenes \cite{cadieu2009learning}. After the extraction step, affinity computation was performed using a SVM, and the association task was formulated as a minimum spanning tree problem. They showed that feature refinement greatly improved the model performance. However, the dataset on which the algorithm was tested is not commonly used and results are hardly comparable to other methods.

\subsubsection{CNNs as visual feature extractors} \label{sec:featext-cnn}

The most widely used methods for feature extraction are based on subtle modifications of convolutional neural networks. One of the first uses of these models can be found in \cite{kim2015multiple}. Here, Kim et al. incorporated visual features into a classical algorithm, called Multiple Hypothesis Tracking, using a pretrained CNN that extracted 4096 visual features from the detections, that were later reduced to 256 using PCA. This modification improved the MOTA score on MOT15 by more than 3 points. By the time that paper was submitted, it was the top ranked algorithm on that dataset. Yu el al. \cite{yu2016poi} used a modified version of GoogLeNet \cite{szegedy2015going}, pretrained on a custom re-identification dataset, built by combining classical person identification datasets (PRW \cite{zheng2017person}, Market-1501 \cite{zheng2015scalable}, VIPeR \cite{gray2008viewpoint}, CUHK03 \cite{li2014deepreid}). Visual features were combined with spatial ones, extracted with a Kalman filter, and then an affinity matrix was computed.

Other examples of the use of CNNs for feature extraction can be found in \cite{chen2017enhancing}, where a custom CNN was used to extract appearance features in a Multiple Hypothesis Tracking framework, in \cite{yang2017hybrid}, whose tracker employed a pretrained region-based CNN \cite{girshick2015region}, or in \cite{wang2017robust}, where a CNN extracted visual features from fish heads, later combined with motion prediction from a Kalman Filter.

\begin{figure}
    \centering
    \includegraphics[width=.9\textwidth]{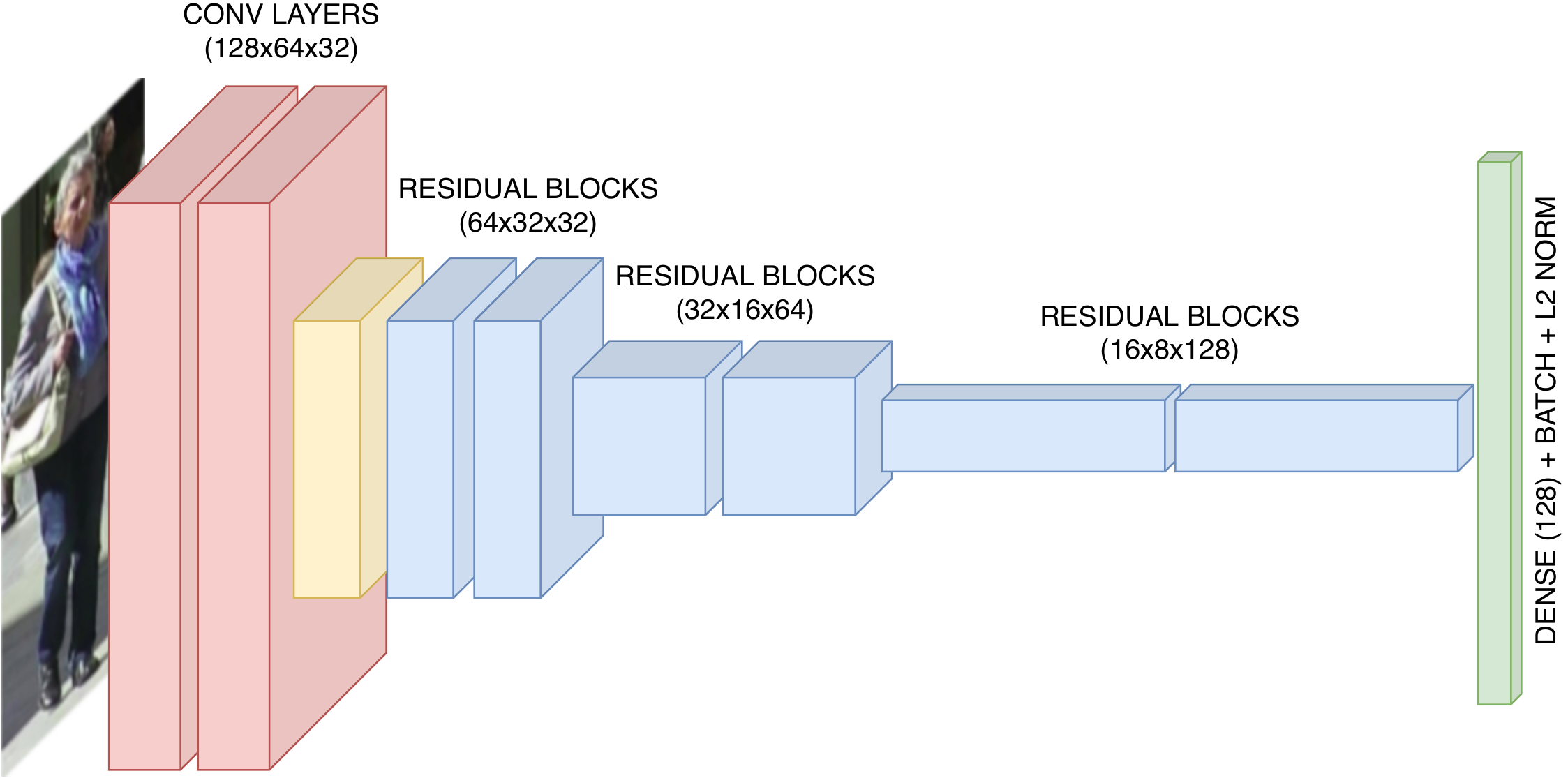}
    \caption{Diagram of DeepSORT \cite{wojke2017simple} CNN-based feature extractor. The red blocks are simple convolutional layers, the yellow block is a max pooling layer, and the blue blocks are residual blocks, that are composed of three convolutional layers each \cite{he2016deep}. The final green block represents a fully-connected layer with batch normalization and L2 normalization. The output size of each block is indicated in parentheses.}
    \label{fig:deepsort-cnn}
\end{figure}

The SORT algorithm \cite{bewley2016simple}, presented in section \ref{sec:det-faster}, was later refined with deep features, and this new version was called DeepSORT \cite{wojke2017simple}. This model incorporated visual information extracted by a custom residual CNN \cite{zagoruyko2016wide}. The CNN provided a normalized vector with 128 features as output, and the cosine distance between those vectors was added to the affinity scores used in SORT. A diagram of the network structure can be found in figure \ref{fig:deepsort-cnn}. The experiments showed that this modification overcame the main drawback of the SORT algorithm, which was a high number of ID switches.

Mahmoudi et al. \cite{mahmoudi2019cnnmtt} also incorporated CNN extracted visual features along with dynamic and position features, and then solved the association problem via Hungarian algorithm. In \cite{kim2018multi}, a ResNet-50 \cite{he2016deep} pretrained on ImageNet was used as visual feature extractor. An extensive explanation of how a CNN can be used to distinguish pedestrians can be found in \cite{bae2017confidence}. In their model, Bae et al. combined the output of the CNN with shape and motion models, and computed an aggregated affinity score for each pair of detections; the association problem was then solved by the Hungarian algorithm. Again, Ullah et al. \cite{ullah2018deep} applied an off-the-shelf version of GoogLeNet \cite{szegedy2015going} for feature extraction. Fang et al. \cite{fang2018recurrent} selected as visual features the output of a hidden convolutional layer of an Inception CNN \cite{xiao2016learning}. Fu et al. \cite{fu2018gm} employed the DeepSORT feature extractor, and measured the correlation of features using a discriminative correlation filter. Afterwards, the matching score was combined with a spatio-temporal relation score, and the final score was used as a likelihood in a Gaussian Mixture Probability Hypothesis Density filter \cite{vo2006gaussian}. The authors in \cite{wen2019learning} used a fine-tuned GoogLeNet on the ILSVRC CLS-LOC \cite{russakovsky2015imagenet} dataset for pedestrians recognition. In \cite{pernici2018memory}, the authors reused the visual features extracted by the CNN-based detector, and the association was performed using a Reverse Nearest Neighbor technique \cite{korn2000influence}. Sheng et al. \cite{sheng2018heterogeneous} employed the convolutional part of GoogLeNet to extract appearance features, using the cosine distance between them to compute an affinity score between pairs of detections, and merging that information with motion prediction in order to compute an overall affinity which serves as edge cost in a graph problem. Chen et al. \cite{chen2019recurrent} utilized the convolutional part of ResNet to build a custom model, stacking a LSTM cell on top of the convolutions, in order to compute simultaneously a similarity score and a bounding box regression.

In \cite{hu2018automatic}, the model learned to distinguish fast moving cells from slow moving cells. After the classification was computed, slow cells were associated using only motion features, since they were almost still, while fast cells were associated using both motion features and visual features extracted by a Fast R-CNN based on VGG-16 \cite{simonyan2014very}, specifically fine-tuned for the cell classification task. Moreover, the proposed model included a tracking optimization step, where false negatives and false positives were reduced by combining possible tracklets that were mistakenly interrupted. 

Ran et al. \cite{ran2019robust} proposed a combination of a classical CNN for visual features extraction and AlphaPose CNN for pose estimation. The output of these two networks was then fed into a LSTM model together with the tracklet information history to compute a similarity, as it is explained in section \ref{sec:aff-rnn}.

An interesting employment of CNNs in feature extraction can be found in \cite{tang2017multiple}. The authors used a pose detector, called DeepCut \cite{pishchulin2016deepcut}, that was a modification of Fast R-CNN; its output consisted in score maps predicting the presence of fourteen body parts. These were combined with the cropped images of detected pedestrians and fed into a CNN. A more detailed explanation of the algorithm is available in section \ref{sec:aff-cnn}. 

\subsubsection{Siamese networks} \label{sec:featext-siamese}

\begin{figure}[bth]
    \centering
    \includegraphics[width=.4\textwidth]{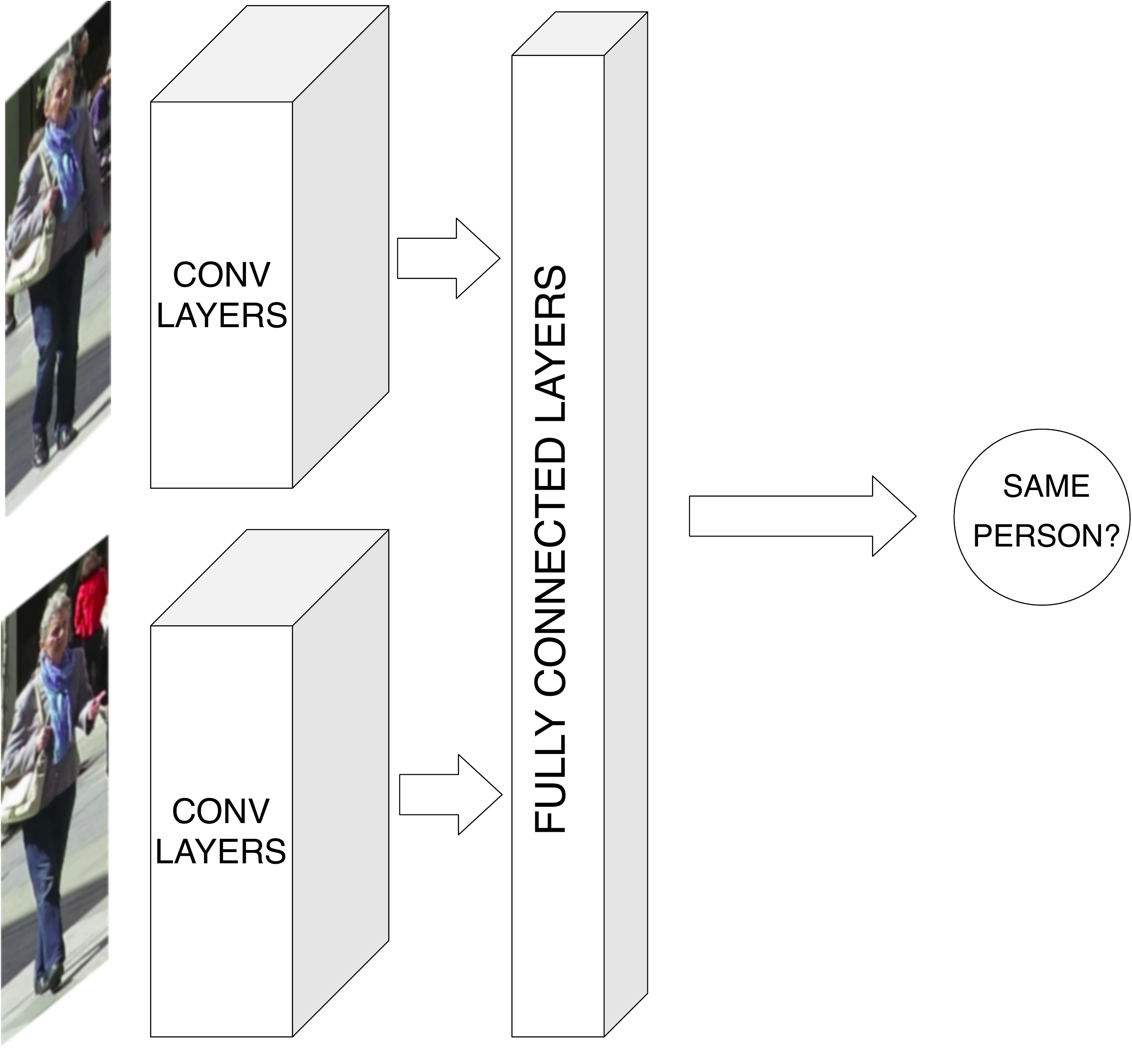}
    \caption{Example of a Siamese CNN architecture. For feature extraction, the network is trained as a Siamese CNN, but at inference time the output probability is discarded, and the last fully connected layer is used as feature vector for a single candidate. When the network is used for affinity computation, the whole structure is preserved during inference.}
    \label{fig:siamese-cnn}
\end{figure}

Another recurrent idea is training CNNs with loss functions that combine information from different images, in order to learn the set of features that best differentiates examples of different objects. These networks are usually called Siamese networks (an example of the architecture is shown in figure \ref{fig:siamese-cnn}). Kim et al. \cite{kim2016similarity} proposed a Siamese network \cite{bromley1994signature} which was trained using a contrastive loss. The network took two images, their IoU score and their area ratio as input, and produced a contrastive loss as output. After the net was trained, the layer that computed the contrastive loss was removed, and the last layer was used as a feature vector for the input image. The similarity score was later computed by combining the Euclidean distance between feature vectors, the IoU score and the area ratio between bounding boxes. The association step was solved using a custom greedy algorithm. Wang et al. \cite{wang2016joint} also proposed a Siamese network which took two image patches and computed a similarity score between them. The score at test time was computed comparing the visual features extracted by the network for the two images, and including temporally constrained information. The distance employed as similarity score was a Mahalanobis distance with a weight matrix, also learned by the model.

Zhang et al. \cite{zhang2016tracking} proposed a loss function called SymTriplet loss. According to their explanation, during the training phase three CNNs with shared weights were used, and the loss function combined the information extracted from two images belonging to the same object (positive pair) and from an image of a different one (two negative pairs). The SymTriplet loss decreased when the distance between the feature vectors of the positive pair was small, and increased when the negative pairs' features were close. Optimizing that function resulted in very similar feature vectors for images of the same object, while producing different vectors for different objects, with a larger distance between them. The dataset on which the tracking algorithm was tested was made of chapters from TV series and music videos from YouTube. Since the videos included different shots, the problem was divided into two stages. First, data association between frames in the same shot were performed. The affinity score in that case was a combination between the Euclidean distance of the feature vectors from the detections, temporal and kinematic information. Afterwards, tracklets were linked across shots, using a Hierarchical Agglomerative Clustering algorithm working over the appearance features.

Leal-Taixé et al. \cite{leal2016learning} proposed a Siamese CNN which received two stacked images as an input, and output the probability of both images belonging to the same person. They trained the network with this output so that it learned the most representative features to distinguish subjects. Afterwards, the output layer was removed and the features extracted by the last hidden layer were used as input for a Gradient Boosting model, together with contextual information, in order to get an affinity score between detections. Then, the association step was solved using Linear Programming \cite{leal2011everybody}.

Son et al. \cite{son2017multi} proposed a new CNN architecture, called Quad-CNN. This model received as input four image patches, where the first three of them were from the same person, but in increasing time order, and the last one from another person. The network was trained using a custom loss, combining information about temporal distances between detections, extracted visual features, and bounding box positions. During the test phase, the network took two detections, and predicted the probability that both detections belonged to the same person, using the learned embedding.

In \cite{zhou2018online} a Siamese network based on Mask R-CNN \cite{he2017mask} was built. After the Mask R-CNN had produced the mask for each detection, three examples were fed into the shallow Siamese net, two from the same object (positive pair) and one from another object (negative pair), again, and a triplet loss was used for training. After the training phase, the output layer was removed, and a 128-d vector was extracted from the last hidden layer. The appearance similarity was then computed using the cosine distance. That similarity was further combined with a motion consistency, which consisted on a score based on the predicted position of the object, assuming linear motion, and with a spatial potential, which was a more complex motion model. The association problem was then solved with a power iteration over a 3-d tensor of computed similarities.

Maksai et al. \cite{maksai2018eliminating} directly used the 128-d feature vector extracted by the ReID triplet CNN proposed in \cite{hermans2017defense}, and combined it with other appearance-based features (as an alternative to an appearance-less version of the algorithm). Those features were further processed by a bidirectional LSTM. In \cite{zhu2018online} a similar approach was followed, with a so-called Spatial Attention Network (SAN). The SAN was a Siamese CNN, which used a pretrained ResNet-50 as base model. That net was truncated so that only the convolutional layers were employed. Then, a Spatial Attention Map was extracted from the last convolutional layers of the model: it represented a measure of the importance of different parts in the bounding box, in order to exclude background and other targets from the extracted features. The features were in fact weighted by this map, acting as a mask. The masked features from both detections were then merged into a fully connected layer which computed the similarity between them. During training, the network was also set to output a classification score, because the authors observed that jointly optimizing classification and affinity computation tasks resulted in a better performance in the latter. The affinity information was further fed into a bidirectional LSTM, as in the previous example. Both will be further discussed in section \ref{sec:affinity-comp}. Ma et al. \cite{ma2018trajectory} also trained a Siamese CNN in order to extract visual features from tracked pedestrians in their model, which is explained in detail in section \ref{sec:ass-rnn}.

In \cite{zhou2018deep}, Zhou et al. proposed a visual displacement CNN, which learned to predict the next position of an object depending on previous positions of the objects, and the influence that an object had over other objects in the scene. That CNN was then used to predict the location of objects in the next frame, taking as input their past trajectories. The network was also capable of extracting visual information from the predicted positions and the actual detections, in order to compute a similarity score, as it is explained in section \ref{sec:aff-cnn}.

Chen et al. \cite{long2018tracking} proposed a two-steps algorithm which employed GoogLeNet trained with triplet loss for feature extraction. In the first step, the model used a R-FCN to predict possible detection candidates using information from the existing tracklets. Then, those detections were combined with the actual detections and NMS was performed. Afterwards, using the customly trained GoogLeNet model, they extracted visual features from the detections, and solved the association problem with a hierarchical association algorithm. When their paper was published, the algorithm was ranked on top among online methods in the MOT16 dataset.

Lee et al. \cite{lee2019multiple} recently explored an interesting approach, combining pyramid and Siamese networks together. Their model, called Feature Pyramid Siamese Network, employed a backbone network (they studied the performance using SqueezeNet \cite{iandola2016squeezenet} and GoogLeNet \cite{szegedy2015going}, but the backbone network can be changed), which extracted visual features from two different images using the same parameters. Afterwards, some of the hidden feature maps from the network were extracted and given to the Feature Pyramid Siamese Network. The network then employed an upsampling and merging strategy to create a feature vector for every stage of the pyramid. Deeper layers were merged with shallower ones in order to enrich the simpler features with more complex ones. Afterwards, affinity score computation was performed, as explained in section \ref{sec:aff-siamesecnn}.

\subsubsection{More complex approaches for visual feature extraction}
\label{sec:featext-complex}

\begin{figure}[thb]
    \centering
    \includegraphics[width=.8\textwidth]{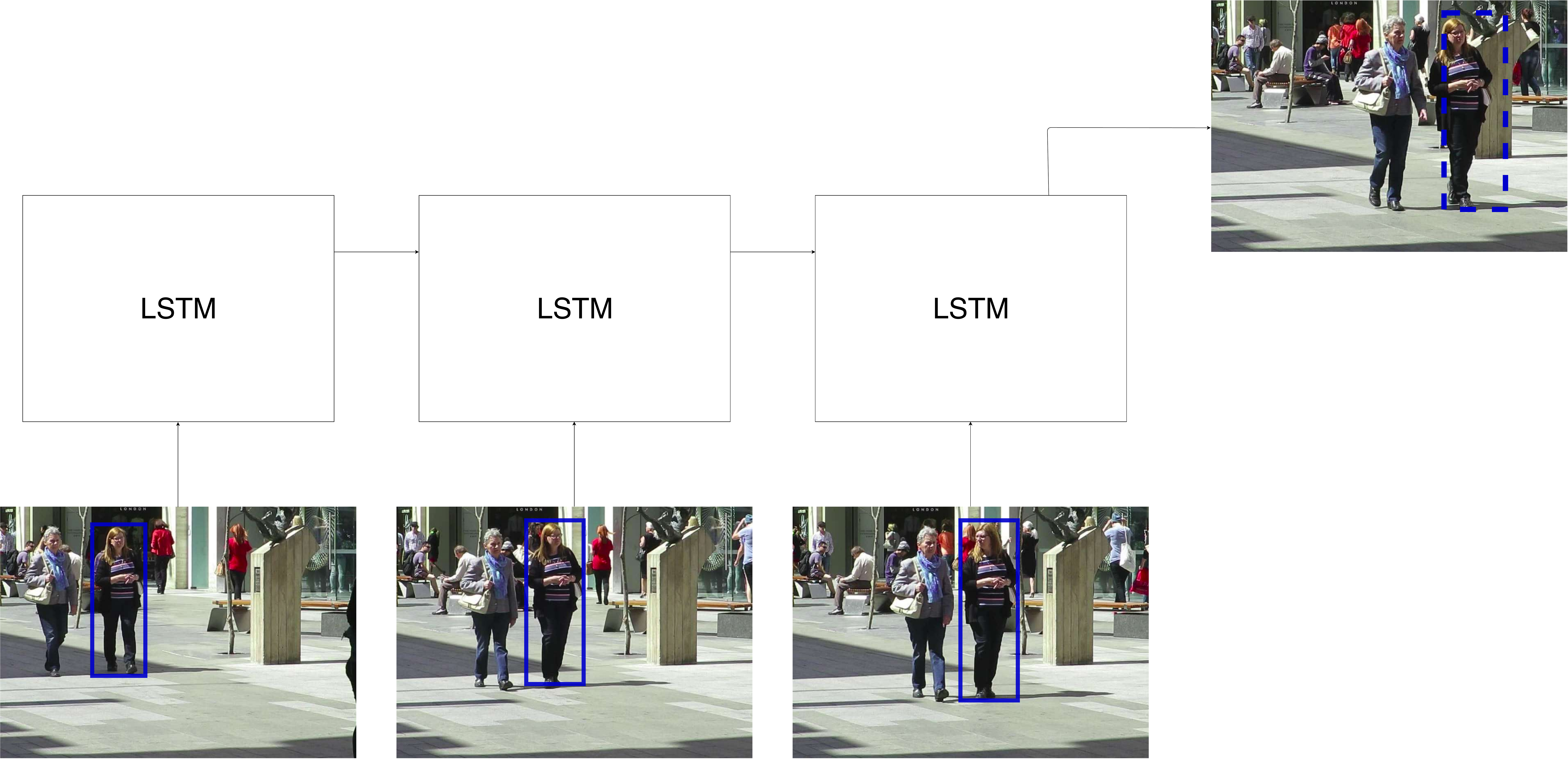}
    \caption{Typical usage of LSTM for motion prediction. A group of bounding boxes are fed into the network, and the produced output is the predicted bounding box in the next frame}
    \label{fig:motion_pred}
\end{figure}
 
 More complex approaches have also been proposed. Lu et al. \cite{lu2017online} employed the class predicted by the SSD in the detection step as a feature, and combined it with an image descriptor extracted with RoI pooling for each detection. Afterwards, the extracted features were used as input for a LSTM network, which learned to compute association features for the detections. Those features were later used for affinity computation, using the cosine distance between them.
 
 In \cite{ullah2017hierarchical}, the shallower layers of GoogLeNet were employed to learn a dictionary of features of the tracked objects. In order to learn the dictionary, the algorithm randomly selected objects in the first 100 frames of the video. The model extracted the feature maps in the first seven layers of the network. Then a dimensionality reduction was performed using Orthogonal Matching Pursuit (OPM) \cite{mallat1993matching} over the features extracted from the objects, and the learned representation was used as a dictionary. During the test phase, the OPM representation was computed for every detected object in the scene, and compared with the dictionary in order to construct a cost matrix, combining visual and motion information extracted by a Kalman filter. Finally, the association was performed using the Hungarian algorithm.
 
 LSTMs are sometimes employed for motion prediction, in order to learn more complex, non-linear motion models from the data. In figure \ref{fig:motion_pred} a scheme of the typical use of LSTMs for motion prediction is shown. An example of this use of recurrent networks is shown by Sadeghian et al. \cite{sadeghian2017tracking}, who proposed a model that employed three different RNNs to compute various types of features, not only motion ones, for each detection. The first RNN was employed to extract appearance features. The input of this RNN was a visual features vector extracted by a VGG CNN \cite{simonyan2014very}, pretrained specifically for person re-identification. The second RNN was a LSTM trained to predict the motion model for every tracked object. In this case, the output of the LSTM was the velocity vector of each object. The last RNN was trained to learn the interactions between different objects on the scene, since the position of some objects could be influenced by the behavior of surrounding items. The affinity computation was then performed by another LSTM, taking the information of the other RNNs as input.
 
 In \cite{chu2017online}, a model of stacked CNNs was proposed. The first section of the model consisted of a pretrained shared CNN which extracted common features for every object in the scene. That CNN was not updated online. Then, a RoI pooling was applied and the RoI features for every candidate were extracted. Afterwards, for every tracked candidate a new specific CNN was instantiated and trained online. Those CNNs extracted both the visibility map and the spatial attention map for its candidate. Finally, after the refined features were extracted, the probability of each new image belonging to every already tracked object was computed, and the association step was finally performed using a greedy algorithm.
 
 Sharma et al. \cite{sharma2018beyond} designed a set of cost functions to compute similarity between detections of vehicles. Those costs combined appearance features, extracted by a CNN, with 3D shape and position features assuming an environment with a moving camera. The defined costs were a 3D-2D cost, were the estimated 3D projection of the bounding box on the previous frame was compared with the 2D bounding box on the new frame, a 3D-3D cost, were the 3D projection of the previous bounding box was overlapped with the 3D projection of the current bounding box, an appearance cost, were the euclidean distance of the extracted visual features was computed, and a shape and pose cost, were the rough shape and position of the object in the bounding boxes were computed and compared. Note that while 3D projections were inferred, the input was still 2D images. After every cost was computed, the final pairwise cost between detections in two subsequent frames was a linear combination of the former costs. The final association problem was solved using the Hungarian algorithm.

Kim et al. \cite{kim2018online} employed the information extracted by the YOLOv2 CNN object detector to build a random ferns classifier \cite{ozuysal2009fast}. The algorithm worked in two steps. In the first step, a so-called teacher-RF was trained in order to differentiate pedestrians from non-pedestrians. After the teacher-RF was trained, for every tracked object, a random ferns classifier was constructed. Those classifiers were called student-RF, and they were smaller than the teacher-RF. They were specialized in distinguishing their tracked object from the rest of the objects in the scene. The decision of having a small random ferns classifier for every object was taken in order to reduce the computational complexity of the overall model, so that it could work in real time.

In \cite{ullah2018directed} the number of affinity computations that the model must compute was reduced by estimating first the position of objects in subsequent frames, using a Hidden Markov Model \cite{rabiner1986introduction}. Then, the feature extraction was performed using a pretrained CNN. After the visual features were extracted, the affinity computation was only computed between feasible pairs, that is, between detections close enough to the HMM prediction to be considered as the same object. The affinity score was obtained using a mutual information function between visual features. When the affinity scores were computed, a dynamic programming algorithm was used to associate detections.

\subsubsection{CNNs for motion prediction: correlation filters} \label{sec:featext-cnnmotion}

Wang et al. \cite{wang2017online} studied the employment of a correlation filter \cite{ma2015hierarchical}, whose output is a response map for the tracked object. That map was an estimation of the new position of the object in the next frame. Such affinity was further combined with an optical flow affinity, computed using the Lucas-Kanade algorithm \cite{lucas1981iterative}, a motion affinity calculated with a Kalman filter, and a scale affinity, represented by a ratio involving height and width of the bounding boxes. The affinity between two detections was computed as a linear combination of the previous scores. There was also a step of false detections removal, using a SVM classifier, and missing detections handling, using for that task the response map calculated in the previous steps. If an object was mistakenly lost and then re-identified, that step could fix the mistake and reconnect the broken tracklet.

In \cite{zhao2018multi}, a correlation filter was also employed to predict the position of the object in subsequent frames. The filter received as input the appearance features extracted by a CNN, previously reduced using PCA, and produced a response map of the predicted position for the object in the next frame as output. The predicted position was later used to compute a similarity score, combining the IoU between prediction and detections, and the APCE score of the response map. After the cost matrix was constructed, computing said score for every pair of detections between frames, the assignment problem was solved using the Hungarian algorithm.

\subsubsection{Other approaches}  \label{sec:featext-other}
Rosello et al. \cite{rosello2018multi} explored a completely different approach, using a reinforcement learning framework to train a set of agents that helped in the feature extraction step. The algorithm was based solely on motion features, without any visual information employed. The motion model was learned using a Kalman filter, whose behavior was managed by an agent, using one agent for each tracked object. The agent learned to decide which action should the Kalman filter take, between a set of actions that included ignoring the prediction, ignoring the new measure, using both information pieces, starting or stopping a track. The authors claimed that their algorithm could solve the tracking task even in non-visual scenarios, in contrast with classical algorithms whose performance was deeply influenced by visual features. However, the experimental results on MOT15 are not reliable and cannot be compared with other models because the model was tested on the training set.

Another algorithm that relied solely on motion features was the one proposed in \cite{babaee2018occlusion}. Babaee et al. presented a LSTM which learned to predict the new position and size of the bounding box for every object in the scene, using information about position and velocity in previous frames. Using the IoU between the predicted bounding box and the real detection, an affinity measure was computed, and the tracks were associated using a custom greedy algorithm. The pipeline was applied on tracking results obtained by other algorithms, in order to handle occlusions, and the authors showed that their method could effectively reduce the number of ID switches.

\subsection{DL in affinity computation}
\label{sec:affinity-comp}
While many works compute affinity between tracklets and detections (or tracklets and other tracklets) by using some distance measure over features extracted by a CNN, there are also algorithms that use deep learning models to directly output an affinity score, without having to specify an explicit distance metric between the features. This section focuses on such works.

In particular, we are first going to describe algorithms that used recurrent neural networks, starting from standard LSTMs (section \ref{sec:aff-rnn}) and then describing uses of Siamese LSTMs (section \ref{sec:aff-siamlstm}) and Bidirectional LSTMs (section \ref{sec:aff-bilstm}). A particular use of LSTM-computed affinities in the context of Multiple Hypothesis Tracking (MHT) frameworks is presented in section \ref{sec:aff-lstmmht}; finally, a few works that employed different kinds of recurrent network for affinity computations are presented in section \ref{sec:aff-otherrnn}.

In the second part of this section we are going to explore instead the uses of CNNs in affinity computation (section \ref{sec:aff-cnn}), including the algorithms that used the output of Siamese CNNs directly as an affinity score (section \ref{sec:aff-siamesecnn}), instead of relying on distance measures over feature vectors like in section \ref{sec:featext-siamese}.

\subsubsection{Recurrent neural networks and LSTMs}\label{sec:aff-rnn}
\begin{figure}
    \centering
    \includegraphics[width=.9\textwidth]{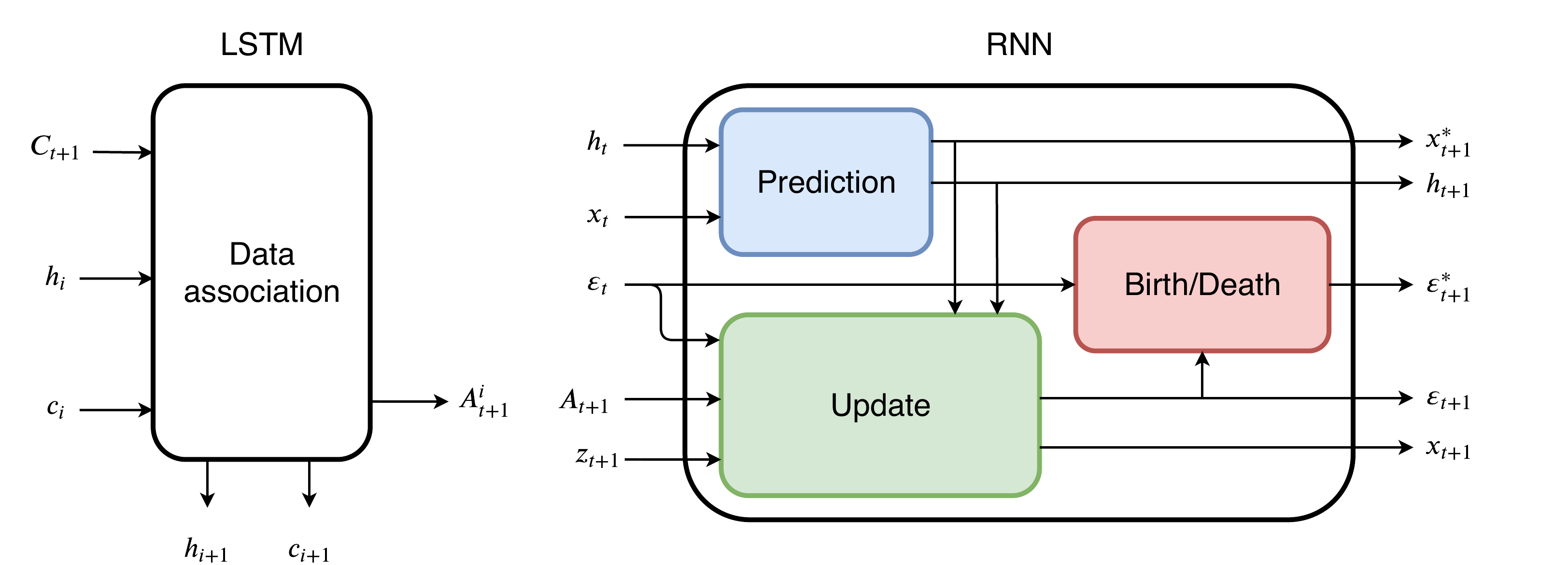}
    \caption{Diagram of the MOT algorithm proposed by Milan et al. \cite{milan2017online} employing a LSTM to predict detection associations. The algorithm used two different RNNs to solve them problem, each one specialized in one subtask. The LSTM (left) learned to associate detections with tracks given predicted positions. It received the pairwise-distance matrix between detections and predictions ($C_{t+1}$), the cell state ($c_i$) and the hidden state ($h_i$) as input, and output the vector $A^i_{t+1}$ representing the probability of associating target $i$ with the detections in the frame. The RNN (right) was trained to predict the position of the targets in the new frame and the possible birth and death of new ones. It received as input the hidden state ($h_t$) and the current position of the target ($x_t$), outputting the predicted position and the new hidden state (blue box). After the associations from the LSTM were computed, using the detections $z_{t+1}$, the positions of targets were updated (green box), and the existence probability $\varepsilon$ was computed to predict death and birth of trajectories (red box).}
    \label{fig:rnn-milan}
\end{figure}

One of the first works to use a deep network to directly compute an affinity is \cite{milan2017online}, where Milan et al. proposed an end-to-end learning approach for online MOT, summarized in figure \ref{fig:rnn-milan}. A recurrent neural network (RNN) based model was used as the main tracker, mimicking a Bayesian filter algorithm, consisting of three blocks: the first was a motion prediction block, that learned a motion model that took as input the state of the target in the past frames (i.e. the old bounding box locations and sizes) and predicted the target state in the next frame without accounting for the detections; the second block refined the state prediction using the detections in the new frame and an association vector containing the probability of associating the target with all such detections (it is evident how this can be considered an affinity score); the third block managed the birth and death of tracks, as it used the previous collected information to predict the probability of existence of the track in the new frame\footnote{To smooth the existence probability predictions and avoid deleting tracks of temporarily occluded objects, the difference between the new and the old existence probability was also output so that it could be minimized during training.}. The association vector was computed using a LSTM-based network, that used the Euclidean distance between the predicted state of the target and the states of the detections in the new frame as input features (besides the hidden state and the cell state, as any standard LSTM). The networks were trained separately using 100K 20-frame long synthetically generated sequences. While the algorithm performed favorably to other techniques, like the combination of a Kalman filter with the Hungarian algorithm, the results on the MOT15 test set did not quite reach top accuracy; however, the algorithm was able to run much faster than other algorithms ($\sim165$ FPS) and did not use any kind of appearance features, leaving room for future improvements.

Among the other works that later used LSTMs there is \cite{sadeghian2017tracking}, that used a LSTM with a fully-connected (FC) layer to fuse features extracted by 3 other LSTMs (as already explained in section \ref{sec:featext-complex}) and output an affinity score\footnote{The paper seems to imply that while the LSTM is trained to predict an affinity score, only the affinity features are extracted and are then used to replace the handcrafted features used in the MDP paper. The algorithm presented in the MDP paper, though, adds on top of those features another FC layer, trained with reinforcement learning to classify the track/detection pair as belonging to the same identity or not. Thus we can consider the overall affinity computation as performed by a deep learning model.}. The overall algorithm is similar to the Markov Decision Processes (MDP) based framework presented in \cite{xiang2015learning}: a single object tracker (SOT) is used to track targets independently; when a target gets occluded, the SOT is stopped and a bipartite graph is built that uses the affinities computed by the LSTM as edge costs and the association problem is then solved with the help of the Hungarian algorithm. The authors showed that using both a combination of all the 3 feature extractors and an LSTM rather than a plain FC layer led to consistently better performance on a MOT15 validation set. The algorithm also reached state-of-the-art MOTA scores on both MOT15 and MOT16 test sets at the time of publication, confirming the validity of the approach.

Another approach using multiple LSTMs is \cite{ran2019robust}, where Ran et al. proposed a Pose-based Triple Stream Network, that computed an affinity combining 3 other affinities output by 3 LSTMs: one for appearance similarity, using CNN features and pose information extracted with AlphaPose \cite{fang2017rmpe}, one for motion similarity, using pose joints velocity, and one for interaction similarity, using an interaction grid. A custom tracking algorithm is then used to associate the detections. The comparison with other state-of-the-art MOT algorithms on their proprietary Volleyball dataset for athlete tracking was favourable.

\subsubsection{Siamese LSTMs}\label{sec:aff-siamlstm}
Liang et al. \cite{liang2018lstm} also used multiple LSTMs to model various features, but they proceeded in a different way. Since extracting appearance features with CNNs is computationally expensive, they proceeded with a so-called \textit{pre-association step}, that used a SVM to predict the association probability between tracklets and detections. The SVM took as input position and velocity similarity scores, computed using two LSTMs for position and velocity prediction. The pre-association step then consisted in discarding the detections with low SVM affinity scores. After this step, an actual association step was performed by using VGG-16 features given as input to a Siamese LSTM, that predicted an affinity score between the tracklet and the detections. Association was performed in a greedy manner, associating the detection with the highest score to the tracklet. Testing was done on the MOT17 datasets, and the results were in line with the top-performing algorithms.

Wan et al. \cite{wan2018online} also used a Siamese LSTM in their algorithm, that was also composed of two steps. In the first step, short reliable tracklets were built by using Hungarian algorithm with affinity measures computed using the IoU between detections and the predicted target positions (obtained with Kalman filter or Lucas-Kanade optical flow). The second step also used the Hungarian algorithm to join the tracklets, but this time the affinity was computed using a Siamese LSTM framework that used motion features concatenated to appearance features extracted by a CNN (like in \cite{zhou2018deep2}), pre-trained on the CUHK03 Re-ID dataset.

\subsubsection{Bidirectional LSTMs}\label{sec:aff-bilstm}
A different usage of LSTMs in the affinity computation phase was presented by Zhu et al. \cite{zhu2018online}. They used a so-called Temporal Attention Network (TAN) to compute attention coefficients to weigh the features extracted by the Spatial Attention Network (SAN) (see section \ref{sec:featext-siamese}) to give less importance to noisy observations. A bidirectional LSTM was employed to this end. The whole network (called Dual Matching Attention Network) was used to recover from occlusions when a modified version of the Efficient Convolution Operators tracker (ECO) \cite{danelljan2017eco}, trained exploiting hard example mining, failed to detect a target. The algorithm obtained results comparable to online state-of-the-art methods on MOT16 and MOT17 according to various metrics (MOTA, IDF1, number of ID switches).

Yoon et al. \cite{yoon2019data} also used a Bidirectional LSTM to compute affinities, on top of some FC layers that encoded non-appearance features (only bounding box coordinates and detection confidences). The association was solved using the Hungarian algorithm. They trained the network on the Stanford Drone Dataset (SDD) \cite{robicquet2016learning} and evaluated it on both SDD and MOT15. They reached comparable results with top algorithms that did not use visual cues, but the performance was still worse than appearance-based methods.

\subsubsection{Uses of LSTMs in MHT frameworks}\label{sec:aff-lstmmht}
In Multiple Hypothesis Tracking approaches, a tree of potential track hypotheses for each candidate target is first built. Then the likelihood of each track is computed and the combination of tracks that has the highest likelihood is chosen as a solution. Various deep learning algorithms have also been employed to enhance MHT based approaches.

Kim et al. \cite{kim2018multi} proposed the use of a so-called \textit{Bilinear LSTM} network as the gating step of the MHT-DAM \cite{kim2015multiple} algorithm, that is, the affinity score computed by the LSTM was used to decide whether to prune or not a certain branch of the hypotheses tree. The LSTM cell had a modified forward pass (inspired by the online recursive least squares estimator proposed in \cite{kim2015multiple}) and took as input the appearance features of a tracklet in the past frames, extracted with a ResNet-50 CNN. The output of the LSTM cell was a feature matrix representing the historical appearance of the tracklet, and such matrix was then multiplied by the vector with the appearance features of the detection that needed to be compared with the tracklet. FC layers on top of that finally computed the affinity score between the tracklet and the detection. The authors claimed that such a modified LSTM is able to store longer-term appearance models than classical LSTMs. They also proposed to add a motion modeling classical LSTM to compute historical motion features (using the normalized bounding box coordinates and sizes), that were then concatenated to the appearance features before proceeding with the FC layers and the final softmax that output the affinity score. The two LSTMs were first trained separately and then fine-tuned jointly. The training data was also augmented including localization errors and missing detections, to resemble real-world data more closely. They used MOT15, MOT17, ETH, KITTI and other minor datasets for training and they evaluated the model on MOT16 and MOT17. They showed that their model is sensitive to the quality of detections, as they improved on the MOTA performance of MHT-DAM when using the public Faster R-CNN and SDP detections, while they performed worse than it on the public DPM detections. Anyway, they seemed to get a higher IDF1 score regardless of the detections used, and their overall results reflected that, since they got the highest IDF1 of all the methods using MHT-based algorithms. However, the tracking quality, as measured both in MOTA and in IDF1, was still lower than other state-of-the-art algorithms.

A similar use of a RNN has been recently presented by Maksai et al. \cite{maksai2018eliminating}, who also employed a LSTM to compute tracklet scores in a variation of the MHT algorithm, that grows and prunes tracklets iteratively and then tries to select the set of tracklets that maximizes said score.\footnote{While it is not an explicit affinity measure, it can still be seen as an evaluation of the effect of merging two tracklets and thus plays a similar role (i.e. taking decisions about associating tracklets) as other affinities presented in this section.} The goal of their work was to solve two frequent problems in training recurrent networks for multiple object tracking: the \textit{loss-evaluation mismatch}, that arises when a network is trained by optimizing a loss that is not well-aligned to the evaluation metric used at inference time (e.g. classification score vs. MOTA); the \textit{exposure bias}, that is present when the model is not exposed to its own errors during the training process. In order to solve the first problem, they introduced a novel way to score tracklets (using a RNN) that is a direct proxy of the IDF1 metric and does not use the ground truth; the network can then be trained to optimize such metric. The second problem was solved instead by adding to the training set of the network the tracklets computed using the current version of the network, together with hard example mining and random tracklet merging during training; in this way, the training set distribution should be more similar to the inference-time input data distribution. The network used was a Bidirectional LSTM, on top of an embedding layer that took as input various features. The authors presented a version of the algorithm using only geometric features and a version that instead used appearance-based features, that performed better. Lots of ablation studies were run, and various alternative approaches were tested. The final algorithm was able to reach top-performance on various MOT datasets (MOT15, MOT17, DukeMTMC \cite{ristani2016performance}) when considering the IDF1 metric, even though it didn't excel in MOTA.

Among the other approaches in the MHT family using RNNs we can also find \cite{chen2019recurrent}, where Chen et al. used a so-called \textit{Recurrent Metric Network} (RMNet) to compute appearance affinity between tracklet hypotheses and detections (together with a motion-based affinity) in their Batch Multi-Hypothesis Tracking strategy. The RMNet is an LSTM that takes as input appearance features of the detection sequence under consideration, extracted with a ResNet CNN, and outputs a similarity score together with bounding box regression parameters. A dual-threshold approach in gating and forming hypothesis was used, and a re-find reward was employed to encourage recovery from occlusions. The hypotheses were selected by casting the problem as a binary linear programming one, solved using \textit{lpsolve}. Kalman filter was finally used to smooth the trajectories. Evaluation was performed on MOT15, PETS2009 \cite{ferryman2009pets2009}, TUD \cite{andriluka2008people} and KITTI, obtaining better results on the IDF1 metric, that gives more weight to people re-identification, than on MOTA.

\subsubsection{Other recurrent networks}\label{sec:aff-otherrnn}
Fang et al. \cite{fang2018recurrent} used instead Gated Recurrent Units (GRUs) \cite{cho2014properties} inside their Recurrent Autoregressive Network (RAN) framework for pedestrian tracking. The GRUs were used to estimate the parameters of autoregressive models, one for motion and the other for appearance for each tracked target, that computed the probability of observing a given detection motion/appearance based on the tracklet's past motion/appearance features. The two probabilities, that can be easily seen as a kind of affinity measure, were then multiplied together to obtain a final association probability, used to solve a bipartite matching problem for association between tracklets and detections following the algorithm in \cite{xiang2015learning}. The RAN training step was formulated as a maximum likelihood estimation problem.

Kieritz et al. \cite{kieritz2018joint} used a recurrent 2-hidden-layer multi-layer perceptron (MLP) to compute an appearance affinity score between a detection and a tracklet. Such affinity was then given as input to another MLP, together with track and detection confidence scores, to predict an aggregate affinity score (called \textit{association metric}). Such score was finally used by the Hungarian algorithm to perform association. The method reached top performance on the UA-DETRAC dataset \cite{wen2015ua}, but the performance on MOT16 was not very good when compared with other algorithms using private detections.

\subsubsection{CNNs for affinity computation}\label{sec:aff-cnn}
Other algorithms used instead CNNs to compute some kind of similarity score. Tang et al. \cite{tang2017multiple} tested the use of 4 different CNNs to compute an affinity score between nodes in a graph, with the association task being formulated as a minimum cost lifted multicut problem \cite{andres2015lifting}: it can be seen as a graph clustering problem, where each output cluster represents a single tracked object. The costs associated to the edges accounted for the similarity between two detections. Such similarity was a combination of person re-identification confidence, deep correspondence matching and spatio-temporal relations. To compute the person re-identification affinity, various architectures were tested (after being trained on a dataset of 2511 identities extracted from MOT15, MOT16, CUHK03, Market-1501 datasets), but the best performing one was the novel StackNetPose. It incorporated body part information extracted using the DeepCut body part detector \cite{pishchulin2016deepcut} (see section \ref{sec:featext-cnn}). The 14 score maps for the body parts of two images were stacked together with the two images themselves to produce a 20-channel input. The network followed the VGG-16 architecture and output an affinity score between the two input identities. Differently from Siamese CNNs, the pair of images were able to `communicate' in the early stages of the network. The authors showed that the StackNetPose network performed better in the person re-identification task, and thus they used it to compute the ReID affinity. The combined affinity score was computed by multiplying a weight vector (learned with logistic regression, and dependent on the time interval between the two detections) with a 14-d vector containing ReID affinity, DeepMatching-based affinity \cite{weinzaepfel2013deepflow}, a spatio-temporal affinity score, the minimum of the two detection confidences and quadratic terms with all the pairwise combinations of the previously mentioned terms. The authors showed that combining all these features produced better results, and together with the improvement in framing the problem as a minimum cost lifted multicut problem (solved heuristically using the algorithm proposed in \cite{keuper2015efficient}), they managed to reach state-of-the-art performance (measured in MOTA score) on the MOT16 dataset at the time of publishing.

Another approach using CNNs was presented in \cite{chen2017online}, where Chen et al. used a Particle Filter \cite{arulampalam2002tutorial} to predict target motion, weighting the importance of each particle using a modified Faster R-CNN network. Such model was trained to predict the probability that the bounding box contains an object, but it was also augmented with a target-specific branch, that took as input features from lower layers of the CNN and merged them with the target historical features to predict the probability of the two objects being the same. The difference with the previous approaches is that here the affinity is computed between the sampled particles and the tracked target, instead of being computed between targets and detections. The detections that did not overlap with the tracked objects were instead used to initialize new tracks or retrieve missing objects. Despite being an online tracking algorithm, it was able to reach top performance on MOT15 at the time of publishing, both when using public detections and when using private ones (obtained from \cite{sanchez2016online}).

Zhou et al. \cite{zhou2018deep} used a visual-similarity CNN, similar to the ResNet-101 based visual-displacement CNN presented in section \ref{sec:featext-siamese}, that outputs affinity scores between the detections and the tracklet boxes predicted by the Deep Continuous Conditional Random Fields. This visual affinity score was merged with a spatial similarity using IoU, and then the detection with the highest score was associated to each tracklet; in case of conflicts, the Hungarian algorithm was employed. The method reached results comparable to state-of-the-art online MOT algorithms on MOT15 and MOT16 in terms of MOTA score.

\subsubsection{Siamese CNNs}\label{sec:aff-siamesecnn}
Siamese CNNs are also a common approach used in affinity computation. An example of Siamese CNN is shown in figure \ref{fig:siamese-cnn}. The approaches presented here decided to directly use the output of the Siamese CNN as an affinity, instead of employing classical distances between feature vectors extracted from the penultimate layer of the network, like the algorithms presented in section \ref{sec:featext-siamese}. For example, Ma et al. \cite{ma2018customized} used one to compute affinities between tracklets in a two-step algorithm. They chose to apply hierarchical correlation clustering, solving two successive lifted multicut problems: local data association and global data association. In the local data association step temporally-close detections were joined together by using the robust similarity measure presented in \cite{tang2016multi}, that uses DeepMatching and detection confidences to compute an affinity score between detections. In this step, only edges between close detections were inserted into the graph. The multicut problem was solved with the heuristic algorithm proposed in \cite{keuper2015efficient}. In the global data association step, local tracks that were split by long-term occlusion needed to be joined together, and a fully-connected graph with all the tracklets was then built. The Siamese CNN was used to compute the affinities that would serve as edge costs in the graph. The architecture was based on GoogLeNet \cite{szegedy2015going} and it was pretrained on ImageNet. The net was then trained on the Market-1501 ReID dataset and then fine-tuned on the MOT15 and MOT16 training sequences. Besides the verification layer, that output a similarity score between the two images, two classification layers were added to the network only during training to classify the identity of each training image; this was shown to improve the network performance in computing the affinity score. This so-called `generic' ReID net was also fine-tuned on each test sequence in an unsupervised manner, without using any ground truth information, to adapt the net to the illumination conditions, resolution, camera angle, etc. of each particular sequence. This was done by sampling positive and negative detection pairs by looking at the tracklets built in the local data association step. The effectiveness of the algorithm was proven by the results obtained on MOT16, where it is at the time of writing the best performing method with a published paper, with a 49.3 MOTA score.

As explained in section \ref{sec:featext-siamese}, Lee et al. \cite{lee2019multiple} used a Feature Pyramid Siamese Network to extract appearance features. When employing this kind of network in the MOT problem, a vector of motion features was concatenated to the appearance features and 3 fully-connected layers were then added on top to predict an affinity score between a track and a detection; the network was trained end-to-end. Detections were then associated iteratively, starting from the pairs with highest affinity scores and stopping when the score got below a threshold. The method obtained top performance results among the online algorithms on the MOT17 dataset at the time of publishing.

\subsection{DL in Association/Tracking step}
\label{sec:association}

Some works, albeit not as many as for the other steps in the pipeline, have used deep learning models to improve the association process performed by classical algorithms, like the Hungarian algorithm, or to manage the track status (e.g. by deciding to start or terminate a track). We are going to present them in this section, including the use of RNNs (section \ref{sec:ass-rnn}), deep multi-layer perceptrons (section \ref{sec:ass-mlp}) and deep reinforcement learning agents (section \ref{sec:ass-rl}).

\subsubsection{Recurrent neural networks} \label{sec:ass-rnn}
A first example of algorithms employing DL to manage the track status is the one presented by Milan et al. in \cite{milan2017online}, already described in section \ref{sec:aff-rnn}, that used a RNN to predict the probability of existence of a track in each frame, thus helping with the decision of when to initiate or terminate the tracks.

Ma et al. \cite{ma2018trajectory} used a bidirectional GRU RNN to decide where to split tracklets. The algorithm proceeded in three main stages: a tracklet generation step, that included a NMS step to remove redundant detections and then employed the Hungarian algorithm with appearance and motion affinity together to form high-confidence tracklets; then, a tracklet cleaving step was performed: since a tracklet might contain an ID switch error due to occlusions, this step aimed to split the tracklets at the point where the ID switch happened, in order to obtain two separate tracklets that contained the same identity; finally, a tracklet reconnection step was employed, using a customized association algorithm that made use of features extracted by a Siamese bidirectional GRU. The gaps within the newly-formed tracklets were then filled with polynomial curve fitting. The cleaving step was performed with a bidirectional GRU RNN, that used features extracted by a Wide Residual Network CNN \cite{zagoruyko2016wide}. The GRU output a pair of feature vectors for each frame (one for each direction of the GRU); then the distance between pairs of such feature vectors was computed and a distance vector was obtained. The highest value in this vector indicated where to split the tracklet, provided that the score was higher than a threshold. The reconnection GRU was similar, but it had an additional FC layer on top of the GRU and a temporal pooling layer to extract a feature vector representing the whole tracklet; the distance between the features of the two tracklets was then used to decide which tracklets to reconnect. The algorithm reached results comparable to state-of-the-art on the MOT16 dataset.

\subsubsection{Deep Multi-Layer Perceptron} \label{sec:ass-mlp}
Despite not being a very common approach, deep multi-layer perceptrons (MLP) have also been employed to guide the tracking process. For example, Kieritz et al. \cite{kieritz2018joint} used a MLP with two hidden layers to compute track confidence scores, taking as input the track score at the previous step and various information about the last associated detection (like association score and detection confidence). This confidence score was then used to manage the termination of tracks: they decided in fact to keep a fixed number of targets through time, replacing with new tracks the older ones that had the lowest confidence scores. The rest of the algorithm has been explained in section \ref{sec:aff-otherrnn}.

\subsubsection{Deep Reinforcement Learning agents} \label{sec:ass-rl}
Some works have used Deep Reinforcement Learning (RL) agents to take decisions in the tracking process. Rosello et al. \cite{rosello2018multi}, as explained in section \ref{sec:featext-other}, used multiple deep RL agents to manage the various tracked targets, deciding when to start and stop tracks and influencing the operation of the Kalman filter. The agent was modeled with a MLP with 3 hidden layers.

Ren et al. \cite{ren2018collaborative} also used multiple deep RL agents in a collaborative environment to manage the association task. The algorithm was mainly composed of two parts: a prediction network and a decision network. The prediction network was a CNN that was learned to predict the movement of the target in the new frame looking at the target and at the new image, and also using the recent tracklet trajectory. The decision network was instead a collaborative system that consisted of multiple agents (one for each tracked target) and the environment. Each agent took decisions based on the information about themselves, the neighbours and the environment; the interactions between the agents and the environment were exploited by maximizing a shared utility function: the agents thus did not operate independently from each other. Every agent/object was represented by a trajectory, its appearance features (extracted using MDNet \cite{nam2016learning}) and its current position. The environment was represented by the detections in the new frame. The detection network took as input, for each target, its predicted location in the new frame (output by the prediction network), the nearest target and the nearest detection, and based on various factors, such as the detection reliability and the target occlusion status, took one among various actions: updating the track and its appearance features using both the prediction and the detection, ignoring the detection and only using the prediction to update the track, detecting an occlusion of the tracked target, deleting the track. The agents were modelled using 3 FC layers on top of the feature extraction part of the MDNet. Various ablation studies showed the effectiveness of using the prediction and detection networks instead of linear motion models and Hungarian algorithm, respectively, and the method obtained very good results on the MOT15 and MOT16 datasets, reaching state-of-the-art performance among online methods, despite suffering from a relatively high number of ID switches.

\subsection{Other uses of DL in MOT}
\label{sec:others}
In this section we will present other interesting uses of deep learning models that don't neatly fit into one of the four common steps of a multiple object tracking algorithm. For this reason, such works have not been included in table \ref{tab:summary_table}, but are summarized instead in table \ref{tab:summary_oth}.

\begin{table}[htb]
\centering
\begin{tabular}{c>{\centering\arraybackslash}p{0.12\textwidth}>{\centering\arraybackslash}p{0.54\textwidth}>{\centering\arraybackslash}c>{\centering\arraybackslash}p{0.08\textwidth}}
\toprule
                          & Detection & Description  & Mode   & Source and data \\ \midrule
\cite{jiang2018precise}   & N/A       & They integrate a bounding box regression step in various existing MOT algorithms. The regression is done using Deep Reinforcement Learning using CNN features.                                                                                                                      & N/A    &                      \\
\cite{lee2016multi}       & Public    & An ensemble of 2 CNNs, color histograms and a KLT motion detector are used to compute likelihoods for a Markov Chain Monte Carlo sampling; the position sampling was used to form short tracklets. A Changing Point Detection algorithm was employed to merge and delete tracklets. & Online &                      \\
\cite{hoak2017image}      & CNN       & Multi-Bernoulli Filter with a novel Interactive Likelihood, computed using a CNN.                                                                                                                                                                                                   & Online &                      \\
\cite{henschel2018fusion} & Public    & Body detections are refined using head detections obtained with a CNN \cite{stewart2016end}. A modified version of the Frank-Wolfe algorithm is used to solve a correlation clustering problem for association, using spatial and temporal costs.                                   & Batch  &                      \\
\cite{gan2018online}      & Public    & Modified MDNet CNN with target-specific branches to compute affinities between targets and candidates extracted with Gaussian sampling. Combination of appearance and motion features to reduce ID Switches.                                                                        & Online &                      \\
\cite{xiang2019online}    & Public    & CNN to extract app features and LSTM to extract motion features. The LSTM is part of a BF-Net, that performs Bayesian filtering and uses the output from Hungarian algorithm for track refinement.                                                                                 & Online &                      \\
\cite{chu2019online}      & Public    & PafNet and PartNet CNNs to distinguish targets from background and among themselves. KCF SOT tracker is used. SVM+Hungarian algorithm for error recovering. CNN trained with RL for model updating.                                                                                 & Online &                     \\ \bottomrule
\end{tabular}
\caption{Information summary about methods using DL that don't fit the 4-step scheme.}
\label{tab:summary_oth}
\end{table}

One example is \cite{jiang2018precise}, where Jiang et al. use a Deep RL agent to perform bounding box regression after the use of one of many MOT algorithms. The procedure is in fact completely independent from the tracking algorithm employed, and can be used a posteriori to increase the accuracy of the model. A VGG-16 CNN was used to extract appearance features from the region enclosed by the bounding box, then those features were concatenated to a vector representing the history of the last 10 actions taken by the agent. Finally a Q-network \cite{watkins1989learning} made of 3 fully-connected layers was used to predict one among 13 possible actions, that included motion and scaling of the bounding box and a termination action, to signal the completion of the regression. The use of this bounding box regression technique on various state-of-the-art MOT algorithms allowed an improvement between 2 and 7 absolute MOTA points on the MOT15 dataset, reaching top score among public detections methods. The authors also showed that their regression approach had better results than using conventional methods, such as the bounding box regression computed by a Faster R-CNN model.

Lee et al. \cite{lee2016multi} proposed a multi-class multi-object tracker that used an ensemble of detectors, including CNN models like VGG-16 and ResNet, to compute the likelihood of each target being at a certain location in the next frame. A Markov Chain Monte Carlo sampling from a distribution that was influenced by said likelihoods was used to predict the next position for each target, and together with an estimation of track birth and death probabilities, short track segments were built. Finally, a changing point detection \cite{takeuchi2006unifying} algorithm was employed to detect abrupt changes in stationary time series representing track segments; this was done in order to detect track drift, to remove unstable track segments and to combine the segments together. The algorithm reached results comparable to state-of-the-art MOT methods using private detections.

Hoak et al. \cite{hoak2017image} proposed a 5-layer custom CNN network, trained on the Caltech pedestrian detection dataset \cite{dollar2009pedestrian}, to compute the likelihood of a target being at a certain location in the image. They used a multi-Bernoulli filter (implemented using the particle filter algorithm presented in \cite{hoseinnezhad2012visual}), and a novel Interactive Likelihood (ILH) was computed for each particle, in order to weigh them based on their distance from particles belonging to other targets; this was done to prevent the algorithm from sampling from areas that belong to different objects. The algorithm obtained good results on the VSPETS 2003 INMOVE soccer dataset\footnote{\url{ftp://ftp.cs.rdg.ac.uk/pub/VS-PETS/}} and the AFL dataset \cite{milan2014improving}.

Henschel et al. \cite{henschel2018fusion} used head detections, extracted with a CNN \cite{stewart2016end}, in addition to the usual body detections to perform pedestrian tracking. The presence/absence of a head and its position relative to the bounding box can help determine if a bounding box is a true or a false positive. The association problem was modelled as a correlation clustering problem on graphs, that the authors solved with a modified version of the Frank-Wolfe algorithm \cite{frank1956algorithm}; the association costs were computed as a combination of spatial and temporal costs: the spatial costs were the distance and the angle between the detected and the predicted head positions; the temporal costs were computed using the correspondences between pixels between the two frames, obtained using DeepMatching \cite{revaud2016deepmatching}. The algorithm reached top MOTA score on MOT17 and second-best score on MOT16 at the time of publication.

Gan et al. \cite{gan2018online} employed a modified MDNet \cite{nam2016learning} in their online pedestrian tracking framework. Besides 3 shared convolutional layers, common to all the targets, each target also had 3 specific FC layers, that were updated online to capture the appearance change of the target. A set of box candidates, including detections intersecting the last bounding box of the target and a set of boxes sampled from a Gaussian distribution with parameters estimated using a linear motion model, were given as input to the network, that output a confidence score for each of them. The candidate with the highest score was considered the optimal estimated target location. To reduce the number of ID switch errors, the algorithms tried to find the past trajectory that was most similar to the estimated box, using another affinity measure between the pairs; such affinity was computed using appearance and motion cues, together with the tracklet confidence score and a collision factor. Detections were also used to initialize new tracklets and to fix the motion prediction errors when occlusions happened.

Xiang et al. \cite{xiang2019online} used a MetricNet to track pedestrians. The model unified an affinity model with trajectory estimation, done with a Bayesian filter. An appearance model, made of a VGG-16 CNN trained for person re-identification on various datasets, extracted features and performed bounding box regression; the motion model instead consisted of two parts: an LSTM-based feature extractor, that took as input the trajectory's past coordinates, and a so-called \textit{BF-Net} on top, made of various FC layers, that combined the features extracted by the LSTM and a detection box (chosen by the Hungarian algorithm) to perform the Bayesian filtering step and output the new position of the target. The MetricNet was trained using a triplet loss, similar to other models presented in the previous sections. The algorithm obtained the best and second-best results among online methods on MOT16 and MOT15, respectively.

Finally, Chu et al. \cite{chu2019online} used three different CNNs in their algorithm. The first one, called PafNet \cite{cao2017realtime}, was used to distinguish the background from the tracked objects. The second one, called PartNet \cite{zhao2017deeply}, was employed to distinguish among the different targets. The third CNN, made of one convolutional layer and one FC layer, was instead used to decide whether to refresh the tracking model or not. The overall algorithm worked as follows: for every tracked target in the past frame, two score maps were computed in the current one, using PafNet and PartNet. Then, using the Kernel Correlation Filter tracker \cite{henriques2014high}, a new position for the object was predicted. Moreover, after a certain number of frames, a so-called detection verification step was performed: the detections output by a detector (in their experiments, they chose to use the public detections provided with the dataset) were assigned to the tracked targets by solving a graph multicut problem. Targets that were not associated to a detection for a certain number of frames were terminated. Then, the third CNN was employed to check if the associated detection box was better than the predicted one. If so, the KCF model parameters were updated to reflect the change in the object characteristics. Such CNN used the maps extracted by PafNet, and was trained using reinforcement learning. Unassociated detections were then employed to recover from target occlusion, using a SVM classifier and the Hungarian algorithm. Finally, the remaining unassociated detections were used to initialize new targets. The algorithm was evaluated both on MOT15 and MOT16 datasets, reaching top performance overall on the first one, and top performance among online methods on the second.

\section{Analysis and comparisons}\label{sec:comparisons}
This section presents a comparison between all the works that have tested their algorithm on one of the MOTChallenge datasets. We will only focus on the MOTChallenge datasets since for other datasets there aren't enough relevant papers using deep learning to perform a meaningful analysis.

We first describe the setup of the experimental analysis, including the considered metrics and the organization of the tables in section \ref{sec:comp-setup}. Section \ref{sec:comp-disc} will then present the actual results and considerations derived from the analysis.

\subsection{Setup and organization}\label{sec:comp-setup}
For a fair comparison, we only show results reported on the whole test sets. Some of the discussed papers report their results using subsets of the test set, or validation datasets extracted from the training splits of the MOTChallenge datasets. These results are discarded as they are not comparable with the others. Moreover, the reported results are divided into algorithms that use public detections and algorithms that use private detections, since the different quality of the detections has a big impact on performance. The results are further split into online and batch methods, since the online methods are at a disadvantage, being only able to access present and past information to assign IDs in each frame.

For each algorithm we indicate the year of the referenced published paper, their mode of operation (batch vs. online); the MOTA, MOTP, IDF1, Mostly Tracked (MT) and Mostly Lost (ML) metrics, expressed in percentages; the absolute number of false positives (FP), false negatives (FN), ID switches (IDS) and fragmentations (Frag); the speed of the algorithm expressed in frames per second (Hz). For each metric, an arrow pointing up ($\uparrow$) indicates that a higher score is better, while an arrow pointing down ($\downarrow$) indicates the opposite. The metrics shown here are the same that can be found on the public leaderboards on the MOTChallenge website. The numerical results presented in the referenced works have been integrated with data from the MOTChallenge leaderboards. 

Attending to the classification presented before, a table for each of the combinations dataset/detection source is shown. Tables \ref{tab:mot15pub} and \ref{tab:mot15priv} show results on MOT15 using public and private detections respectively; tables \ref{tab:mot16pub} and \ref{tab:mot16priv} do the same on MOT16; finally, table \ref{tab:mot17pub} shows results on MOT17, who currently only has published algorithms that use public detections. Each table groups online and batch methods separately, and for each group the papers are sorted by year, and then by ascending MOTA score if the papers are from the same year, since it is the main metric considered in the MOTChallenge datasets\footnote{If not differently specified, when we use in this section expressions such as "best performing" or similar, we are always referring to a higher MOTA score, since it's the main evaluation metric used in the MOTChallenge benchmark.}. If a work presents multiple results on the same dataset, using the same set of detections and the same mode of operation, we only show the result with the highest MOTA. The best performance for each metric is highlighted in \textbf{bold}, while the best performance among papers operating in the same mode (batch/online) is \underline{underlined}. It is important to note though that comparisons on the Hz metric may not be reliable since the performance is usually reported only for the tracking part of the algorithms, without the detection step and sometimes without including the runtime of deep learning models, that are usually the most computational intensive part of the algorithms presented in this survey; moreover, the algorithms were run on widely different hardware.

\begin{table}[htb]
\centering
\begin{tabular}{ccc
*{5}{S[table-format=2.1, table-number-alignment=center]}
S[table-format=4.0, table-number-alignment=center]
S[table-format=5.0, table-number-alignment=center]
*{2}{S[table-format=4.0, table-number-alignment=center]}
S[table-format=3.1, table-number-alignment=center]}
\toprule
 & \head{Year} & \head{Mode} & \head{MOTA $\uparrow$} & \head{MOTP $\uparrow$} & \head{IDF1 $\uparrow$} & \head{MT $\uparrow$} & \head{ML $\downarrow$} & \head{FP $\downarrow$} & \head{FN $\downarrow$} & \head{IDS $\downarrow$} & \head{Frag $\downarrow$} & \head{Hz $\uparrow$} \\ \midrule
\cite{kim2015multiple}       & 2015 & \multirow{15}{*}{Online} & 32.4 & 71.8 & 45.3 & 16.0 & 43.8 & 9064  & 32060 & 435  & \Uline{826}  & 0.7  \\
\cite{milan2017online}       & 2017 &        & 19.0 & 71.0 & 17.1 & 5.5  & 45.6 & 11578 & 36706 & 1490 & 2081 & 165.2\\
\cite{wang2017online}        & 2017 &        & 31.6   & 71.8   &        & 10.1 & 46.3 &      &       & 491    & 994 &          \\
\cite{bae2017confidence}     & 2017 &        & 32.8   & 70.7   & 38.8   & 9.7  & 42.2 & 4983 & 35690 & 614    & 1583     & 2.3      \\
\cite{chu2017online}         & 2017 &        & 34.3   & 70.5   & 48.3   & 11.4 & 43.4 & 5154 & 34848 & \bfseries 348    & 1463     & 0.5      \\
\cite{yang2017hybrid}        & 2017 &        & 35.0 & \Uline{72.6} & 47.7 & 11.4 & 42.2 & 8455 & 31140 & 358 & 1267 & 4.6 \\
\cite{sadeghian2017tracking} & 2017 &        & 37.6   & 71.7   & 46.0   & 15.8 & \bfseries 26.8 & 7933 & \bfseries 29397 & 1026   & 2024     & 1.0      \\
\cite{chen2017online}        & 2017 &        & 38.5   & \Uline{72.6} & 47.1   & 8.7  & 37.4 & \bfseries 4005 & 33204 & 586    & 1263     & 6.7      \\
\cite{zhou2018deep}          & 2018 &        & 33.6   & 70.9   & 39.1 & 10.4 & 37.6 & 5917 & 34002 & 866    & 1566     &  0.1 \\
\cite{fang2018recurrent}     & 2018 &        & 35.1   & 70.9   & 45.4  & 13.0 & 42.3 & 6771 & 32717 & 381    & 1523     &  5.4  \\
\cite{ren2018collaborative}  & 2018 &        & 37.1   & 71.0   &        & 14.0 & 31.3 & 7036 & 30440 &        &          &          \\ 
\cite{jiang2018precise}      & 2018 &        & \bfseries 42.3 &  & 47.7 & 13.6 & 39.7 & & & & & 3.1\\ 
\cite{yoon2019data}          & 2019 &        & 22.5 & 70.9   & 25.9   & 6.4  & 61.9 & 7346 & 39092 & 1159   & 1538     & \bfseries 172.8    \\
\cite{xiang2019online}       & 2019 &        & 37.1   & 72.5   & \bfseries 48.4   & 12.6 & 39.7 & 8305 & 29732 & 580    & 1193     & 1.0      \\
\cite{chu2019online}         & 2019 &        & 38.9   & 70.6   &  44.5  & \bfseries 16.6 & 31.5 & 7321 & 29501 & 720    & 1440     & 0.3 \\
\midrule
\cite{leal2016learning}      & 2016 & \multirow{5}{*}{Batch} & 29.0   & 71.2   & 34.3    & 8.5 & 48.4 & \Uline{5160} & 37798  & 639    & 1316     &  \Uline{52.8} \\
\cite{wang2016joint}         & 2016 &        & 29.6   & 71.8   & 36.8  & 11.2 & 44.0 & 7786 & 34733 & 712    & 943      & 1.7 \\
\cite{son2017multi}          & 2017 &        & \Uline{33.8} & 73.4   & \Uline{40.4} & \Uline{12.9} & \Uline{36.9} & 7898 & \Uline{32061} & 703    & 1430     & 3.7      \\
\cite{maksai2018eliminating} & 2018 &        & 22.2   & 71.1   & 27.2 & 3.1  & 61.6 & 5591 & 41531 & 700    & 1240     & 8.9      \\ 
\cite{chen2019recurrent}     & 2019 &        & 28.1   & \bfseries 74.3   & 38.7   &      &       & 6733 & 36952 & \Uline{477} & \bfseries 790      & 16.9     \\
\bottomrule
\end{tabular}
\caption{Experimental results of MOT algorithms using deep learning and public detections on MOT15 dataset.}
\label{tab:mot15pub}
\end{table}

\begin{table}[htb]
\centering
\begin{tabular}{ccc
*{5}{S[table-format=2.1, table-number-alignment=center]}
S[table-format=4.0, table-number-alignment=center]
S[table-format=5.0, table-number-alignment=center]
*{2}{S[table-format=4.0, table-number-alignment=center]}
S[table-format=3.1, table-number-alignment=center]}
\toprule
 & \head{Year} & \head{Mode} & \head{MOTA $\uparrow$} & \head{MOTP $\uparrow$} & \head{IDF1 $\uparrow$} & \head{MT $\uparrow$} & \head{ML $\downarrow$} & \head{FP $\downarrow$} & \head{FN $\downarrow$} & \head{IDS $\downarrow$} & \head{Frag $\downarrow$} & \head{Hz $\uparrow$} \\ \midrule
\cite{bewley2016simple}      & 2016 &  \multirow{6}{*}{Online} & 33.4 & 72.1 & 40.4 & 11.7 & 30.9 & 7318  & 32615 & 1001 & 1764 & \bfseries 260.0 \\
\cite{bullinger2017instance} & 2017 &  & 32.1 & 70.9 &      & 13.2 & 30.1 & 6551  & 33473 & 1687 & 2471 &    \\
\cite{bae2017confidence}     & 2017 &  & 51.3 & 74.2 & 54.1 & 36.3 & 22.2 & 7110  & 22271 & 544  & \bfseries 1335 & 1.3  \\
\cite{chen2017online}        & 2017 &  & 53.0 & \bfseries 75.5 & 52.2 & 29.1 & 20.2 & \bfseries 5159  & 22984 & 708  & 1476 & 6.7  \\
\cite{zhao2018multi}         & 2018 &  & 32.7 &      & 38.9 & 26.2 & 19.6 &       &       &      &      & 11.1 \\
\cite{fang2018recurrent}     & 2018 &  & \bfseries 56.5 & 73.0 & \bfseries 61.3 & \bfseries 45.1 & \bfseries 14.6 & 9386  & \bfseries 16921 & \bfseries 428  & 1364 & 5.1 
\\
\bottomrule
\end{tabular}
\caption{Experimental results of MOT algorithms using deep learning and private detections on MOT15 dataset.}
\label{tab:mot15priv}
\end{table}

\begin{table}[htb]
\centering
\begin{tabular}{ccc
*{5}{S[table-format=2.1, table-number-alignment=center]}
S[table-format=5.0, table-number-alignment=center]
S[table-format=5.0, table-number-alignment=center]
*{2}{S[table-format=4.0, table-number-alignment=center]}
S[table-format=2.1, table-number-alignment=center]}
\toprule
 & \head{Year} & \head{Mode} & \head{MOTA $\uparrow$} & \head{MOTP $\uparrow$} & \head{IDF1 $\uparrow$} & \head{MT $\uparrow$} & \head{ML $\downarrow$} & \head{FP $\downarrow$} & \head{FN $\downarrow$} & \head{IDS $\downarrow$} & \head{Frag $\downarrow$} & \head{Hz $\uparrow$} \\ \midrule
\cite{kim2016similarity}     & 2016 & \multirow{13}{*}{Online} & 35.3 & 75.2 &      & 7.4  & 51.1 & 5592  & 110778 & 1598 & 5153   & 7.9    \\
\cite{sanchez2016online}     & 2016 &  & 38.8 & 75.1 &      & 7.9  & 49.1 & 8114  & 102452 & 965  & 1657   & 11.8   \\
\cite{bae2017confidence}     & 2017 &  & 43.9 & 74.7 & 45.1 & 10.7 & 44.4 & 6450  & 95175  & 676  & 1795   & 0.5    \\
\cite{chu2017online}         & 2017 &  & 46.0 & 74.9 & 50.0 & 14.6 & 43.6 & 6895  & 91117  & \Uline{473}  & 1422   & 0.2    \\
\cite{sadeghian2017tracking} & 2017 &  & 47.2 & 75.8 & 46.3 & 14.0 & 41.6 & \bfseries 2681  & 92856  & 774  & 1675   & 1.0    \\
\cite{gan2018online}         & 2018 &  & 44.2 & \Uline{78.3} &      & 15.2 & 45.7 & 7912  & 93215  & 560  & 1212   &        \\
\cite{zhou2018deep}          & 2018 &  & 44.8 & 75.6 & 39.7 & 14.1 & 42.3 & 5613  & 94125  & 968  & 1378   & 0.1    \\
\cite{fang2018recurrent}     & 2018 &  & 45.9 & 74.8 & 48.8 & 13.2 & 41.9 & 6871  & 91173  & 648  & 1992   & 0.9    \\
\cite{zhu2018online}         & 2018 &  & 46.1 & 73.8 & \bfseries 54.8 & \Uline{17.4} & 42.7 & 7909  & 89874  & 532  & 1616   & 0.3    \\
\cite{ren2018collaborative}  & 2018 &  & 47.3 & 74.6 &      & \Uline{17.4} & 39.9 & 6375  & 88543  &      &        &        \\
\cite{long2018tracking}      & 2018 &  & 47.6 & 74.8 & 50.9 & 15.2 & 38.3 & 9253  & \Uline{85431}  & 792  & 1858   & \bfseries 20.6   \\
\cite{xiang2019online}       & 2019 &  & 48.3 & 76.7 & 50.9 & 15.4 & 40.1 & 2706  & 91047  & 543  & \Uline{896}    & 0.5    \\
\cite{chu2019online}         & 2019 &  & \Uline{48.8} & 75.7 & 47.2 & 15.8 & \Uline{38.1} & 5875  & 86567  & 906  & 1116   & 0.1    \\ \midrule
\cite{son2017multi}          & 2017 & \multirow{10}{*}{Batch} & 44.1 & 76.4 & 38.3 & 14.6 & 44.9 & 6388  & 94775  & 745  & 1096   & 1.8    \\
\cite{chen2017enhancing}     & 2017 &  & 45.3 & 75.9 & 47.9 & 17.0 & 39.9 & 11122 & 87890  & 639  & 946    & 1.8    \\
\cite{tang2017multiple}      & 2017 &  & 48.8 & \bfseries 79.0 &      & 18.2 & 40.1 & 6654  & 86245  & 481  & 595    & 0.5    \\
\cite{kim2018multi}          & 2018 &  & 42.1 &      & 47.8 & 14.9 & 44.4 & 11637 & 93172  & 753  & 1156   & 1.8    \\
\cite{babaee2018occlusion}   & 2018 &  & 46.9 & 76.4 & 46.8 & 16.1 & 43.2 & 6257  & 91669  & 549  & 757    &        \\
\cite{sheng2018heterogeneous}& 2018 &  & 47.2 & 75.7 & \Uline{52.4} & 18.6 & 42.8 & 12586 & 83107 & 542 & 787 & 0.5 \\
\cite{wen2019learning}       & 2018 &  & 47.5 &      & 43.6 & \bfseries 19.4 & \bfseries 36.9 & 13002 & \bfseries 81762  & 1035 & 1408   & 0.8    \\
\cite{henschel2018fusion}    & 2018 &  & 47.8 & 75.5 & 44.3 & 19.1 & 38.2 & 8886  & 85487  & 852  & 1534   & 0.6    \\
\cite{ma2018trajectory}      & 2018 &  & 48.2 & 77.5 & 48.6 & 12.9 & 41.1 & \Uline{5104}  & 88586  & 821  & 1117   & \Uline{2.8}    \\
\cite{ma2018customized}      & 2018 &  & \bfseries 49.3 & \bfseries 79.0 & 50.7 & 17.8 & 39.9 & 5333  & 86795  & \bfseries 391  & \bfseries 535    & 0.8    \\ \bottomrule
\end{tabular}
\caption{Experimental results of MOT algorithms using deep learning and public detections on MOT16 dataset.}
\label{tab:mot16pub}
\end{table}

\begin{table}[htb]
\centering
\begin{tabular}{ccc
*{5}{S[table-format=2.1, table-number-alignment=center]}
S[table-format=5.0, table-number-alignment=center]
S[table-format=5.0, table-number-alignment=center]
*{2}{S[table-format=4.0, table-number-alignment=center]}
S[table-format=2.1, table-number-alignment=center]}
\toprule
 & \head{Year} & \head{Mode} & \head{MOTA $\uparrow$} & \head{MOTP $\uparrow$} & \head{IDF1 $\uparrow$} & \head{MT $\uparrow$} & \head{ML $\downarrow$} & \head{FP $\downarrow$} & \head{FN $\downarrow$} & \head{IDS $\downarrow$} & \head{Frag $\downarrow$} & \head{Hz $\uparrow$} \\ \midrule
 \cite{yu2016poi}              & 2016 & \multirow{8}{*}{Online} & \Uline{66.1} & \Uline{79.5} & 65.1 & 34.0 & 20.8 & \Uline{5061}  & 55914 & 805  & 3093   & 9.9    \\ 
\cite{wojke2017simple}        & 2017 &         & 61.4 & 79.1 & 62.2 & 32.8 & \bfseries 18.2 & 12852 & 56668 & 781  & 2008   & 17.4   \\
\cite{kieritz2018joint}       & 2018 &         & 39.1 &      &      & 11.1 & 41.1 & 9411  & 99727 & 1906 &        & 4.5   \\
\cite{ujiie2018interpolation} & 2018 &         & 55.0 & 76.7 &      & 20.4 & 24.5 & 15766 & 65297 & 1024 & 1594   & 16.9   \\
\cite{wan2018online}          & 2018 &         & 62.6 & 78.3 &      & 32.7 & 21.1 & 10604 & 56182 & 1389 & 1534   &        \\
\cite{fang2018recurrent}      & 2018 &         & 63.0 & 78.8 & 63.8 & 39.9 & 22.1 & 13663 & 53248 & \Uline{482}  & 1251   & 1.6    \\
\cite{zhou2018online}         & 2018 &         & 64.8 & 78.6 & \bfseries 73.5 & \Uline{40.6} & 22.0 & 13470 & \Uline{49927} & 794  & \Uline{1050}  & \bfseries 39.4   \\
\cite{mahmoudi2019cnnmtt}     & 2019 &         & 65.2 & 78.4 & 62.2 & 32.4 & 21.3 & 6578  & 55896 & 946  & 2283   & 11.2   \\
\midrule
\cite{lee2016multi}           & 2016 & \multirow{4}{*}{Batch} & 62.4 & 78.3 & 51.6 & 31.5 & 24.2 & 9855  & 57257 & 1394 & 1318   & \Uline{34.9}   \\
\cite{yu2016poi}              & 2016 &         & 68.2 & 79.4 & 60.0 & 41.0 & \Uline{19.0} & 11479 & 45605 & 933  & 1093   & 0.7    \\
\cite{tang2017multiple}       & 2017 &         & \bfseries 71.0 & \bfseries 80.2 & \Uline{70.1} & \bfseries 46.9 & 21.9 & 7880 & \bfseries 44564 & \bfseries 434 & \bfseries 587 & 0.5\\ 
\cite{babaee2018occlusion}    & 2018 &         & 58.1 & 77.2 & 47.4 & 23.1 & 33.3 & \bfseries 4883 & 70207 & 1624 & 2539 &          \\
\bottomrule
\end{tabular}
\caption{Experimental results of MOT algorithms using deep learning and private detections on MOT16 dataset.}
\label{tab:mot16priv}
\end{table}

\begin{table}[htb]
\centering
\begin{tabular}{ccc
*{5}{S[table-format=2.1, table-number-alignment=center]}
S[table-format=5.0, table-number-alignment=center]
S[table-format=6.0, table-number-alignment=center]
*{2}{S[table-format=4.0, table-number-alignment=center]}
S[table-format=2.1, table-number-alignment=center]}
\toprule
 & \head{Year} & \head{Mode} & \head{MOTA $\uparrow$} & \head{MOTP $\uparrow$} & \head{IDF1 $\uparrow$} & \head{MT $\uparrow$} & \head{ML $\downarrow$} & \head{FP $\downarrow$} & \head{FN $\downarrow$} & \head{IDS $\downarrow$} & \head{Frag $\downarrow$} & \head{Hz $\uparrow$} \\ \midrule
\cite{gan2018online}         & 2018 & \multirow{5}{*}{Online} & 44.9 & \bfseries 78.9 &      & 13.8 & 44.2 & \bfseries 22085 & 287267 & \Uline{1537} & \Uline{3295}   &        \\
\cite{fu2018gm}              & 2018 &  & 46.5 & 77.2 &      & 16.9 & 37.2 & 23859 & 272430 & 5649 & 9298   & 1.6    \\
\cite{zhu2018online}         & 2018 &  & 48.2 & 75.7 & \Uline{55.7} & \Uline{19.3} & 38.3 & 26218 & 263608 & 2194 & 5378   & 0.3    \\
\cite{long2018tracking}      & 2018 &  & \Uline{50.9} & 76.6 & 52.7 & 17.5 & \Uline{35.7} & 24069 & \Uline{250768} & 2474 & 5317 & \bfseries 18.3 \\
\cite{lee2019multiple}       & 2019 &  & 44.9 & 76.6 & 48.4 & 16.5 & 35.8 & 33757 & 269952 & 7136 & 14491  & 10.1   \\ \midrule
\cite{maksai2018eliminating} & 2018 & \multirow{5}{*}{Batch} & 44.2 & 76.4 & \bfseries 57.2 & 16.1 & 44.3 & 29473 & 283611 & \bfseries 1529 & \bfseries 2644   & \Uline{4.8}    \\
\cite{kim2018multi}          & 2018 &  & 47.5 &      & 51.9 & 18.2 & 41.7 & 25981 & 268042 & 2069 & 3124   & 1.9    \\
\cite{liang2018lstm}         & 2018 &  & 50.3 &      & 47.9 & 21.8 & 36.2 & \Uline{22204} & 249342 & 3243 & 3155   & 1.9    \\
\cite{henschel2018fusion}    & 2018 &  & 51.3 & \Uline{77.0} & 47.6 & 21.4 & \bfseries 35.2 & 24101 & 247921 & 2648 & 4279   & 0.2    \\
\cite{sheng2018heterogeneous} & 2018 &  & \bfseries 51.8 & \Uline{77.0} & 54.7 & \bfseries 23.4 & 37.9 & 33212 & \bfseries 236772 & 1834 & 2739 & 0.7 \\ \bottomrule
\end{tabular}
\caption{Experimental results of MOT algorithms using deep learning and public detections on MOT17 dataset.}
\label{tab:mot17pub}
\end{table}

\subsection{Discussion of the results}\label{sec:comp-disc}
\subsubsection*{General observations}
\begin{figure}
    \centering
    \begin{subfigure}[t]{0.33\textwidth}
        \includegraphics[width=\textwidth]{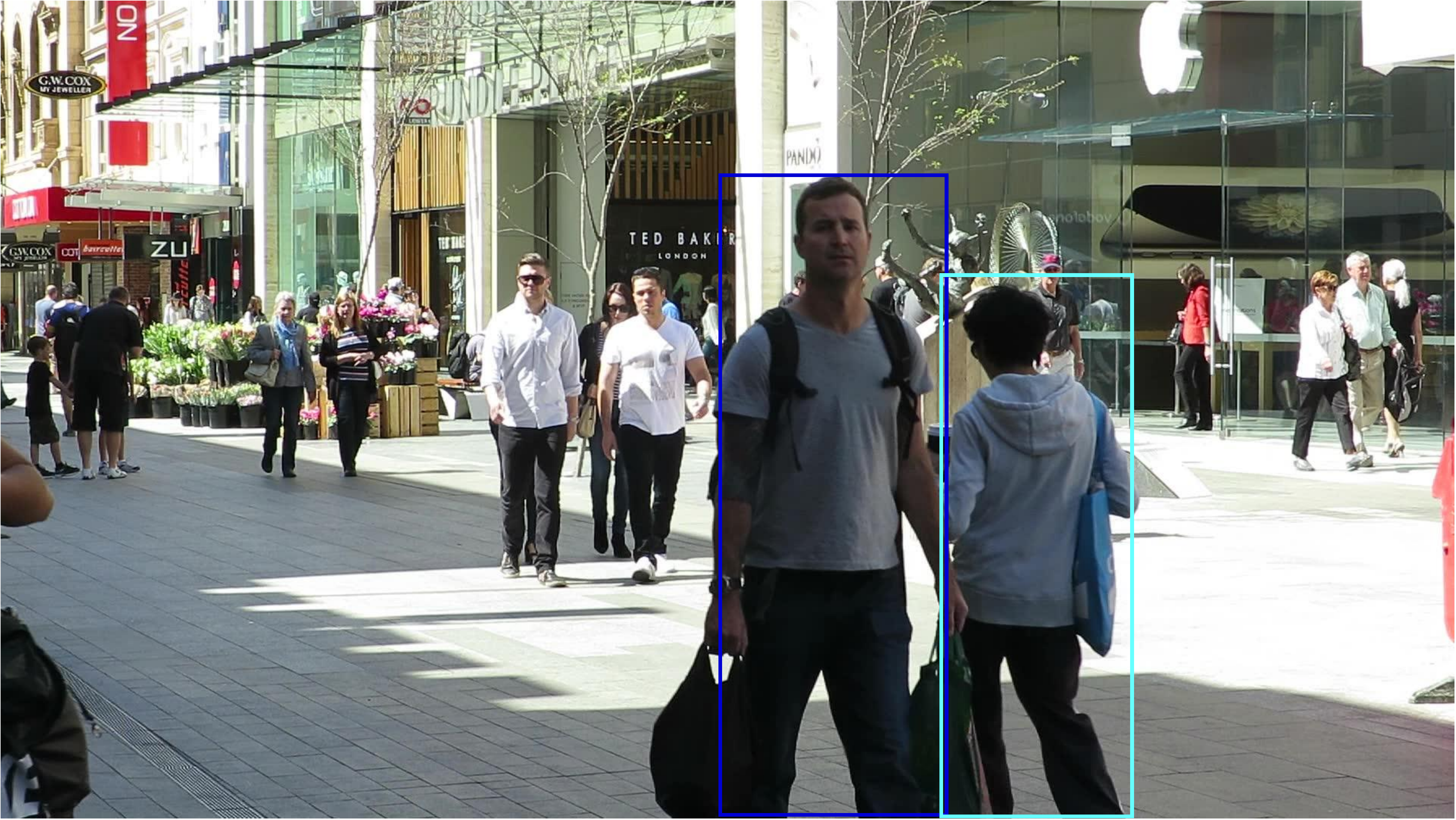}
        \caption{MOTDT output before occlusion}
    \end{subfigure}
    \hfill
    \begin{subfigure}[t]{0.33\textwidth}
        \includegraphics[width=\textwidth]{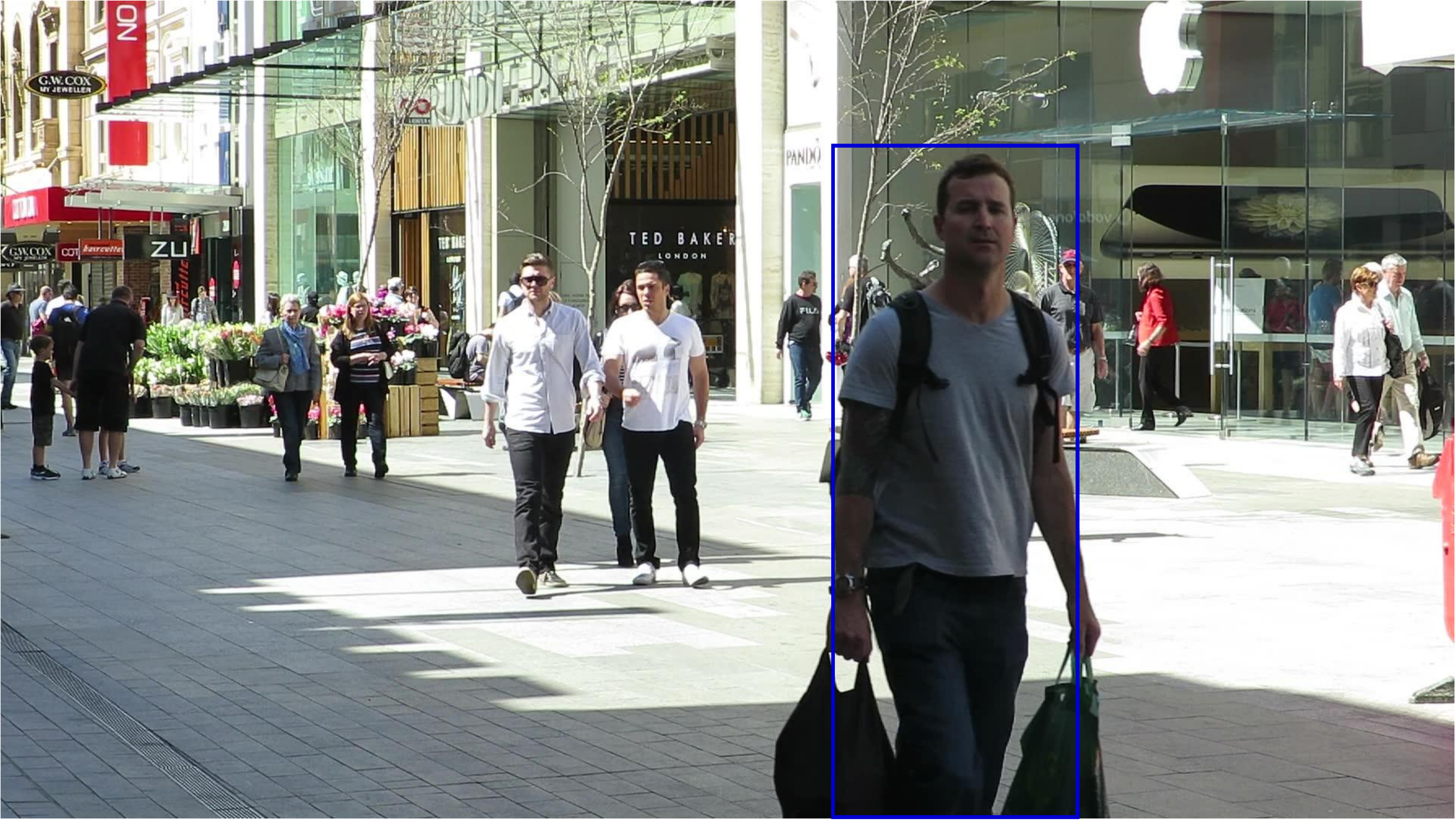}
        \caption{MOTDT output during occlusion}
    \end{subfigure}
    \begin{subfigure}[t]{0.33\textwidth}
        \includegraphics[width=\textwidth]{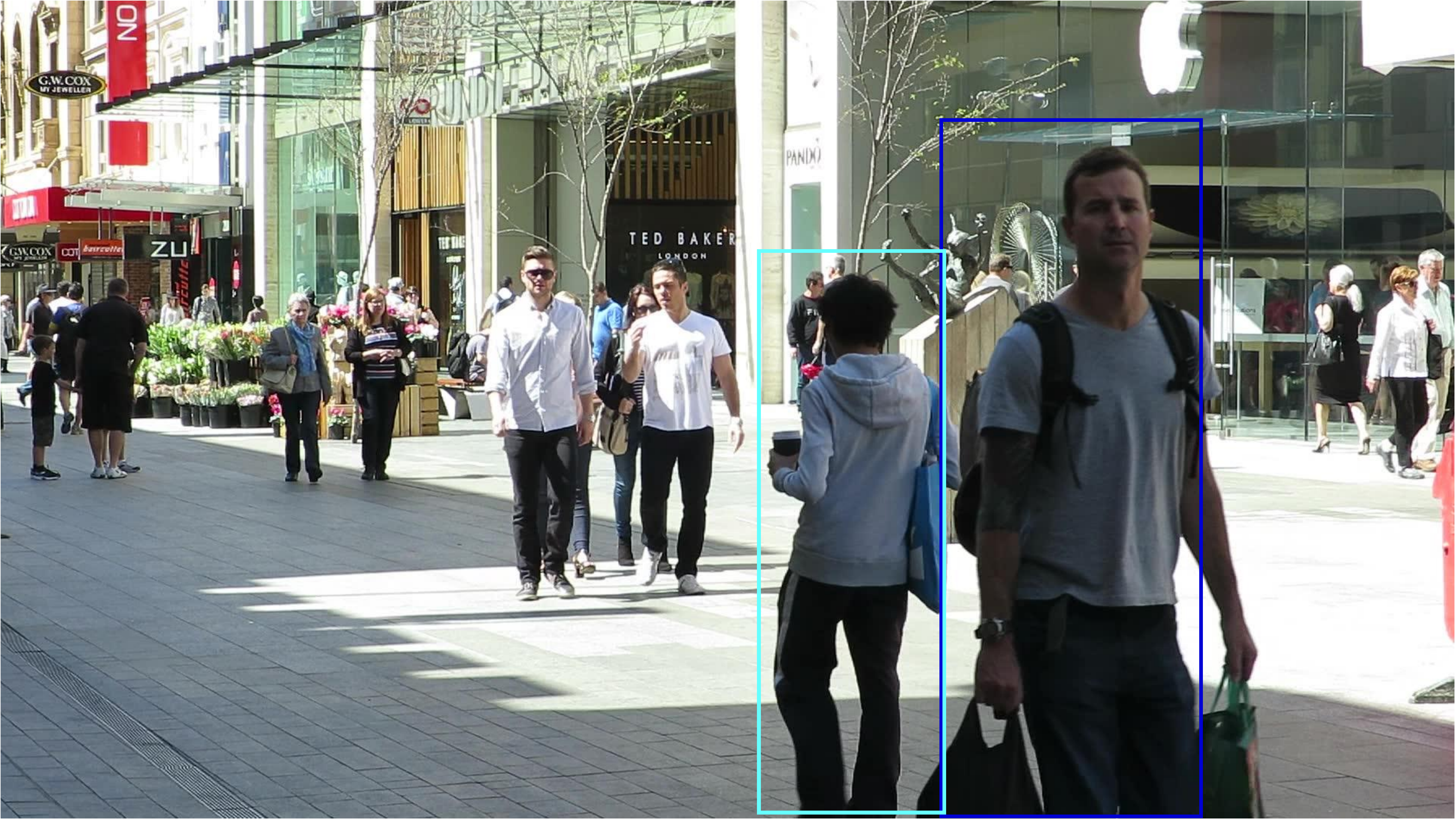}
        \caption{MOTDT output after occlusion}
    \end{subfigure}
    \begin{subfigure}[t]{0.33\textwidth}
        \includegraphics[width=\textwidth]{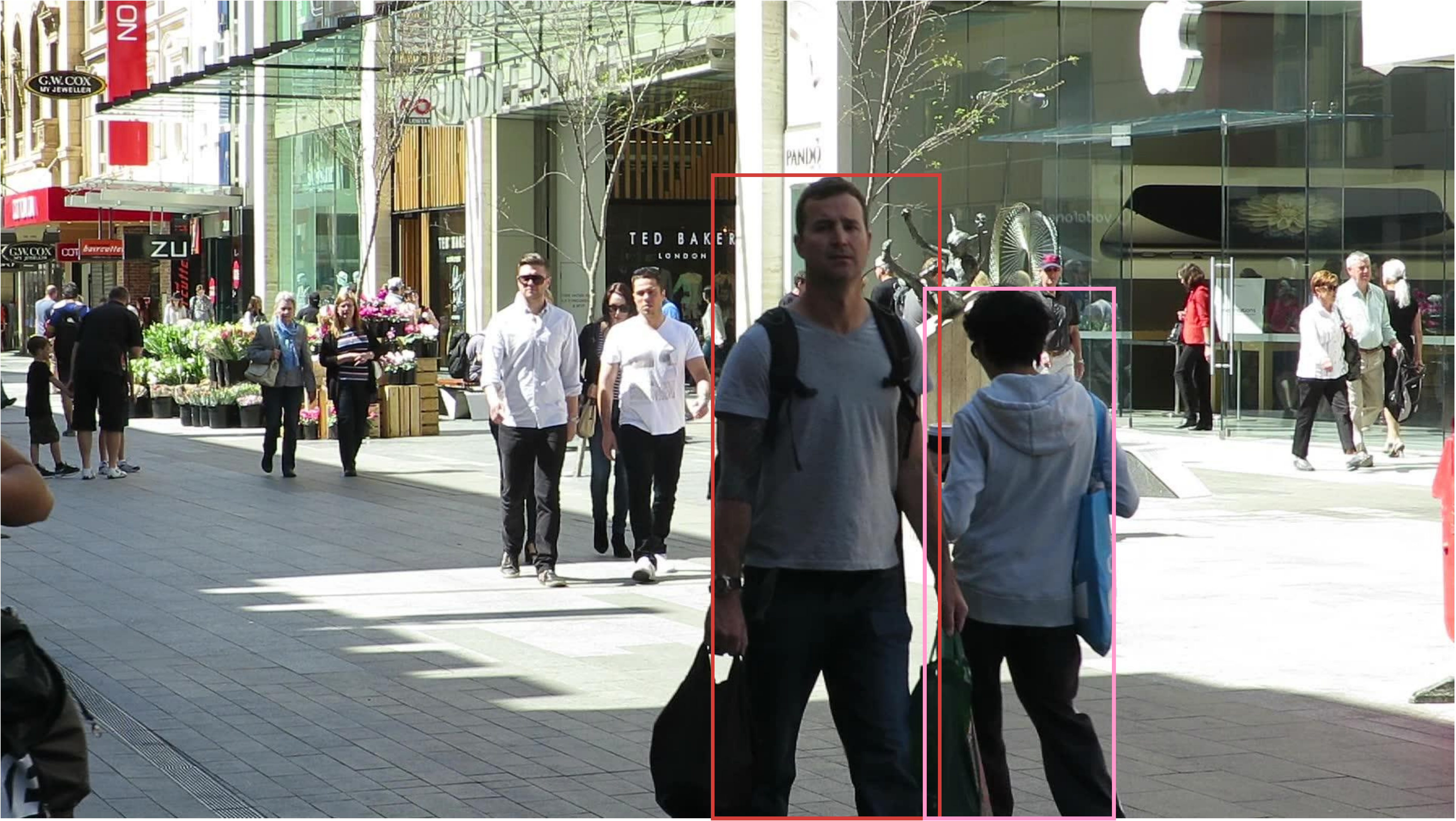}
        \caption{eHAF16 output before occlusion}
    \end{subfigure}
    \hfill
    \begin{subfigure}[t]{0.33\textwidth}
        \includegraphics[width=\textwidth]{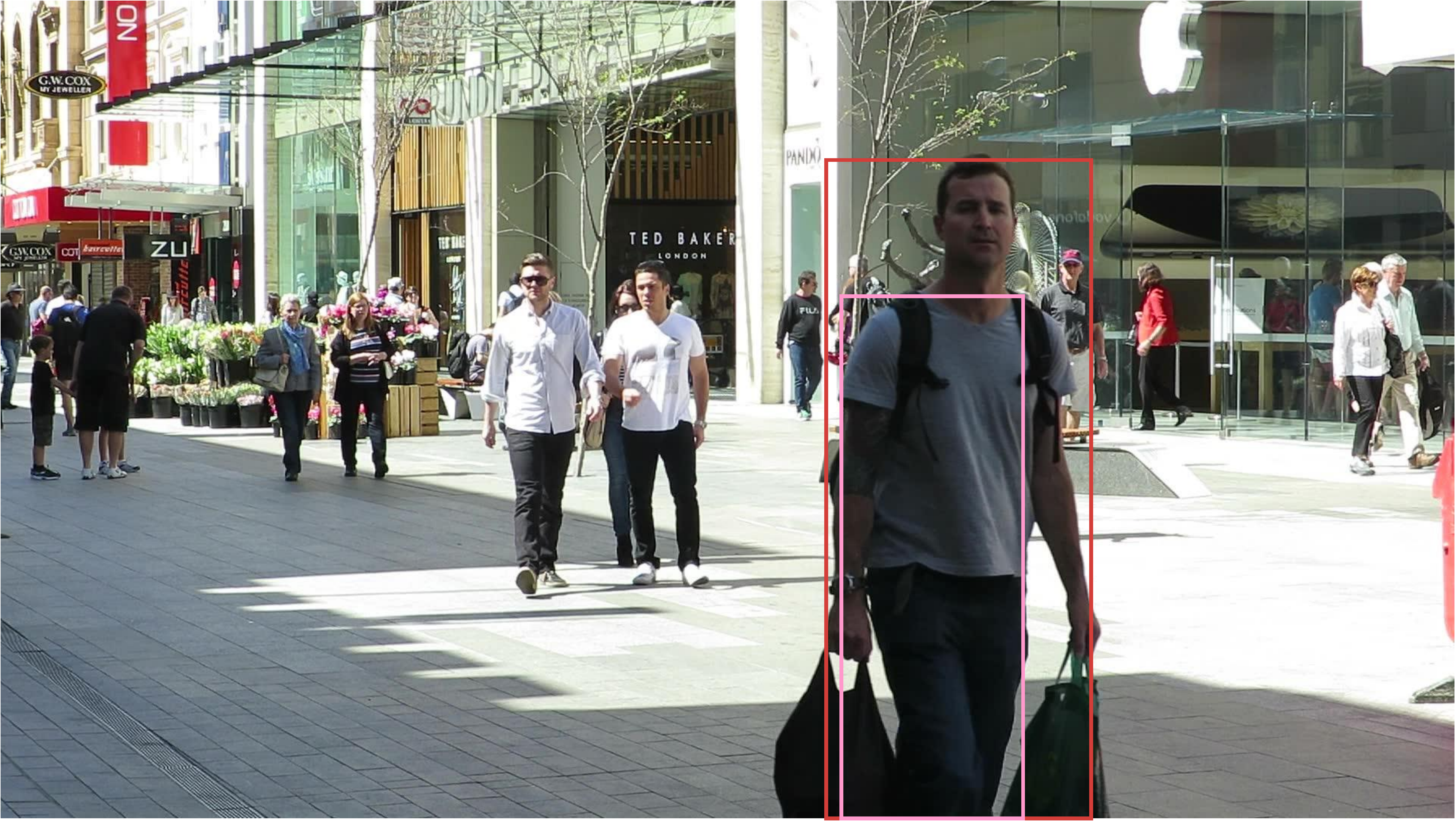}
        \caption{eHAF16 output during occlusion}
    \end{subfigure}
    \begin{subfigure}[t]{0.33\textwidth}
        \includegraphics[width=\textwidth]{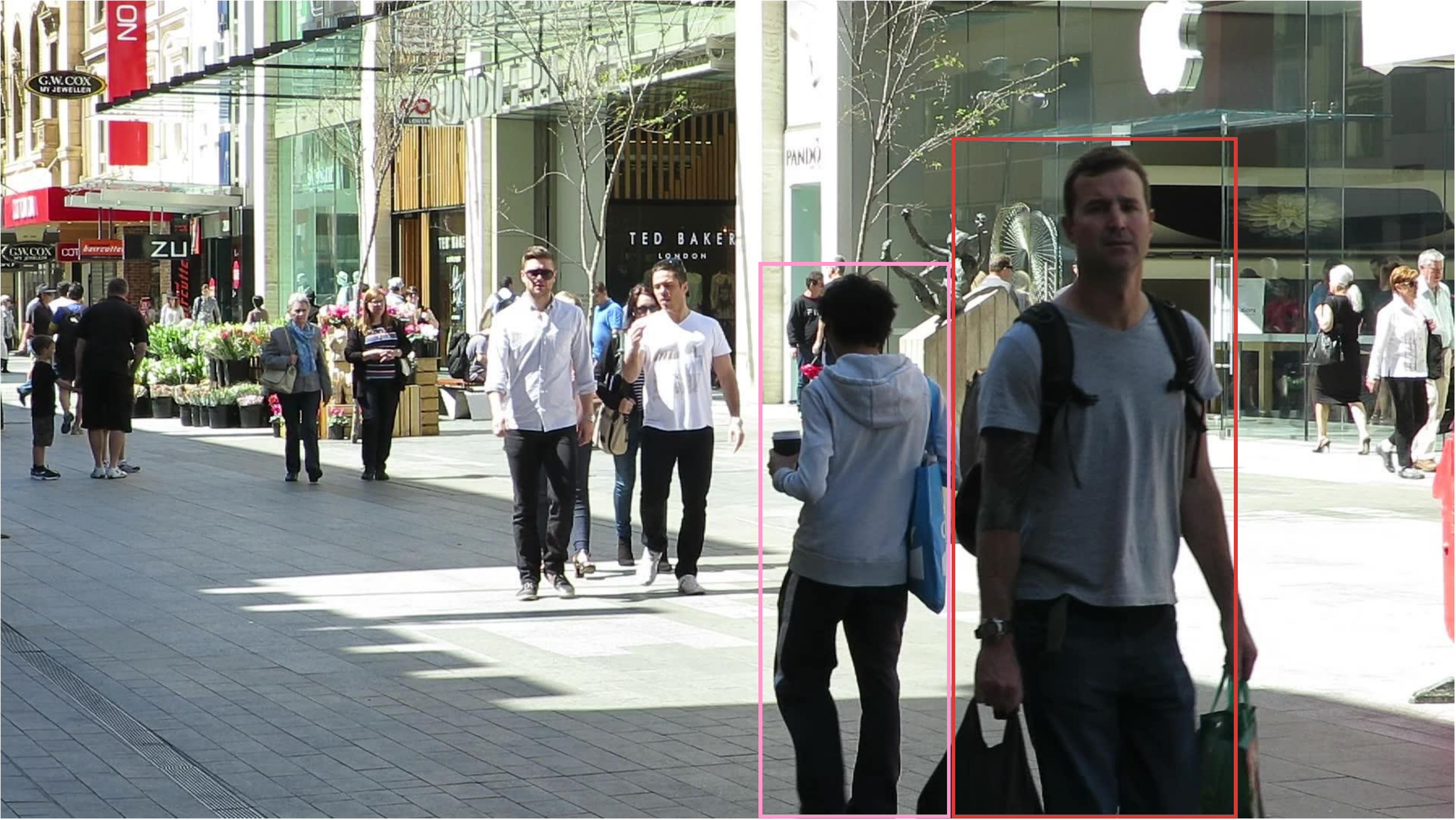}
        \caption{eHAF16 output after occlusion}
    \end{subfigure}
    \caption{Example of fragmentation produced by an online method during occlusion. Above: tracking results for MOTDT \cite{long2018tracking}, online algorithm. Below: tracking results for eHAF16 \cite{sheng2018heterogeneous}, batch algorithm. From left to right, frames 50, 60 and 70 of the MOT16-08 video are shown for both methods. Only the relevant boxes are shown to avoid clutter. As we can see in the image, while some online algorithms are able to re-identify lost targets after occlusions, they are usually unable to track them while the target is not visible, and this results in a fragmentation. Batch methods, on the other hand, are capable of reconstructing a fragmented trajectory by inferring the position of the target given past and future information.}
    \label{fig:frag-example}
\end{figure}

As expected, the best performing algorithms on each dataset use private detections, confirming the fact that the detection quality dominates the overall performance of the tracker: 56.5\% MOTA vs. 42.3\% for MOT15 and 71.0\% vs. 49.3\% for MOT16. Moreover, on MOT16 and MOT17 the batch algorithms slightly outperform the online ones, even though the online methods are progressively getting closer to the performance of the batch ones. In fact, on MOT15 the best reported algorithm that uses deep learning runs in an online fashion. However, this can be an effect of a greater focus on developing online methods, which is a trend in the MOT deep learning research community. A common problem among online methods, that is not reflected in the MOTA score, is the higher number of fragmentations, as we can see in table \ref{tab:frag}. This happens because, when occlusions occur or detections are missing, online algorithms cannot look ahead in the video, re-identify the lost targets and interpolate the missing part of the trajectories \cite{chu2017online, yang2017hybrid, bae2017confidence}. We can see in figure \ref{fig:frag-example} an example of trajectory that is fragmented by an online method, MOTDT \cite{long2018tracking}, while it is correctly tracked by a batch method, eHAF16 \cite{sheng2018heterogeneous}.

\begin{figure}
    \centering
    \begin{subfigure}[t]{0.48\textwidth}
        \includegraphics[width=\textwidth]{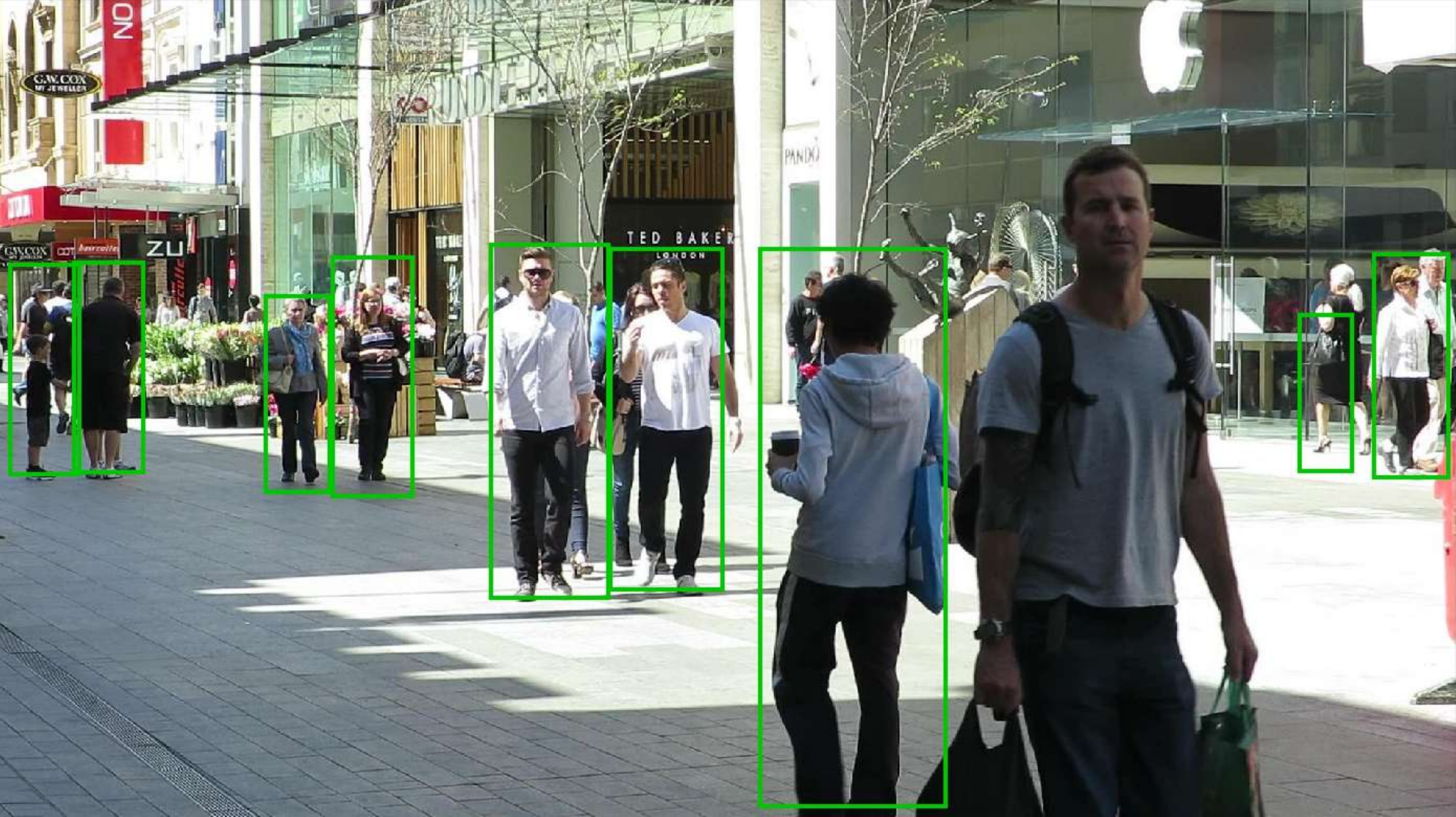}
        \caption{Public detections}
    \end{subfigure}
    \hfill
    \begin{subfigure}[t]{0.48\textwidth}
        \includegraphics[width=\textwidth]{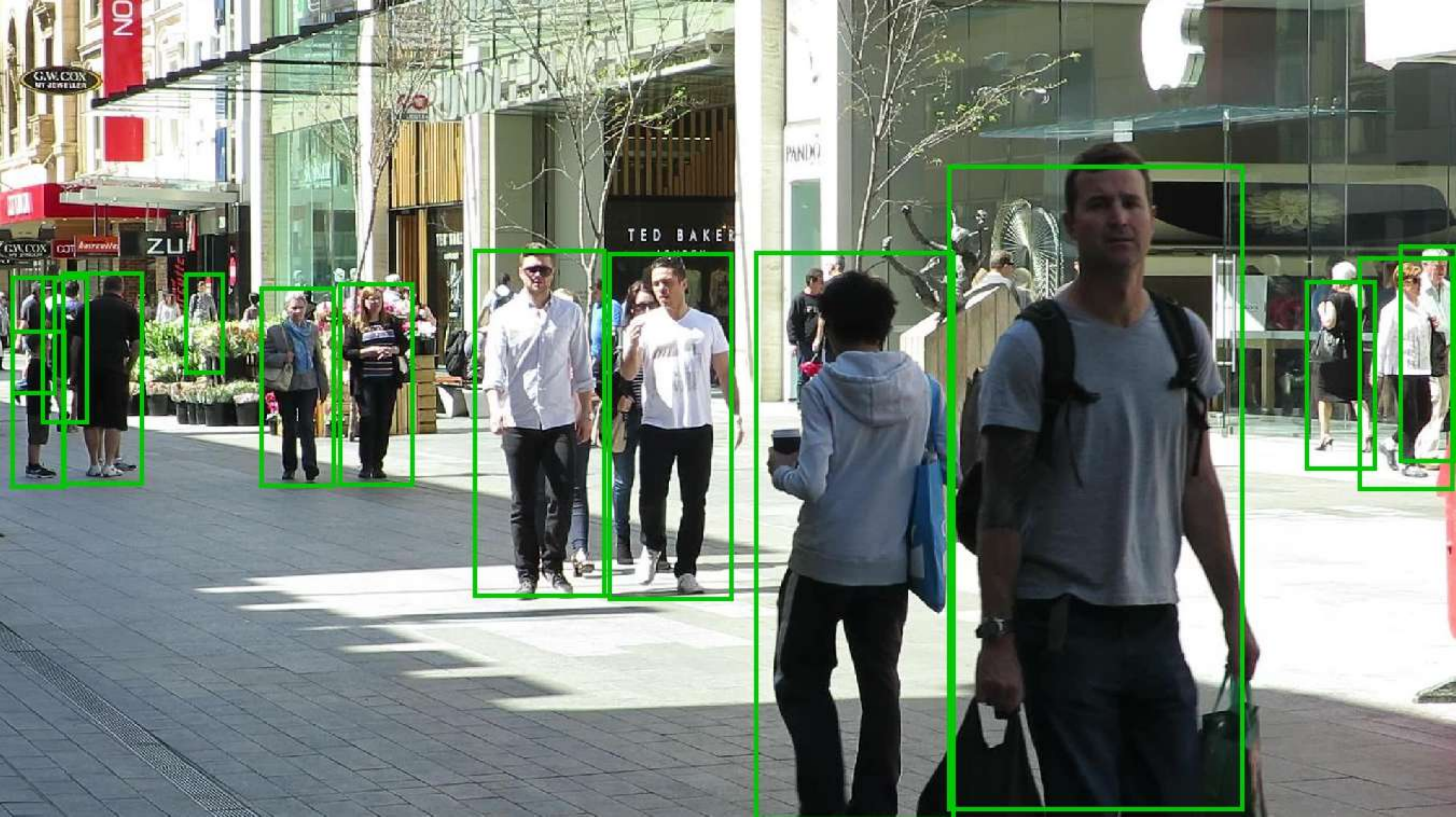}
        \caption{Detections from \cite{yu2016poi}}
    \end{subfigure}
    \begin{subfigure}[t]{0.48\textwidth}
        \includegraphics[width=\textwidth]{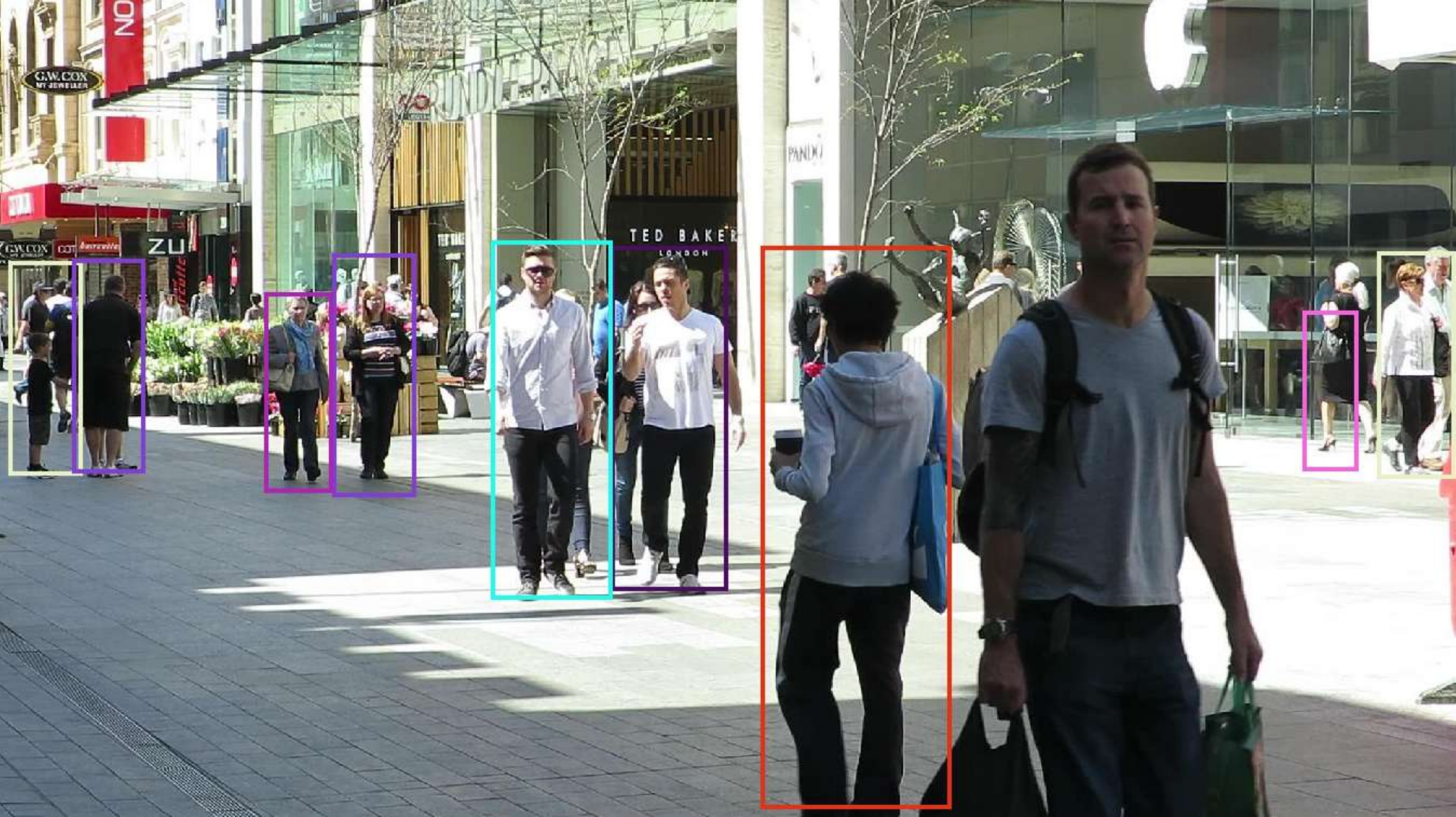}
        \caption{Output tracks with public detections}
    \end{subfigure}
    \hfill
    \begin{subfigure}[t]{0.48\textwidth}
        \includegraphics[width=\textwidth]{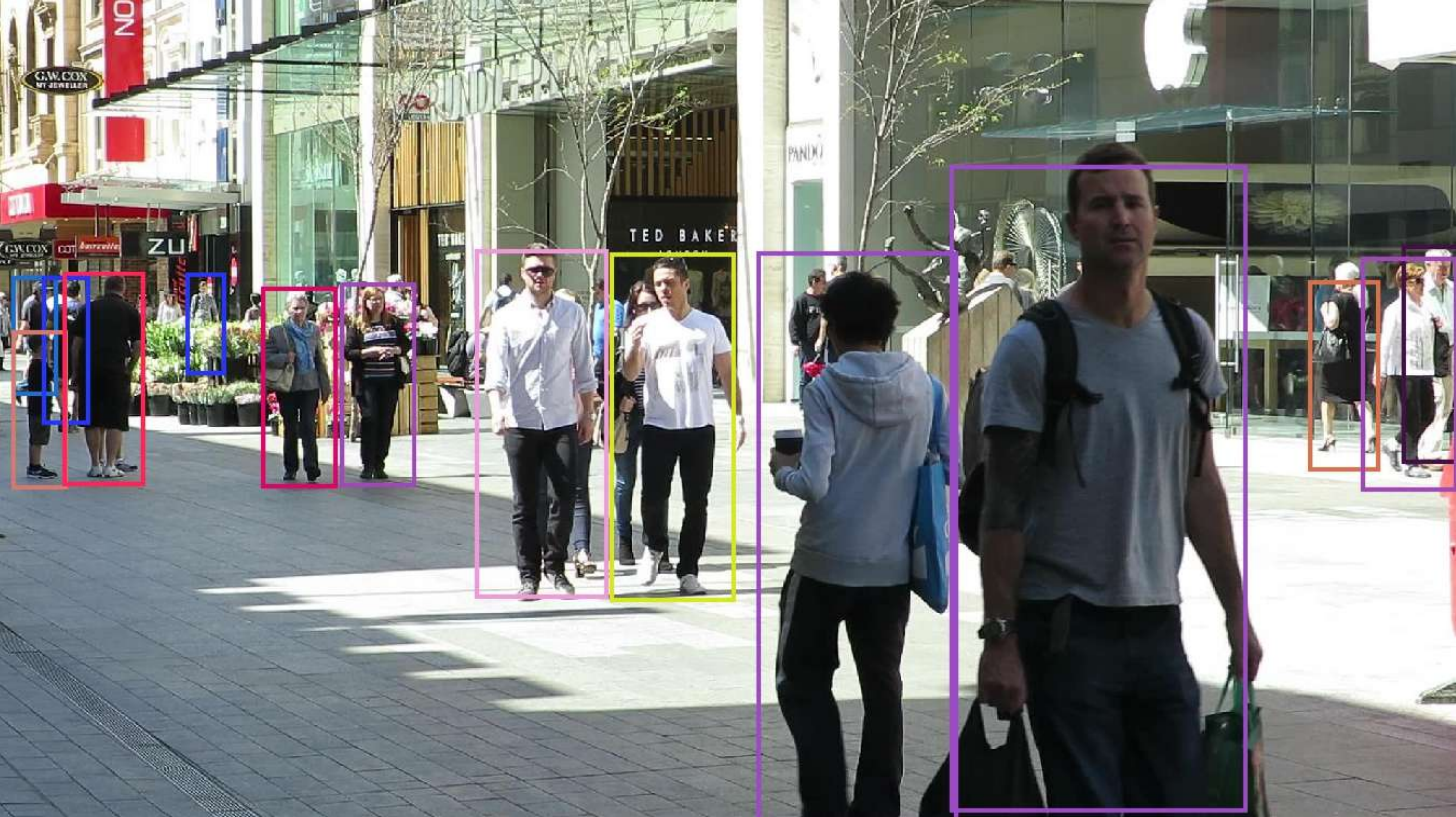}
        \caption{Output tracks with detections from \cite{yu2016poi}}
    \end{subfigure}
    \caption{Above: Public detections (generated using DPM v5 \cite{girshick2012DPMv5}) and private detections (obtained by \cite{yu2016poi} with a customized Faster R-CNN trained on multiple datasets) for frame 70 of MOT16-08 sequence. It can be observed that the man in the foreground is correctly detected by the custom Faster R-CNN detector (b), while it is ignored by DPM (a). Below: Results of tracking for the two detection sets using the SORT \cite{bewley2016simple} algorithm, whose performance is heavily dependent on detection errors. We can indeed see that the mentioned missing detection produces a corresponding false negative in the tracking output (c), while in (d) the man is correctly tracked.}
    \label{fig:pub_priv_dets}
\end{figure}

Another interesting thing to notice is that since the MOTA score is basically a normalized sum of FPs, FNs and ID switches, and since the number of FNs is usually at least an order of magnitude higher than the FPs and two order of magnitudes higher than the ID switches, the methods that manage to strongly reduce the number of FNs are the ones that obtain the best performances. We can in fact observe a strong correlation between MOTA and number of FNs, in accordance to what was found in \cite{leal2017tracking}: MOTA and FN values are linked by a Pearson correlation coefficient of $-0.95$ on MOT15, $-0.98$ on MOT16 and $-0.95$ on MOT17. So, while there have been limited improvements in the reduction of FNs using public detections, the most effective way is still building and training a custom detector; the halving of the number of FNs is in fact the main reason why private detectors have lead to better tracking performances, being able to identify previously uncovered targets. In figure \ref{fig:pub_priv_dets} we can see how the SORT algorithm, that is particularly sensitive to missing detections, is not able to detect a target as soon as the relative detection is missing.

 To avoid this issue, many new algorithms are including new strategies to tackle this problem. In fact, while basic approaches that perform interpolation are able to recover missing boxes during occlusions, this is still insufficient to detect targets that are not covered by even a single detection, that have been shown to be 18\% of the total on MOT15 and MOT16 \cite{leal2017tracking}. For example, the eHAF16 algorithm presented by Sheng et al. \cite{sheng2018heterogeneous} employed a superpixel extraction algorithm to complement the publicly provided detections and was in fact able to significantly reduce the number of false negatives on MOT17, reaching top MOTA score on the dataset. The MOTDT algorithm \cite{long2018tracking} instead used a R-FCN to integrate the missing detections with new candidates, and was able to reach best MOTA and lowest number of false negatives among online algorithms on MOT17. The AP-RCNN algorithm \cite{chen2017online} was also able to avoid the problems caused by missing detections by employing a Particle Filter algorithm and relying on detections only to initialize new targets and to recover lost ones. The algorithm presented in \cite{ren2018collaborative} also reduces FNs by designing a deep prediction network, whose aim is to learn the motion model of the objects. At test time, the network is capable to predict the position of them in subsequent frames, and thus reducing the amount of false negatives produced by missing detections. In fact, it is the second best among online methods in MOT16 regarding this metric.

\begin{table}[htb]
\centering
\begin{tabular}{lrrr}
\toprule
Mode   &  MOT15  &  MOT16  &  MOT17  \\
\midrule
Batch  &  1143.8 &  1104.9 &  3188.2 \\
Online &  1509.5 &  1820.2 &  7555.8 \\
\bottomrule
\end{tabular}

\caption{Average number of fragmentations for online and batch methods in the three considered datasets.}
\label{tab:frag}
\end{table}

An important observation must be made regarding the training strategy for affinity networks. As noted by \cite{kim2018multi}, training a network using ground truth trajectories to predict affinities might produce suboptimal results, as at test time those networks would be exposed to a different data distribution, made of noisy trajectories that can include missing/wrong detections. Many algorithms in fact have chosen to train networks using either actual detections \cite{fang2018recurrent} or by manually adding noise and errors to the ground truth trajectories \cite{kim2018multi, zhu2018online}, although this may slow the training procedure sometimes and not always be feasible \cite{kieritz2018joint}.

\subsubsection*{Best approaches in the four MOT steps}
Speaking of private detections, the tables show that the best performing detectors are currently Faster R-CNN and its variants. In fact, the algorithm presented in \cite{yu2016poi}, that uses a modified Faster R-CNN, has held its top ranking position among online methods on MOT16 for 3 years, and many of the other top-performing MOT16 algorithms have employed the same detections. In contrast, algorithms that employed the SSD detector, such as the ones presented in \cite{kieritz2018joint} and \cite{zhao2018multi}, tend to perform worse. A big advantage of SSD, though, is its faster speed: thanks to that the algorithm by Kieritz et al. \cite{kieritz2018joint} was able to reach near real-time performance (4.5 FPS) \textit{including} the detection step\footnote{We remind the reader that the FPS reported by many algorithms tend to exclude the detection step, that can easily be the most computationally expensive part of the algorithm.}. Despite the great number of online methods, a major issue in using deep learning techniques in a MOT pipeline is still the difficulty in obtaining real-time predictions, making such algorithms not usable in most practical online scenarios.

Regarding feature extraction, all the top performing methods on the three considered datasets have used a CNN to extract appearance features, where GoogLeNet is the most common one. Methods that do not exploit appearance (either extracted with deep or classical methods) tend to perform worse. However, visual features are not enough: many of the best algorithms also employ other types of features to compute affinity, especially motion ones. In fact, algorithms like LSTMs and Kalman Filters are often employed to predict the position of the target in the next frame and this often helps in improving the quality of the association. Various Bayesian filter algorithms, such as particle filter and hypotheses density filter, are also used to predict target motion, and they benefit from the use of deep models \cite{xiang2019online, chen2017online, fu2018gm}. Nonetheless, even when employed together with non-visual features, appearance still plays a major role in improving the overall performance of the algorithm \cite{sadeghian2017tracking, xiang2019online}, especially in avoiding ID switches \cite{kim2015multiple} or to re-identify targets after long occlusions \cite{wojke2017simple}. In the latter case, simple motion predictors do not work since the linear motion assumption is easily broken, as noted by Zhou et al. \cite{zhou2018online}.

While deep learning plays an important role in detection and feature extraction, the use of deep networks to learn affinity functions is less ubiquitous and has not yet been proven to be essential for a good MOT algorithm. Many algorithms in fact rely on a combination of hand-crafted distance metrics on a variety of deep and non-deep features. However, some works have already demonstrated how using affinity networks can produce top-performing algorithms \cite{ma2018customized, chen2017online, tang2017multiple, fang2018recurrent}, with approaches ranging from the use of Siamese CNNs to recurrent neural networks. In particular, the adapted Siamese network proposed by Ma et al. \cite{ma2018customized} was able to produce reliable similarity measures that helped with the person re-identification after occlusions and allowed the algorithm to reach the highest MOTA score on MOT16. The integration of body part information was also crucial for the StackNetPose CNN proposed by Tang et al. \cite{tang2017multiple}: it served as an attention mechanism that allowed the network to focus on the relevant parts of the input images, thus producing more accurate similarity measures. The algorithm was able to reach top performance on MOT16 using private detections.

Very few works have instead explored the use deep learning models to guide the association process, and this could be a interesting research direction for future approaches.

\subsubsection*{Other trends in top-performing algorithms}
\begin{figure}
    \centering
    \begin{subfigure}[t]{0.2\textwidth}
        \includegraphics[width=\textwidth]{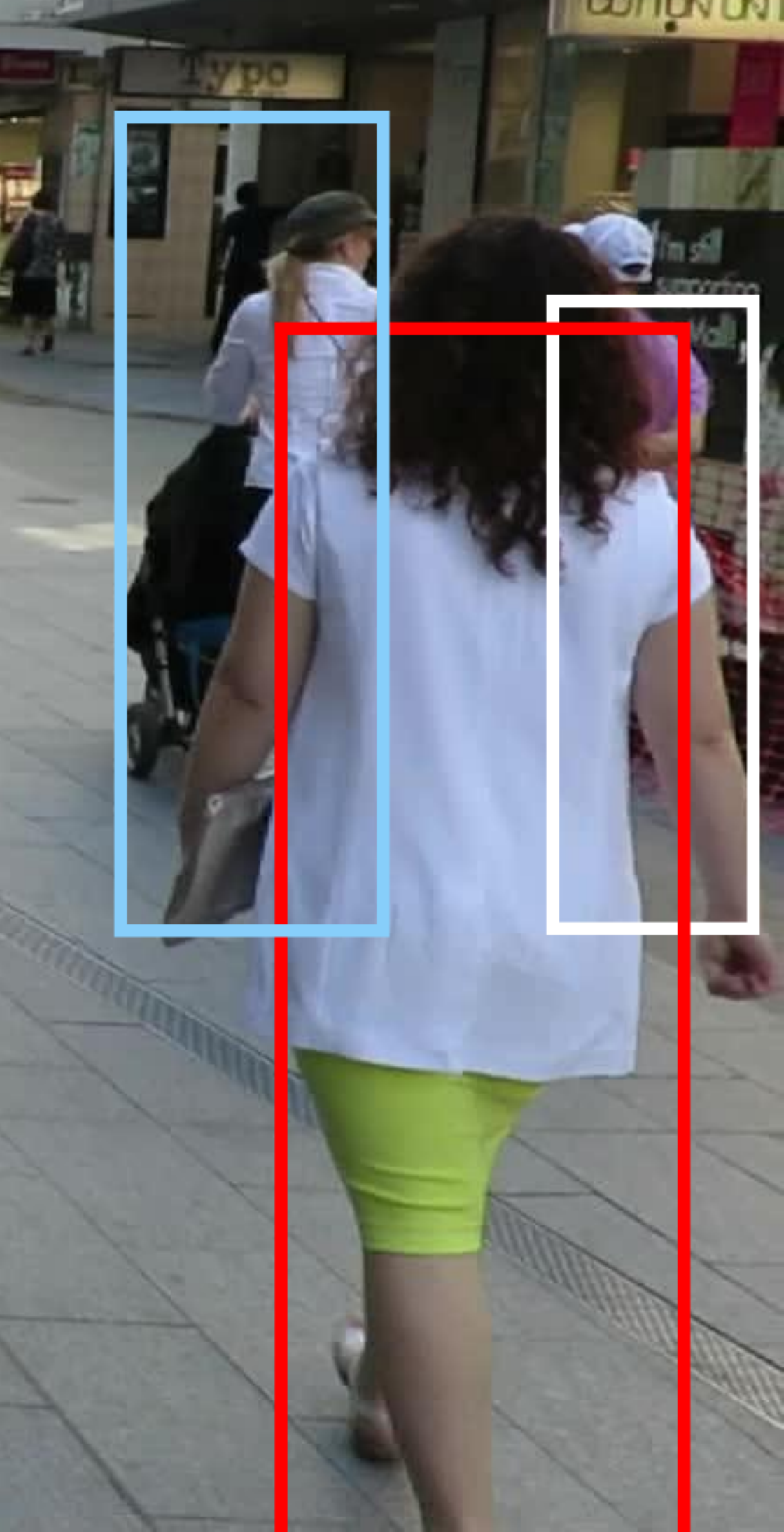}
        \caption{Frame 14}
    \end{subfigure}
    \begin{subfigure}[t]{0.2\textwidth}
        \includegraphics[width=\textwidth]{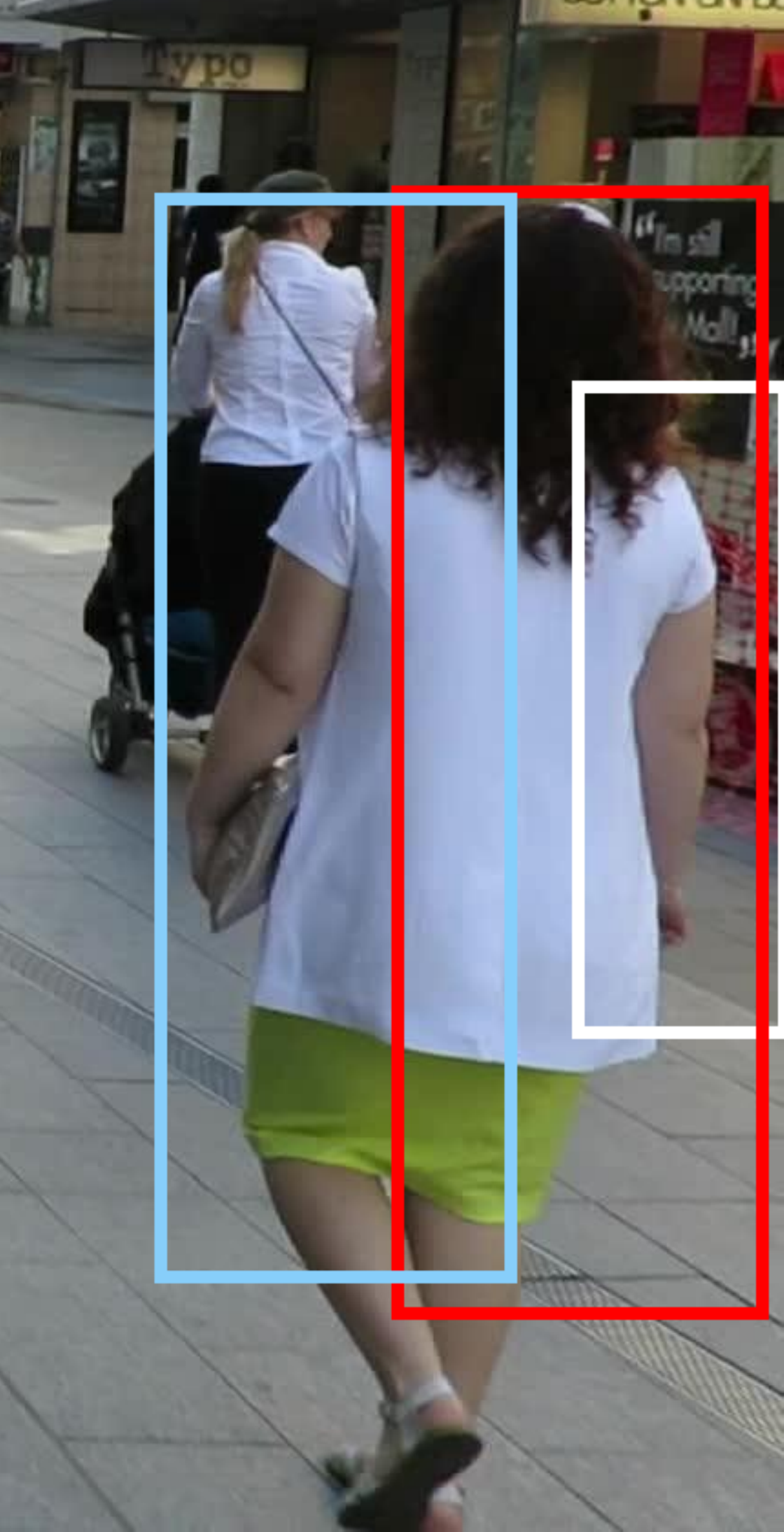}
        \caption{Frame 22}
    \end{subfigure}
    \begin{subfigure}[t]{0.2\textwidth}
        \includegraphics[width=\textwidth]{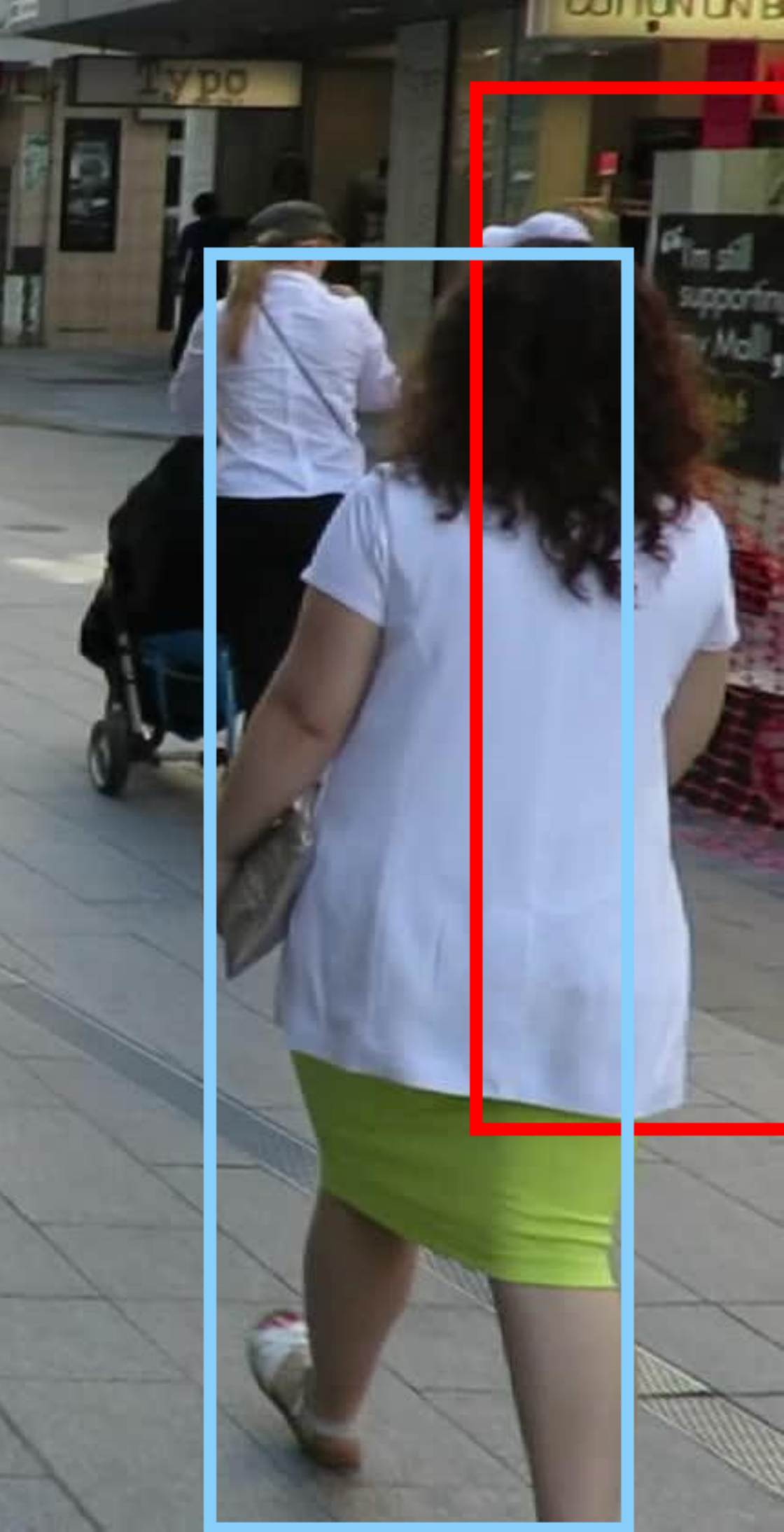}
        \caption{Frame 28}
    \end{subfigure}
    \begin{subfigure}[t]{0.2\textwidth}
        \includegraphics[width=\textwidth]{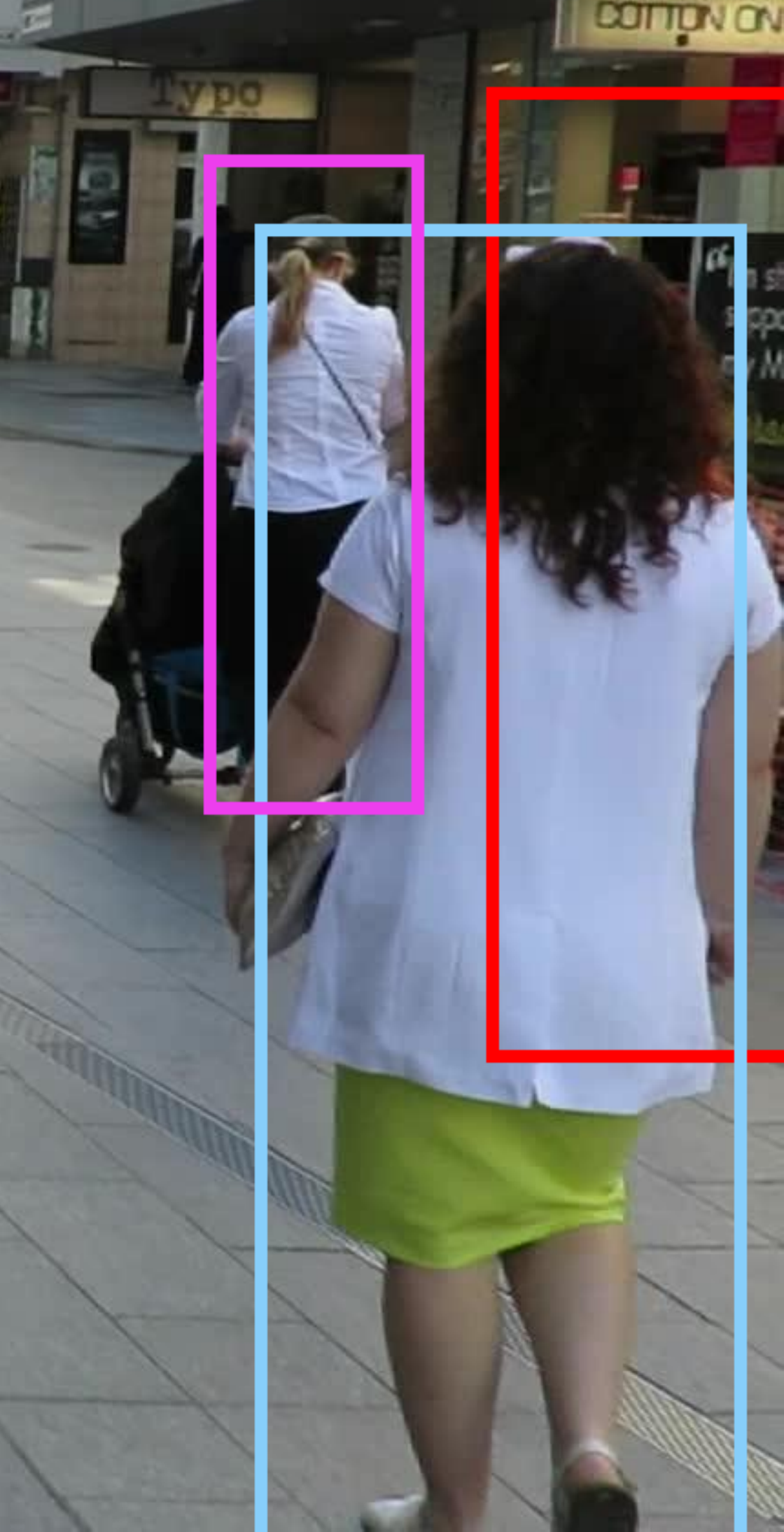}
        \caption{Frame 34}
    \end{subfigure}
    \caption{Example of SOT drift in the context of a MOT algorithm (KFC16 \cite{chu2019online}). The four images are cropped from the MOT16-07 video, and are best viewed in color, since each color represents a different target ID. At first (a) the three persons are tracked. After a few frames (b) the \textit{red} box starts drifting towards the occluded man, while the \textit{light blue} box starts drifting towards the foreground woman. In (c) the white track is interrupted and the first two ID switches are completed. In (d) a new identity is assigned to the woman in the background, causing a third ID switch.}
    \label{fig:SOT_drift}
\end{figure}

Some other trends can be identified among current top ranked methods. For example, a successful approach in online methods is the use of Single Object Tracking algorithms, properly modified in order to solve the MOT task. Some of the top performing online algorithms on the 3 datasets have in fact employed a SOT tracker augmented with deep learning techniques to recover from occlusions or to refresh the target models \cite{chu2019online, zhu2018online, sadeghian2017tracking}. Interestingly, to the best of our knowledge, no adapted SOT algorithm has been used to perform tracking with private detections. As we have already observed, the use of private detections reduces the number of completely uncovered targets; since SOT trackers don't usually need detections to keep following a target once it's been identified, the reduction in uncovered targets might translate in a much lower number of lost tracks, that in turn would enhance the overall performance of tracker. The application of a SOT tracker on private detections could thus be a good research direction to try to further improve the results on the MOTChallenge datasets. A batch method could also exploit a SOT tracker to look at past frames in order to recover missed detections before the target was first identified by the detector. However, SOT-based MOT trackers can sometimes still be prone to tracking drift and produce a higher number of ID switches. For example, the KCF16 algorithm \cite{chu2019online}, while reaching top MOTA score among online methods on MOT16 on public detections, it still produces a relatively high number of switches due to tracker drift, as can be seen in figure \ref{fig:SOT_drift}. Moreover, SOT-based MOT algorithms must be careful not to keep tracking spurious trajectories, caused by the inevitably higher number of false positive detections predicted by higher-quality detectors, for too many frames, as this might offset the reduction in the number of FNs. Current approaches \cite{chu2019online, zhu2018online} still tend to use detection overlap (e.g. in how many recent frames the trajectory is covered by a detection) to understand if a trajectory is a true or a false positive in the long run, but better solutions should be investigated to avoid exclusive reliance on detections.

While many methods perform association by formulating the task as a graph optimization problem, batch methods benefit in particular from this, since they can perform global optimization on them. For example, the minimum cost lifted multicut problem has reached top performance on MOT16, helped by CNN-computed affinities \cite{ma2018customized, tang2017multiple}, while heterogeneous association graph fusion and correlation clustering are used on the two top MOT17 methods \cite{sheng2018heterogeneous, henschel2018fusion}.

Finally, it can be noticed that the accuracy of bounding boxes radically affects the final performance of the algorithms. In fact, the top ranked tracker on MOT15 \cite{yang2017hybrid} obtained a relatively high MOTA score by just performing a bounding box regression step on the output of a previous state-of-the-art algorithm \cite{yang2017hybrid} using a deep RL agent. Developing an effective bounding box regressor to be incorporated in future MOT algorithms could be an interesting research direction that has not yet been explored thoroughly. Moreover, instead of relying on a single frame to fix the boxes, that could make them enclose the wrong target in case of an occlusion, batch methods could also try to exploit future and past target appearance to more accurately regress the bounding boxes around the right target.

\section{Conclusion and future directions}\label{sec:conclusion}
We have presented a comprehensive description of all MOT algorithms employing deep learning techniques, focusing on single-camera videos and 2D data. Four main steps have been shown to characterize a generic MOT pipeline: detection, feature extraction, affinity computation, association. The use of deep learning in each of these four steps has been explored. While most of the approaches have focused on the first two, some applications of deep learning to learn affinity functions are also present, but only very few approaches use deep learning to directly guide the association algorithm. A numerical comparison of the results on the MOTChallenge datasets has also been provided, showing that, despite the wide variety of approaches, some commonalities can be found among the presented methods:
\begin{itemize}
    \item \textbf{detection quality is important}: the amount of false negatives still dominates the MOTA score. While deep learning has allowed for some improvement in this regard for algorithms employing public detections, the use of higher quality detections is still the most effective way to reduce false negatives. Thus, a careful use of deep learning in the detection step can considerably improve the performance of a tracking algorithm;
    \item \textbf{CNNs are essential in feature extraction}: the use of appearance features is also fundamental for a good tracker and CNNs are particularly effective at extracting them. Moreover, strong trackers tend to use them in conjunction with motion features, that can be computed using LSTMs, Kalman filter or other Bayesian filters;
    \item \textbf{SOT trackers and global graph optimization work}: the adaptation of SOT trackers to the MOT task, with the help of deep learning, has recently produced good-performing online trackers; batch methods have instead benefited from the integration of deep models in global graph optimization algorithms.
\end{itemize}

As deep learning has been introduced only recently in the field of MOT, a number of promising future research directions have also been identified:
\begin{itemize}
    \item \textbf{researching more strategies to mitigate detection errors}: although modern detectors are constantly reaching better and better performances, they are still prone to produce a significant number of false negatives and false positives in complex scenarios such as dense pedestrian tracking. Some algorithms have provided solutions to reduce the exclusive reliance on detections by integrating them with information extracted from other sources (e.g. superpixels \cite{sheng2018heterogeneous}, R-FCN \cite{long2018tracking}, Particle Filter \cite{chen2017online}, etc.), but further strategies should be investigated;
    \item \textbf{applying DL to track different targets}: most of DL-based MOT algorithms have focused on pedestrian tracking. Since different types of targets pose different challenges, possible improvements in tracking vehicles, animals, or other objects with the use of deep networks should be investigated;
    \item \textbf{investigating the robustness of current algorithms}: how do current methods perform under different camera conditions? How do a varying contrast, illumination, the presence of noisy/missing frames affect the result of current algorithms? Are existing DL networks able to generalize to different tracking contexts? For example, the vast majority of people tracking frameworks are trained to follow pedestrians or athletes, but tracking could be useful in other scenarios. A possible new application could be helping with scene understanding in different contexts: inside movies, in order to generate textual descriptions to provide a coarse way of searching for a scene in a movie; or on social networks, in order to generate descriptions for blind users or to detect inappropriate videos that should be removed from the platform. These different scenarios would probably require changes to the current detection and tracking algorithms, since the people could appear in unusual poses and behaviors that are not present in the existing datasets for MOT;
    \item \textbf{applying DL to guide association}: the use of deep learning to guide the association algorithm and to directly perform tracking is still in its infancy: more research is needed in this direction to understand if deep algorithms can be useful in this step too;
    \item \textbf{combining SOT trackers with private detections}: a possible way to reduce the number of lost tracks, and thus reduce the false negatives, could be the combination of SOT trackers with private detections, especially in a batch setting, where it would be possible to recover past detections that were previously missed;
    \item \textbf{investigating bounding box regression}: the use of bounding box regression has been shown to be a promising step in obtaining a higher MOTA score, but this has not yet been explored in detail and further improvements should be investigated, e.g. the use of past and future information to guide the regression;
    \item \textbf{investigating post-tracking processing}: in batch contexts, it is possible to apply correction algorithms on the output of a tracker to increase its performance. This has already been shown by Babaee et al. \cite{babaee2018occlusion}, that have applied occlusion handling on top of existing algorithms, and by Jiang et al. \cite{jiang2018precise} with the aforementioned bounding box regression step. More complex processing could be applied on the results from a tracker to further improve the results.
\end{itemize}

Finally, as very few of the presented algorithms have provided public access to their source code, we would like to encourage future researchers to publish their code in order to allow for better reproducibility of their results and benefit the whole research community.

\section*{Acknowledgements}
This research work is partially supported by the Spanish Ministry of
Science and Technology under the project TIN2017-89517-P and the project
DeepSCOP-Ayudas Fundaci\'on BBVA a Equipos de Investigaci\'on
Cient\'ifica en Big Data 2018. Siham Tabik was supported by the Ramon y
Cajal Programme (RYC-2015-18136).

\bibliographystyle{unsrt}  
\bibliography{references}

\newpage
\appendix

\section{Appendix}
\label{app:summary}
Here we present a table containing a summary of the techniques used by each algorithm presented in this paper. The table follows the order of presentation of the papers. Since we think that the publication of open source code can greatly help the research community, we have also provided links to the source codes for the papers that provide them.
\renewcommand{\arraystretch}{1.5}
\begin{longtable}{c>{\centering\arraybackslash}p{0.12\textwidth}>{\centering\arraybackslash}p{0.18\textwidth}>{\centering\arraybackslash}p{0.18\textwidth}>{\centering\arraybackslash}p{0.18\textwidth}c>{\centering\arraybackslash}p{0.08\textwidth}}
\toprule
 & Detection & Feature extr. / mot. pred. & Affinity / cost computation & Association / Tracking & Mode & Source and data \\ \midrule \endhead
 \hline
 \endfoot
 \endlastfoot
\cite{bewley2016simple} & Faster R-CNN & Kalman filter & IoU & Hungarian algorithm & O & \href{https://github.com/abewley/sort}{Source} \\
\cite{yu2016poi} & Modified Faster R-CNN & Modified GoogLeNet, Kalman filter & Cosine distance + IoU & Hungarian algorithm (online), modified H2T \cite{wen2014multiple} (batch) & O+B & \href{https://drive.google.com/file/d/0B5ACiy41McAHMjczS2p0dFg3emM/view}{Detections and appearance features} \\
\cite{ran2019robust} & Faster R-CNN & CNN (app.), AlphaPose CNN, pose joints velocities, interaction grid & Pose-based Triple Stream Network (LSTM-based) & Custom algorithm & O & \\
\cite{hu2018automatic} & Faster R-CNN & CNN & Euclidean distance, cosine distance & Multifeature fusion re-tracking algorithm & B & \\
\cite{zhang2019automatic} & CNN & HOG + Colour Names & Variation of Discriminative Correlation Filter & Custom algorithm + Hungarian algorithm & O & \\
\cite{lu2017online} & SSD & SSD, LSTM & Cosine similarity & Hungarian algorithm & O & \\
\cite{kieritz2018joint} & SSD & SSD & RNN & Hungarian algorithm, MLP (track scores) & O & \\
\cite{zhao2018multi} & SSD & SSD + Correlation Filter & IoU + APCE & Hungarian algorithm & O & \\
\cite{zhou2018online} & Public / Mask R-CNN & Siamese Mask R-CNN & App. affinity, mot. consistency, spatial structural potential & Tensor-based high-order graph matching & O & Code will be released \\
\cite{kim2018online} & YOLOv2 & Tiny Yolo, Particle filter, Random Ferns, KLT & Pairwise overlap ratio, student Random Ferns, Euclidean distance & Greedy bipartite assignment & O & \\
\cite{sharma2018beyond} & RRC or SubCNN & Feature-based odometry, Pose Adjustment CNN, stacked-hourglass CNN & 3D-2D cost + 3D-3D cost + appearance, shape and pose costs & Hungarian algorithm & O & \href{https://github.com/JunaidCS032/MOTBeyondPixels}{Source} \\
\cite{pernici2018memory} & DPM or Tiny (CNN) & DPM or Tiny (CNN) & Implicit in Reverse Nearest Neighbour & Reverse Nearest Neighbour Matching & O & Code will be released \\
\cite{min2018new} & ViBe + SVM + CNN & & IoU & Region Matching algorithm & O & \\
\cite{bullinger2017instance} & Multi-task Network Cascades (CNN) & Optical flow & Overlap of segmentation instances & Hungarian algorithm & O & \\
\cite{wang2014learning} & Dalal-Triggs detector & Autoencoders & SVM & Minimum spanning tree & O & \\
\cite{kim2015multiple} & Public & CNN + PCA & Multi-Output Regularized Least Squares & Variation of Multiple Hypothesis Tracking & O & \href{http://rehg.org/mht/}{Source} \\
\cite{chen2017enhancing} & Public & CNN, Kalman Filter & Multi-Output Regularized Least Squares + Kalman Filter + detection-scene score & Maximum Weighted Independent Set & B & \\
\cite{yang2017hybrid} & Public & R-CNN & Observation cost + transition cost + birth-death cost & Min-cost multi commodity flow problem, solved with Dantzig-Wolfe decomposition & O & \\
\cite{wang2017robust} & DoH \cite{qian2014automatically} & CNN & CNN + Kalman filter & Custom algorithm, SVM & B & \\
\cite{wojke2017simple} & From \cite{yu2016poi} & Kalman filter, Wide Residual Net & Mahalanobis dist. (mot.) + cosine distance (app.), IoU & Hungarian algorithm & O & \href{https://github.com/nwojke/deep_sort}{Source} \\
\cite{mahmoudi2019cnnmtt} & From \cite{yu2016poi} & CNN & appearance + motion + dynamic affinity & Hungarian algorithm & O & \\
\cite{kim2018multi} & Public & CNN & Bilinear LSTM & Variant of MHT-DAM \cite{kim2015multiple} & B & \\
\cite{bae2017confidence} & Public / SDP+RPN & CNN & Appearance + motion + shape affinities & Hungarian algorithm & O & \href{https://cvl.gist.ac.kr/project/cmot.html}{Source} \\
\cite{ullah2018deep} & Public & GoogLeNet CNN & App. similarity & Bayesian inference using \cite{pirsiavash2011globally} & B & \\
\cite{fang2018recurrent} & Public / Faster R-CNN & GoogLeNet CNN & Recurrent Autoregressive Networks (GRU-based) & Bipartite graph matching & O & \\
\cite{fu2018gm} & Public & CNN & Hybrid Likelihood Function (Discriminative Correlation Filter + Gaussian Mixture Probability Hypothesis Density) & Hungarian algorithm & O & \\
\cite{wen2019learning} & Public & CNN & app. + HSV histogram + motion similarities & Pairwise update algorithm + SSVM & B & Will be available at \href{https://github.com/longyin880815}{this link} \\
\cite{sheng2018heterogeneous} & Public & GoogLeNet CNN, Optical flow & Distance between app. features, common superpixels, optical flow predictions & Multiple Hypotheses Tracking & B & \\
\cite{chen2019recurrent} & Public & CNN & LSTM (app.) + motion affinity & Batch Multi-Hypothesis & B & \\
\cite{tang2017multiple} & Public / From \cite{yu2016poi} & DeepCut CNN \cite{pishchulin2016deepcut}, StackNetPose CNN & StackNetPose CNN & Lifted multicut problem, solved as in \cite{keuper2015efficient} & B & \href{https://www.mpi-inf.mpg.de/departments/computer-vision-and-multimodal-computing/research/people-detection-pose-estimation-and-tracking/multiple-people-tracking-with-lifted-multicut-and-person-re-identification/}{Source} \\
\cite{kim2016similarity} & Public & Siamese CNN & Euclidean distance (app. feat.) + IoU + box area ratio & Custom greedy algorithm & O & \\
\cite{wang2016joint} & DPM & Siamese CNN with temporal constraints & Mahalanobis distance (app. feat.) + motion affinity & Generalized Linear Assignment solved with Softassign \cite{gold1996softmax}, Dual-threshold strategy \cite{huang2008robust} & B & \\
\cite{zhang2016tracking} & HeadHunter \cite{benenson2012pedestrian} & CNN & Euclidean distance (app. feat.), temporal and kinematic affinities & Hungarian algorithm, Agglomerative clustering & B & \href{https://github.com/shunzhang876/AdaptiveFeatureLearning}{Source} \\
\cite{leal2016learning} & Public & Siamese CNN, contextual features & Gradient Boosting & Linear programming & B & \\
\cite{son2017multi} & Public & CNN, sequence-specific statistics, optical flow, FC layers & FC layer combining app. and mot. distances & Minimax label propagation & B & \\
\cite{maksai2018eliminating} & Public & CNN + various app. and non-app. feat. & embedding layer + bidirectional LSTM & Variation of Multiple Hypothesis Tracking & B & \\
\cite{zhu2018online} & Public & Linear motion model, Spatial Attention Network CNN & Temporal Attention Network (bidirectional LSTM) & Custom greedy algorithm, ECO (SOT tracker) & O & \href{https://github.com/jizhu1023/DMAN_MOT}{Source} \\
\cite{ma2018trajectory} & Public & Siamese CNN, LSTM, WRN CNN, Siamese Bi-GRU + CNN & Euclidean dist. (app. feat.), spatial distance, GRU feature matching & Hungarian algorithm, bi-GRU RNN (track split), custom algorithm & B & \\
\cite{zhou2018deep} & Public & DCCRF, visual-displacement CNN & Visual-similarity CNN, IoU & Hungarian algorithm & O & \\
\cite{long2018tracking} & Public & R-FCN + Kalman Filter, GoogleNet & Eucl. dist. (app. feat.), IoU & Hierarchical Data Association & O & \href{https://github.com/longcw/MOTDT}{Source} \\
\cite{lee2019multiple} & Public & Feature Pyramid Siamese Network, motion features & Feature Pyramid Siamese Network & Custom greedy algorithm & O & \\
\cite{ullah2017hierarchical} & Public & Kalman Filter, GoogLeNet & Distance between sparse coding of features using a learned dictionary & Hungarian algorithm & B & \\
\cite{sadeghian2017tracking} & Public & 3 LSTMs (app., mot., interaction features) using CNN, bb velocity, occupancy map & LSTM & Hungarian algorithm, SOT tracker \cite{xiang2015learning} & O & \\
\cite{chu2017online} & Public & Linear motion model, CNN & CNN & Association to highest classification score & O & \\
\cite{ullah2018directed} & Manually generated & Hidden Markov Models, CNN & Mutual information (app. feat.) & Dynamic programming algorithm from \cite{pirsiavash2011globally} & B & \\
\cite{wang2017online} & Public & LK optical flow, Convolutional Correlation Filter CNN, Kalman filter & Optical flow aff., app. feat. aff., IoU, scale affinity, distance between detections & Custom algorithm (with Hungarian alg.) & O & \href{http://faculty.neu.edu.cn/ise/wanglu/CCF_MOT.htm}{Source} \\
\cite{rosello2018multi} & Public & Kalman filter + Deep RL agent & IoU & Hungarian algorithm + Deep RL agent & O & \\
\cite{babaee2018occlusion} & N/A & LSTM (mot.) & Stitching score using IoU & Custom iterative tracklet-stitching algorithm & B & \\
\cite{milan2017online} & Public & RNN (mot.) & LSTM & RNN & O & \href{https://bitbucket.org/amilan/rnntracking}{Source} \\
\cite{liang2018lstm} & Public & 2 LSTMs, VGG16 CNN & SVM, Siamese LSTM & Greedy association & B & \\
\cite{wan2018online} & From \cite{yu2016poi} & Kalman filter or LK optical flow, CNN + motion features & IoU, Siamese LSTM & Hungarian algorithm & B & \\
\cite{yoon2019data} & Public & & FC layers + Bi-directional LSTM & Hungarian algorithm & O & \\
\cite{chen2017online} & Public / from \cite{sanchez2016online} (combines DPM, SDP and ACF) & Modified Faster R-CNN & Modified Faster R-CNN & Particle filter & O & \\
\cite{ma2018customized} & Public & DeepMatching, Siamese CNN & Edge potential as in \cite{tang2016multi}, Siamese CNN & Lifted multicut & B & \\
\cite{ren2018collaborative} & Public & CNN (motion pred.), part of MDNet (CNN) & N/A & Deep RL agents & O & \\
\bottomrule
\caption{Information summary about the methods commented in section \ref{sec:main}. In each column, the approach for each paper in that step is shown. \textit{app.} means appearance, \textit{mot.} means motion, \textit{feat.} means features, \textit{pred.} means prediction; \textit{O} and \textit{B} in the Mode column indicate Online and Batch methods respectively. Text in the last column is clickable and contains links to the specified data.}\\
\label{tab:summary_table}\\
\end{longtable}

\end{document}